\documentclass[review]{elsarticle}
\usepackage[latin9]{inputenc}
\usepackage{array}
\usepackage{float}
\usepackage{units}
\usepackage{textcomp}
\usepackage{url}
\usepackage{multirow}
\usepackage{amsmath}
\usepackage{amsthm}
\usepackage{graphicx}

\makeatletter

\DeclareTextSymbolDefault{\textquotedbl}{T1}
\providecommand{\tabularnewline}{\\}
\floatstyle{ruled}
\newfloat{algorithm}{tbp}{loa}
\providecommand{\algorithmname}{Algorithm}
\floatname{algorithm}{\protect\algorithmname}

\theoremstyle{plain}
\newtheorem{thm}{\protect\theoremname}
\theoremstyle{remark}
\newtheorem{rem}[thm]{\protect\remarkname}
\theoremstyle{definition}
\newtheorem{defn}[thm]{\protect\definitionname}

\journal{Journal of Artificial Intelligence}

\usepackage{times}
\usepackage{latexsym}
\usepackage[algo2e]{algorithm2e}

\usepackage{color}
\usepackage{units}
\usepackage{comment}

\usepackage{graphicx}
\usepackage{subfigure}
\usepackage{booktabs}% for professional tables

\makeatother

\providecommand{\definitionname}{Definition}
\providecommand{\remarkname}{Remark}
\providecommand{\theoremname}{Theorem}

\begin{document}
\begin{frontmatter}

\title{Learning to Play Two-Player Perfect-Information Games without Knowledge}

\author[mymainaddress]{Quentin Cohen-Solal}

\address[mymainaddress]{LAMSADE, Université Paris-Dauphine, PSL, CNRS, France}

\address{CRIL, Univ. Artois and CNRS, F-62300 Lens, France\\
Normandie University, UNICAEN, CNRS, GREYC, 14 000 Caen, France.}

\ead[url]{quentin.cohen-solal@dauphine.psl.eu}
\begin{abstract}
In this paper, several techniques for learning game state evaluation
functions by reinforcement are proposed. The first is a generalization
of tree bootstrapping (tree learning): it is adapted to the context
of reinforcement learning without knowledge based on non-linear functions.
With this technique, no information is lost during the reinforcement
learning process. The second is a modification of Unbounded Best-First
Minimax extending the best sequences of actions to the terminal states.
This modified search is intended to be used during the learning process.
The third is to replace the classic gain of a game ($+1$ / $-1$)
with a \emph{reinforcement heuristic}. We study particular reinforcement
heuristics such as: quick wins and slow defeats ; scoring ; mobility
or presence. The fourth is the completion technique which takes into
account the resolution of states. The fifth is a new action selection
distribution. The conducted experiments suggest that these new techniques
improve the level of play. 

Moreover, we combine all these techniques within an algorithm called
Athénan. We compare Athénan with ExIt, a state-of-the-art algorithm
for self-play reinforcement learning without knowledge. This comparison
shows that Athénan is more performant. Finally, we apply Athénan to
design program-players to the following games: Hex, Othello, Arimaa.
The state-of-the-art level of theses games are surpassed without using
prior knowledge. We also apply Athénan to the single-player game Morpion
Solitaire which allowed to reach the state-of-the-art level without
using prior knowledge too. 
\end{abstract}
\end{frontmatter}

\noindent\begin{minipage}[t]{1\columnwidth}%
\global\long\def\et{\ \wedge\ }%

\global\long\def\terminal{\mathrm{t}}%

\global\long\def\joueur{\mathrm{j}}%

\global\long\def\joueurUn{\mathrm{1}}%

\global\long\def\joueurDeux{\mathrm{2}}%

\global\long\def\fbin{\mathrm{f_{b}}}%

\global\long\def\Actions{\mathcal{A}}%

\global\long\def\etats{\mathcal{S}}%

\global\long\def\random{\mathrm{random}\left(\right)}%
\end{minipage}

\global\long\def\terminalrandom{\mathrm{t_{r}}}%
\global\long\def\hfb{\mathrm{b_{t}}}%
\global\long\def\hfp{\mathrm{p_{t}}}%
\global\long\def\hadapt{f_{\theta}}%
\global\long\def\hadaptnum#1{f_{\theta_{#1}}}%
\global\long\def\itreeset{\mathbf{T_{i}}}%
\global\long\def\treeset{\mathbf{T}}%
\global\long\def\rootset{\mathbf{R}}%
\global\long\def\hterminal{f_{\mathrm{t}}}%
\global\long\def\rnap{\mathrm{p_{rna}}}%
\global\long\def\rnab{\mathrm{b_{rna}}}%
\global\long\def\ubfm{\mathrm{UBFM}}%
\global\long\def\ubfmt{\ubfm_{\mathrm{s}}}%
\global\long\def\argmax{\operatorname*{\mathrm{arg\,max}}}%
\global\long\def\argmin{\operatorname*{\mathrm{arg\,min}}}%
\global\long\def\liste#1#2{\left\{  #1\,|\,#2\right\}  }%
\global\long\def\fterminal{\hterminal}%
\global\long\def\fadapt{\hadapt}%
\global\long\def\id{\mathrm{ID}}%
\global\long\def\minimum{\operatorname*{\mathrm{min}}}%

\section{Introduction}

One of the most difficult tasks in artificial intelligence is the
\emph{sequential decision making problem} \citep{littman1996algorithms},
whose applications include robotics and games. As for games, the successes
are numerous. Machine surpasses man for several games, such as backgammon,
checkers, chess, and go \citep{silver2017mastering}. A major class
of games is the set of two-player games in which players play in turn,
without any chance or hidden information. This class is sometimes
called two-player \emph{perfect information}\footnote{With some definitions, perfect information games include games with
chance. This is not the case in this article} games \citep{mycielski1992games} or also two-player combinatorial
games. There are still many challenges for these games. For example,
at the \emph{game of Hex}, computers have only been able to beat strong
humans since 2020~\citep{cazenave2020polygames}. In the context
of \emph{general game playing} \citep{genesereth2005general}, i.e.
the context of playing a unknown game with only a short time to learn
how to play well: man is always superior to machine (even restricted
to games with perfect information). In this article, we focus on two-player
zero-sum games with perfect information. Note that the algorithms
we propose also apply in the single-player case. We demonstrate this
in the context of Morpion Solitaire.

The first approaches used to design a game-playing program are based
on a game tree search algorithm, such as \emph{minimax}, combined
with a handcrafted game state evaluation function based on expert
knowledge. A notable use of minimax with an handcrafted evaluation
is the Deep Blue chess program \citep{campbell2002deep}. However,
the success of Deep Blue is largely due to the raw power of the computer,
which could analyze two hundred million game states per second. In
addition, this approach is limited by having to design an evaluation
function manually (at least partially). The design of evaluation functions
is a very complex task, which must, in addition, be carried out for
each different game. Several works have thus focused on automatic
learning of evaluation functions \citep{mandziuk2010knowledge}. One
of the first successes about learning evaluation functions is on the
Backgammon game \citep{tesauro1995temporal}. However, for many games,
such as Hex or Go, minimax-based approaches, with or without machine
learning, have failed to overcome human. Two causes have been identified
\citep{baier2018mcts}. Firstly, the very large number of possible
actions at each game state prevents an exhaustive search at a significant
depth (the game can only be anticipated a few turns in advance). Secondly,
for these games, no sufficiently powerful evaluation function could
be identified. An alternative approach has been proposed to solve
the two problems, called Monte Carlo tree search and denoted MCTS
\citep{Coulom06,browne2012survey}. MCTS explores the game tree non-uniformly,
which is a solution to the problem of the very large number of actions.
In addition, it evaluates the game states from victory statistics
of a large number of random end-game simulations. Thus, it does not
need an evaluation function. MCTS gived notably good results to Hex
and Go. However, this was not enough to go beyond the level of human
players. Several variants of Monte Carlo tree search were then proposed,
using in particular knowledge to guide the exploration of the game
tree and/or random end-game simulations \citep{browne2012survey}.
Recent improvements in Monte Carlo tree search have focused on the
automatic learning of MCTS knowledge and their uses. This knowledge
was first generated by \emph{supervised learning} \citep{clark2015training,gao2017move,gao2018three,cazenave2018residual,tian2015better}
then by supervised learning followed by \emph{reinforcement learning}
\citep{silver2016mastering}, and finally by only reinforcement learning
\citep{silver2017mastering,anthony2017thinking,silver2018general}.
These learning-based improvements allowed programs to reach and surpass
the level of world champion at the game of Go~\citep{silver2016mastering,silver2017mastering}.
In particular, the program AlphaGo Zero \citep{silver2017mastering},
which only uses reinforcement learning, did not need any knowledge
to reach its level of play. This last success, however, required 29
million games of self-play (with 1,600 state evaluations per move).
The AlphaGo Zero approach has also been generalized and applied to
chess \citep{silver2017mastering2} under the name of \emph{AlphaZero}.
The resulting program broke the best chess program (based on minimax).
 Another state-of-the-art zero-knowledge reinforcement learning algorithm
is the ExIt algorithm~\citep{anthony2017thinking}, also based on
Monte Carlo tree search.

It is therefore questionable whether minimax is totally out of date
or whether the spectacular successes of recent programs are more based
on reinforcement learning than Monte Carlo tree search. In particular,
it is interesting to ask whether reinforcement learning would enhance
minimax enough to make it competitive with Monte Carlo tree search
on games where it dominates minimax so far, such as Go or Hex.

In this article\footnote{Note that this paper is an extended, improved, and english version
of~\citep{cohen2019apprendre}.}, we therefore focus on reinforcement learning within the minimax
framework. We propose and asses new techniques for reinforcement learning
of evaluation functions, which we then combine together to form the
Athénan algorithm. In particular, we show that Athénan makes it possible
to go beyond the state-of-the-art of reinforcement learning without
knowledge, in particular the ExIt algorithm, and the state-of-the-art
dedicated to the following games: Hex, Arimaa, Othello, and Morpion
Solitaire.  Note that, we showed, in collaboration with Tristan
Cazenave, in an another article produced after this article but published
before, that Athénan is more efficient and performant than AlphaZero~\citep{cohen2023minimax}.
This further study complements the studies carried out within this
article.

In the next section, we briefly present game algorithms and in particular
\emph{Unbounded Best-First Minimax} on which we base several of our
experiments. We also present reinforcement learning in games and the
games on which our experiments are performed. In the following sections,
we propose different techniques aimed at improving learning performances
and we expose the experiments carried out using these new techniques.
In particular, in Section~\ref{sec:Data-Usage}, we extends the \emph{tree
bootstrapping} (\emph{tree learning}) technique to the context of
reinforcement learning without knowledge based on non-linear functions.
In Section~\ref{sec:Search-Algorithms-for-Learning}, we present
a new search algorithm, a variant of Unbounded Best-First Minimax
called \emph{Descent}, intended to be used during the learning process.
In Section~\ref{sec:Reinforcement-Heuristic}, we introduce \emph{reinforcement
heuristics.} Their usage is a simple way to use general or dedicated
knowledge in reinforcement learning processes. We study several reinforcement
heuristics in the context of different games.  In Section~\ref{subsec:Completion},
we introduce the\emph{ completion technique }for taking into account
the resolution of states in the cases of Unbounded Minimax and Descent\emph{.}
Section~\ref{sec:Ordinal-Distribution} introduce a new action selection
distribution. 

Then, we combine previously introduced techniques into the algorithm
called \emph{Athénan} and apply it in different contexts. In Section~\ref{sec:Application-to-Hex},
we apply Athénan to design program-players to the game of Hex (size
$11$ and $13$) and compare them to Mohex 3HNN \citep{gao2018three},
the best Hex program (before Athénan), notably champion at Hex (size
11 and 13) during the $2018$ Computer Olympiad \citep{gao2019hex}.
In Section~\ref{sec:Comparison-with-ExIt}, we compare Athénan with
ExIt~\citep{anthony2017thinking}, the state-of-the-art reinforcement
learning algorithm without knowledge at Hex. Then, in Section~\ref{sec:Application-at-Othello},
we apply Athénan to surpass the dedicated state-of-the-art at Othello.
In Section~\ref{sec:Application-at-Arimaa}, we apply Athénan to
surpass the dedicated state-of-the-art at Arimaa. In Section~\ref{fig:morpion},
we apply Athénan to reach the level of the dedicated state-of-the-art
at Morpion Solitaire.   

Moreover, in Section~\ref{sec:Computer-Olympiad-Results}, we present
the results of Athénan during the worldwide artificial intelligence
competition on board games: the Computer Olympiad.

Finally, in Section~\ref{sec:Conclusion}, we conclude and expose
the different research perspectives.

\section{Background and Related Work\label{sec:RW}}

In this section, we briefly present game tree search algorithms, reinforcement
learning in the context of games, and the games used in this paper
(for more details about game algorithms, see \citep{yannakakis2018artificial}).

Games can be represented by their \emph{game tree} (a node corresponds
to a game state and the children of a node are the states that can
be reached by an action). From this representation, we can determine
the action to play using a game tree search algorithm. In order to
win, each player tries to maximize his score (i.e. the value of the
game state for this player at the end of the match). As we place ourselves
in the context of two-player zero-sum games, to maximize the score
of a player is to minimize the score of his opponent (the score of
a player is the negation of the score of his opponent).

\subsection{Game Tree Search Algorithms}

The central algorithm is \emph{minimax} which recursively determines
the value of a node from the value of its children and the functions
$\min$ and $\max$, up to a limit recursion depth. With minimax,
the game tree is uniformly explored. A better implementation of minimax
uses \emph{alpha-beta pruning} \citep{knuth1975analysis,yannakakis2018artificial}
which makes it possible not to explore the sections of the game tree
which are less interesting given the values of the nodes already met
and the properties of $\min$ and $\max$. Many variants and improvements
of minimax have been proposed \citep{millington2009artificial}. For
instance, \emph{iterative deepening} \citep{slate1983chess,korf1985depth}
allows one to use minimax with a time limit. It sequentially performs
increasing depth alpha-beta searches as long as there is time. It
is generally combined with the \emph{move ordering} technique \citep{fink1982enhancement},
which consists of extending the best move from the previous search
first, which accelerates the new search. 

Some variants perform a search with unbounded depth (that is, the
depth of their search is not fixed) \citep{van2008proof,schaeffer1990conspiracy,berliner1981b}.
Unlike minimax with or without alpha-beta pruning, the exploration
of these algorithms is non-uniform. One of these algorithms is the\emph{
best-first minimax search }\citep{korf1996best}\footnote{At the time of the first experiments of this article, we were not
aware that the Unbounded Best-First Minimax algorithm had already
been proposed, so we rediscovered it independently during this work.}. To avoid any confusion with some best-first approaches at fixed
depth, we call this algorithm \emph{Unbound Best-First Minimax}, or
more succinctly $\ubfm$. $\ubfm$ iteratively extends the game tree
by adding the children of one of the leaves of the game tree having
the same value as that of the root (minimax value). These leaves are
the states obtained after having played one of the best sequences
of possible actions given the current partial knowledge of the game
tree. Thus, that algorithm iteratively extends the \emph{a priori}
best sequences of actions. The best sequences usually change at each
extension. Thus, the game tree is non-uniformly explored by focusing
on the \emph{a priori} most interesting actions without exploring
just one sequence of actions. 

In this article, we use the anytime version of $\ubfm$ \citep{korf1996best},
i.e. we leave a fixed search time for $\ubfm$ to decide the action
to play. We also use transposition tables \citep{greenblatt1988greenblatt,millington2009artificial}
with $\ubfm$, which makes it possible not to explicitly build the
game tree and to merge the nodes corresponding to the same state.
(MCTS and Alpha-Beta in this paper also use transposition tables).
Note that positive impact of transposition tables has been evaluated
\citep{cohen2025little}. Algorithm~\ref{alg:-Unbounded-Best-First-Mininmax}
is an implementation of $\ubfm$\footnote{This implementation is a slight variant of Korf and Chickering algorithm.
Both algorithms behave identically if the evaluation function gives
a different value to each state and if transposition tables are not
used. In practice, in the context of using transpositions, our variant
improves win performance by a few percentage points \citep{cohen2025little}.
If the evaluation function gives a different value to each state and
transpositions are not used, Korf and Chickering algorithm is slightly
faster (carelessly faster in some contexts).}. $\ubfm$ has been improved within the algorithm called \emph{Unbounded
Minimax with Safe Decision}~\citep{cohen2025study}, denoted $\ubfmt$.
This variant performs the same search in the game tree but plays a
different action after performing this search. Instead of playing
the action that leads to the state of best value, the action played
is the one that was most selected from the current state during the
search. This variant of Unbounded Minimax was originally proposed
in the old, unpublished versions of this article~\citep{cohen2020learning}.
It has been removed from this document to simplify it. The removed
results about Unbounded Minimax with Safe Decision that appeared in
the old versions of this article have been improved and detailed in
the article that deals with its new introduction~\citep{cohen2025study}.
\begin{table}
\begin{centering}
{\footnotesize{}}%
\begin{tabular}{|c|c|}
\hline 
{\footnotesize{}Symbols} & {\footnotesize{}Definition}\tabularnewline
\hline 
\hline 
{\footnotesize{}$\mathrm{actions}\left(s\right)$} & {\footnotesize{}action set of the state $s$ for the current player}\tabularnewline
\hline 
{\footnotesize{}$\mathrm{first\_player}\left(s\right)$} & {\footnotesize{}true if the current player of the state $s$ is the
first player}\tabularnewline
\hline 
{\footnotesize{}$\mathrm{terminal\left(s\right)}$} & {\footnotesize{}true if $s$ is an end-game state}\tabularnewline
\hline 
{\footnotesize{}$a(s)$} & {\footnotesize{}state obtained after playing the action $a$ in the
state $s$}\tabularnewline
\hline 
{\footnotesize{}$\mathrm{time}\left(\right)$} & {\footnotesize{}current time in seconds}\tabularnewline
\hline 
{\footnotesize{}$\random$} & {\footnotesize{}returns a uniformly random number from $[0,1]$}\tabularnewline
\hline 
{\footnotesize{}$S$} & {\footnotesize{}keys of the transposition table $T$}\tabularnewline
\hline 
{\footnotesize{}$T$} & {\footnotesize{}transposition table (contains states labels as function
$v$ ; depends on the used search algorithm)}\tabularnewline
\hline 
{\footnotesize{}$\tau$} & {\footnotesize{}search time per action}\tabularnewline
\hline 
{\footnotesize{}$t$} & {\footnotesize{}time elapsed since the start of the reinforcement
learning process}\tabularnewline
\hline 
{\footnotesize{}$t_{\max}$} & {\footnotesize{}chosen total duration of the learning process }\tabularnewline
\hline 
{\footnotesize{}$n(s,a)$} & {\footnotesize{}number of times the action $a$ is selected in state
$s$ (initially, $n(s,a)=0$ for all $s$ and $a$}\tabularnewline
\hline 
{\footnotesize{}$v(s)$} & {\footnotesize{}value of state $s$ in the game tree}\tabularnewline
\hline 
{\footnotesize{}$v(s,a)$} & {\footnotesize{}value obtained after playing action $a$ in state
$s$}\tabularnewline
\hline 
{\footnotesize{}$c(s)$} & {\footnotesize{}completion value of state $s$ ($0$ by default)}\tabularnewline
\hline 
{\footnotesize{}$c(s,a)$} & {\footnotesize{}completion value obtained after playing action $a$
in state $s$}\tabularnewline
\hline 
{\footnotesize{}$r(s)$} & {\footnotesize{}resolution value of state $s$ ($0$ by default)}\tabularnewline
\hline 
{\footnotesize{}$r(s,a)$} & {\footnotesize{}resolution value obtained after playing action $a$
in state $s$}\tabularnewline
\hline 
{\footnotesize{}$f(s)$} & {\footnotesize{}the used evaluation function (first player point of
view)}\tabularnewline
\hline 
{\footnotesize{}$\hadapt(s)$} & {\footnotesize{}adaptive evaluation function (of non-terminal game
tree leaves ; first player point of view)}\tabularnewline
\hline 
{\footnotesize{}$\hterminal(s)$} & {\footnotesize{}evaluation of terminal states, e.g. gain game (first
player point of view)}\tabularnewline
\hline 
{\footnotesize{}Gain function $\hfb(s)$} & {\footnotesize{}$0$ if $s$ is a draw, $1$ if $s$ is winning for
the first player, $-1$ if $s$ is losing for the first player}\tabularnewline
\hline 
\multirow{3}{*}{{\footnotesize{}search($s$, $S$, $T$, $\hadapt$, $\hterminal$)}} & {\footnotesize{}a seach algorithm (it extends the game tree from $s$,
by adding new states in $S$ }\tabularnewline
 & {\footnotesize{}and labeling its states, in particular, by a value
$v(s)$, stored in $T$, }\tabularnewline
 & {\footnotesize{}using $\hadapt$ as evaluation of the non-terminal
leaves and $\hterminal$ as evaluation of terminal states}\tabularnewline
\hline 
{\footnotesize{}action\_selection($s$, $S$, $T$)} & {\footnotesize{}decides the action to play in the state $s$ depending
on the partial game tree, i.e. on $S$ and $T$}\tabularnewline
\hline 
{\footnotesize{}processing($D$)} & {\footnotesize{}various optional data processing: data augmentation
(symmetry), experience replay, ... }\tabularnewline
\hline 
{\footnotesize{}update($\hadapt,D$)} & {\footnotesize{}updates the parameter $\theta$ of $\hadapt$ in order
for $\hadapt(s)$ is closer to $v$ for each $(s,v)\in D$}\tabularnewline
\hline 
\end{tabular}{\footnotesize\par}
\par\end{centering}
\caption{Index of symbols\label{tab:Index-of-symbols}}
\end{table}

\begin{algorithm}
\DontPrintSemicolon\SetAlgoNoEnd
\SetKwFunction{iterationUM}{UBFM\_iteration}
\SetKwFunction{UM}{UBFM}\SetKwFunction{time}{time}\SetKwFunction{actions}{actions}\SetKwFunction{bestaction}{best\_action}\SetKwFunction{terminal}{terminal}\SetKwFunction{premier}{first\_player} \SetKwProg{myproc}{Function}{}{}
\myproc{\iterationUM{$s$, $S$, $T$}}{
\eIf{\terminal{$s$}}{return $f(s)$}{
\eIf{$s\notin S$}{
$S\leftarrow S\cup\{s\}$\;

\ForEach{$a\in$ \actions{$s$}}{$v(s,a)\leftarrow f\left(a(s)\right)$\; }
}{
$a_{b}\leftarrow$ \bestaction{$s$}\;

$v(s,a_{b})\leftarrow$ \iterationUM{$a_{b}(s)$, $S$, $T$}\;

} 

$a_{b}\leftarrow$ \bestaction{$s$}\; 

return $v(s,a_{b})$\;
}
}
\;
\myproc{\bestaction{$s$, $S$, $T$}}{
\eIf{\premier{$s$}}{return ${\displaystyle \argmax_{a\in\actions{s}}v\left(s,a\right)}$\;}{
return ${\displaystyle \argmin_{a\in\actions{s}}v\left(s,a\right)}$\;
}
}
\;
\myproc{\UM{$s$, $\tau$}}{
$t=$ \time{}\;
\lWhile{\time{}$-\,t<\tau$}{\iterationUM{$s$, $S$, $T$}}
return \bestaction{$s$, $S$, $T$}\;
}\;

\caption{$\protect\ubfm$ (Unbounded Best-First Minimax) algorithm : it computes
the best action to play in the generated non-uniform partial game
tree (see Table~\ref{tab:Index-of-symbols} for the definitions of
symbols ; at any time $T=\protect\liste{v(s,a)}{s\in S,a\in\mathrm{actions}(s)}$).\label{alg:-Unbounded-Best-First-Mininmax}}
\end{algorithm}

\subsection{Learning of Evaluation Functions\label{subsec:RW-Apprentissage-de-fonctions}}

Reinforcement learning of evaluation functions can be done by different
techniques \citep{mandziuk2010knowledge,silver2017mastering,anthony2017thinking,young2016neurohex}.
The general idea of reinforcement learning of state evaluation functions
is to use a game tree search algorithm and an adaptive evaluation
function to play a sequence of matches (for example against oneself,
which is the case in this article). Adaptive evaluation functions
$\hadapt$ are usually neural networks, of parameter $\theta$. Each
match generates pairs $(s,v)$ where $s$ is a game state and $v$
is the value of $s$ calculated by the chosen search algorithm using
the evaluation function $\hadapt$. Such pairs are learned for updating
$\hadapt$ in order to improve it. The set of states used for updating
$\hadapt$ varies depending on the learning technique \citep{tesauro1995temporal,veness2009bootstrapping}.
For example, in the case of \emph{root bootstrapping} (technique that
we call \emph{root learning} in this article), the set of pairs learned
during the learning phase is $D=\liste{\left(s,v\right)}{s\in\rootset}$
with $\rootset$ the set of states of the match. In the case of the
\emph{tree bootstrapping} (\emph{tree learning}) technique \citep{veness2009bootstrapping},
the set of pairs learned during the learning phase is the set of states
of the partial game tree built to decide which actions to play: $D=\liste{\left(s,v\right)}{s\in\treeset}$
with $\treeset$ the set of states of the search trees. Note that
states of the match search trees include states of the match (since
the states of the match are the roots of the search trees). Thus,
contrary to root bootstrapping, tree bootstrapping does not discard
most of the information used to decide the actions to play. The values
of the generated states can be their minimax values in the partial
game tree built to decide which actions to play~\citep{veness2009bootstrapping,tesauro1995temporal}.
Work on tree bootstrapping has been limited to reinforcement learning
of linear functions of state features. It has not been formulated
or studied in the context of reinforcement learning without knowledge
and based on non-linear functions. Note that, in the case of AlphaGo
Zero, the learned states are the states of the sequence of the match
(as with root learning), but the target value for each of these states
is the value of the terminal state of the match \citep{silver2017mastering}.
We call that technique terminal learning. 

Generally between two consecutive matches (between \emph{match phases}),
a \emph{learning phase} occurs, using the pairs of the last match.
Each learning phase consists in modifying $\hadapt$ so that for all
pairs $(s,v)\in D$, $\hadapt(s)$ sufficiently approaches $v$ to
constitute a good approximation. Note that, learning phases can also
use the pairs of \emph{several} matches. This technique is called
\emph{experience replay} \citep{mnih2015human}. Note that, adaptive
evaluation functions $\hadapt$ only serve to evaluate non-terminal
states since we know the true value of terminal states (it is important:
using the exact terminal evaluation function improves confrontation
performance \citep{cohen2025little}). 
\begin{rem}
There is another reinforcement learning technique for games: the Temporal
Differences algorithm $\mathrm{TD}(\lambda)$~\citep{tesauro1995temporal}.
It corresponds to a root learning algorithm without search (i.e. with
a search of depth $1$) where the target of a state is a weighted
average of the values of its successor states in the match, parameterized
by the constant $\lambda$. A variant of $\mathrm{TD}(\lambda)$ has
also been proposed~\citep{baxter2000learning}, called $\mathrm{TD_{Leaf}}(\lambda)$,
where the temporal difference technique $\mathrm{TD}(\lambda)$ is
modified by applying it with a minimax search (i.e. with a search
of depth $d\geq1$). A comparison between these techniques was made~\citep{veness2009bootstrapping}.
Finally, in \citep{bjornsson2003learning} a method is described for
automatically tuning \emph{search-extension} parameters, to decide
which branches of the game tree must be explored during the search.
\end{rem}

\subsection{Action Selection Distribution\label{subsec:Distribution-de-s=00003D0000E9lection}}

One of the problems related to reinforcement learning is the \emph{exploration-exploitation
dilemma} \citep{mandziuk2010knowledge}. It consists of choosing between
exploring new states to learn new knowledge and exploiting the acquired
knowledge. Many techniques have been proposed to deal with this dilemma
\citep{mellor2014decision}. However, most of these techniques do
not scale because their application requires memorizing all the encountered
states. For this reason, in the context of games with large numbers
of states, some approaches use probabilistic exploration \citep{young2016neurohex,silver2017mastering,mandziuk2010knowledge,schraudolph2001learning}.
With this approach, to exploit is to play the best action and to explore
is to play uniformly at random. More precisely, a parametric probability
distribution is used to associate with each action its probability
of being played. The parameter associated with the distribution corresponds
to the exploration rate (between $0$ and $1$), which we denote $\epsilon$
(the exploitation rate is therefore $1-\epsilon$, which we denote
$\epsilon'$). The rate is often experimentally fixed. \emph{Simulated
annealing} \citep{kirkpatrick1983optimization} can, however, be applied
to avoid choosing a value for the parameter. In that case, at the
beginning of reinforcement learning, the parameter is $1$ (we are
just exploring). It gradually decreases until reaching $0$ at the
end of the learning process. The simplest action selection distribution
is \emph{$\epsilon$-greedy} \citep{young2016neurohex} (of parameter
$\epsilon$). With the distribution \emph{$\epsilon$-greedy}, the
action is chosen uniformly with probability $\epsilon$ and the best
action is chosen with probability $1-\epsilon$ (see also Algorithm~\ref{alg:e-greedy}). 

\begin{algorithm}[!bh]
\DontPrintSemicolon\SetAlgoNoEnd

\SetKwFunction{egreedy}{$\epsilon$\_greedy}\SetKwFunction{actions}{actions}\SetKwFunction{premier}{first\_player}\SetKwProg{myproc}{Function}{}{}

\myproc{\egreedy{$s$, $v$}}{

\eIf{ $\random\leq\frac{t}{t_{\max}}$ }{

\eIf{\premier{$s$}}{

return $\argmax_{a\in\text{\actions{\text{s}}}}v\left(s,a\right)$
\;

}{

return $\argmin_{a\in\text{\actions{\text{s}}}}v\left(s,a\right)$
\;

}

}{

return $a\in\text{\actions{\text{s}}}$ uniformly chosen.\;

}

}\;

\protect\protect

\caption{$\epsilon$-greedy algorithm with simulated annealing ($\epsilon=1-\frac{t}{t_{\max}}$)
used in the experiments of this article (see Table~\ref{tab:Index-of-symbols}
for the definitions of symbols).\label{alg:e-greedy}}
\end{algorithm}
The $\epsilon$-greedy distribution has the disadvantage of not differentiating
the actions (except the best action) in terms of probabilities. Another
distribution is often used, correcting this disadvantage. This is
the \emph{softmax} distribution \citep{schraudolph2001learning,mandziuk2010knowledge}.
It is defined by $P\left(a_{i}\right)=\frac{e^{\nicefrac{v(s,a_{i})}{\tau}}}{\sum_{j=1}^{n}e^{\nicefrac{v(s,a_{j})}{\tau}}}$
with $n$ the number of children of the current state $s$, $P\left(a_{i}\right)$
the probability of playing the action $a_{i}$ , $v(s,a_{i})$ the
value of the state obtained after playing $a_{i}$ in the current
state $s$, $i\in\{0,\ldots,n-1\}$, and $\tau\in]0,+\infty[$ a parameter
called temperature ($\tau\simeq0$ : exploitation, $\tau\simeq+\infty$
: exploration).

\subsection{Games of Paper Experiments}

We now briefly present the games on which experiments are performed
in this article, namely: Hex, Arimaa, Morpion Solitaire, Othello,
Surakarta, Outer Open Gomoku, Clobber, Breakthrough, Amazons, Lines
of Action, and Santorini. They are all board games. All of these games
are present and recurring (except Morpion Solitaire) at the Computer
Olympiad, the worldwide multi-games event in which computer programs
compete against each other. Moreover, all these games (and their rules)
are included (and available for free) in Ludii~\citep{Piette2020Ludii},
a general game system. 

\subsubsection{Hex}

Hex is a two-player combinatorial strategy game played on an $n\times n$
hexagonal board. Each player takes turns placing a stone of their
color on an empty cell. The objective is to be the first to connect
the two opposite sides of the board corresponding to their color.
A common variant of Hex includes the swap rule, where the second player
can choose, on their first turn, to swap roles and sides with the
first player. This rule helps balance the game by mitigating the first
player\textquoteright s advantage and is typically used in competitions.
It is still used at the Computer Olympiad, and we use it in this paper.

\subsubsection{Arimaa}

Arimaa is a game with similarities with Chess. Each player has a different
set of pieces (identical between the two players): one elephant, one
camel, two horses, two dogs, two cats, and eight rabbits. The pieces
types are fully ordered: elephant $>$ camel $>$ horse $>$ dog $>$
cat $>$ rabbit. A higher piece can move (push or pull) a lower piece.
A piece which is orthogonally adjacent to a stronger opposing piece
is frozen, unless it is also adjacent to a friendly piece. If a piece
is on a trap square without being adjacent to a friendly piece, it
is removed from the board. A player wins when one of its rabbits reaches
the other side of the board. A player without rabbits loses. It is
played on a $8\times8$ grid like Chess (the cases $(2,2)$, $(2,5)$,
$(5,2)$, $(5,5)$ are traps). At the start, the board is empty. The
first player places its pieces on its side of the board then the second
player does the same. On his turn, the player can make 4 moves of
one square (with one or more pieces). Pushing or pulling an opposing
piece counts as two moves. Details about rules are available in \url{https://arimaa.com/arimaa/}.

\subsubsection{Morpion Solitaire}

The game is played on a grid of unlimited size. The starting configuration
includes a set of points already placed on the grid in the shape of
a Greek cross whose side has 4 points (the cross is made up of a total
of 36 points).

A move in this game consists of two steps. First, the player places
a new point on the grid so as to form an alignment of five adjacent
points horizontally, vertically or diagonally (which have not already
been connected by a line). Then, he connects this alignment by drawing
a line. An alignment cannot be placed as an extension of a previous
alignment. 

The aim of the game is to place as many points as possible before
reaching a situation where no new points can be placed.

\subsubsection{Othello}

Othello (also called Reversi) is a territory and linear encirclement
game whose goal is to have more pieces than its opponent. In his turn,
a player places a piece of his color on the board (only if he can
make an encirclement, otherwise he pass his turn). There is an encirclement
if an opponent's line of pieces has at one of its ends the piece that
the current player has just placed, and has at the other end, another
piece of the current player. As a result of this encirclement, the
encircled opponent's pieces are replaced by pieces from the current
player.

\subsubsection{Surakarta}

Surakarta is a move and capture game (like draughts). The goal of
the game is to take all the opposing pieces. In his turn, a player
can either move a piece to an empty square at a distance of $1$ or
move a piece to a square occupied by an opponent's piece under certain
conditions and according to a mechanism specific to Surakarta (based
on a movement circuit dedicated only to capture), allowing \textquotedblleft long
distance\textquotedblright{} capture.

\subsubsection{Outer Open Gomoku}

Outer Open Gomoku is an alignment game. The goal of the game is to
line up at least $5$ pieces of its color. On his turn, a player places
a piece of his color. In the first turn, the first player can only
place a piece at a distance of $2$ from the sides of the board.

\subsubsection{Clobber}

Clobber is a move and capture game. The goal is to be the last player
to have played. A player can play if he can orthogonally move one
of his pieces onto a neighboring square on which there is an opponent's
piece. This movement is always a capture (the opponent's piece is
removed from the board).

\subsubsection{Breakthrough}

Breakthrough is a move and capture game. The goal of the game is to
be the first to make one of his pieces reach the other side of the
board. A piece can only move by moving forward one square (straight
or diagonal). A capture can only be done diagonally.

\subsubsection{Amazons}

Amazons is a move and blocking game. In turn, a player moves one of
his pieces in a straight line in any direction (like the queen of
the game of Chess). Then he places a neutral piece in any direction
starting from the new position of the piece just moved (always in
the manner of the queen of the game of Chess ). Any piece (neutral
or player-owned) blocks placements and movements. The goal of the
game is to be the last to play.

\subsubsection{Lines of Action}

Lines of Action is a game of movement and regrouping. On his turn,
a player can move one of his pieces in one direction as many squares
as there are pieces in that direction. A piece cannot move if there
is an opponent's piece in its path, unless it is the square to arrive
(in which case a capture is made). The goal is to have all of its
pieces connected.

\subsubsection{Santorini}

Santorini is a three-dimensional building and moving game. The goal
of the game is to reach the 3rd floor of a building. In his turn,
a player moves one of his pieces by one square then places the first
floor on an adjacent empty square or increases a pre-existing construction
by one floor (on which no player's piece is located). A piece cannot
move to a square having strictly more than one floor more than the
square where it is located (a piece only go up one floor at a time
and can descend as many floors as wanted). A move cannot be made to
a square with 4 floors. A construction cannot be done on a square
of 4 floors. A player who cannot play loses. The advanced mode (i.e.
the use of power cards) is not used in the experiments in this article.

\section{Data Use in Game Learning\label{sec:Data-Usage}}

In this section, we adapt and study tree learning (see Section~\ref{subsec:RW-Apprentissage-de-fonctions})
in the context of reinforcement learning and the use of non-linear
adaptive evaluation functions. For this purpose, we compare it to
root learning and terminal learning. We start by adapting tree learning,
root learning, and terminal learning. We also define in Section~\ref{alg:smooth-exp_replay}
a variant of experience replay that we call stratified experience
replay. Next, we describe the experiment protocol common to several
sections of this article. Finally, we expose the comparison of tree
learning with root learning and terminal learning.

\subsection{Tree Learning\label{subsec:Tree-Learning}}

As we saw in Section~\ref{subsec:RW-Apprentissage-de-fonctions},
tree learning consists in learning the value of the states of the
search tree. However, our tree learning technique has two main differences
with classic tree learning (in addition to the generalization to reinforcement
learning without knowledge). On the one hand, the learning phase is
carried out only after each game (and not after each search, i.e.
not after each action play). On the other hand, the data of the non-terminal
leaves of the tree are not learned. Thus learning is performed on
the partial game tree obtained at the end of the game from which the
non-terminal leaves have been removed. It drastically reduces the
number of data to learn and avoids an overlearning effect: our version
significantly changes the learning performance. As a reminder, root
learning consists in learning the values of the states of the sequence
of states of the match (the value of each state is its value in the
search tree). Terminal learning consists in learning the values of
the sequence of states of the match but the value of each state is
the value of the terminal state of the match (i.e. the gain of the
match). To generalize tree learning and root learning, in a manner
similar to the use of terminal learning in the AlphaZero framework,
data to learn after each game can be modified by some optional data
processing methods, such as experience replay (which involves learning
a sample of previous matches instead of just learning the entire data
from the previous match). In addition, the learning phase uses a particular
update method so that the adaptive evaluation function fit the chosen
data. The adaptation of tree learning, root learning, and terminal
learning are given respectively in Algorithm~\ref{alg:tree-learning},
Algorithm~\ref{alg:root-learning}, and Algorithm~\ref{alg:terminal-learning}.
The data processing method Experience replay is described in Algorithm~\ref{alg:exp_replay}
(its parameter are the memory size $\mu$ and the sampling rate $\sigma$).
In addition, we use in this article a stochastic gradient descent
as update method (see Algorithm~\ref{alg:sgd} ; its parameter is
$B$ the batch size). Formally, in Algorithm~\ref{alg:tree-learning},
Algorithm~\ref{alg:root-learning}, and Algorithm~\ref{alg:terminal-learning},
we have: processing($D)$ is experience\_replay($D$, $\mu$, $\sigma$)
and update($\hadapt$, $D$) is stochastic\_gradient\_descent($\hadapt$,
$D$, $B$). Finally, we use $\epsilon$-greedy as default action
selection method (i.e. action\_selection($s$, $S$, $T$) is $\epsilon$-greedy($s$,
$T.v$) (see Algorithm~\ref{alg:e-greedy} ; recall that $T$ stores
the children value function $v$)).

\begin{algorithm}[!bh]
\DontPrintSemicolon\SetAlgoNoEnd

\SetKwFunction{treelearning}{tree\_learning}\SetKwFunction{search}{search}\SetKwFunction{initial}{initial\_game\_state}\SetKwFunction{learningtime}{learning\_time}\SetKwFunction{processing}{processing}\SetKwFunction{update}{update}

\SetKwFunction{actionselection}{action\_selection}\SetKwFunction{terminal}{terminal}\SetKwFunction{leaf}{leaf}
\SetKwProg{myproc}{Function}{}{}

\myproc{\treelearning{$t_{\max}$, $\tau$}}{

$t_{0}\leftarrow$ \time{}\;

\While{\time{}$-\,t_{0}<t_{\max}$ }{

$s\leftarrow$\initial{}\;

$S\leftarrow\emptyset$\;

$T\leftarrow\{\}$\;

\While{$\neg$\terminal{$s$}}{

$S,\,T\leftarrow$\search{$s$, $S$, $T$, $\hadapt$, $\hterminal$,
$\tau$}\;

$a\leftarrow$\actionselection{$s$, $S$, $T$}\;

$s\leftarrow a(s)$

}\;

$D\leftarrow\{\left(s,v(s)\right)\ |\ s\in S\}$\;

$D\leftarrow$ \processing{$D$}

\update{$\hadapt$, $D$}

}\;

}\;

\protect\protect

\caption{Tree learning (tree bootstrapping) algorithm (see Table~\ref{tab:Index-of-symbols}
for the definitions of symbols). For tree learning, $S$ is the set
of states which are non-leaves or terminal.\label{alg:tree-learning}}
\end{algorithm}
\begin{algorithm}[!bh]
\DontPrintSemicolon\SetAlgoNoEnd

\SetKwFunction{rootlearning}{root\_learning}\SetKwFunction{search}{search}\SetKwFunction{initial}{initial\_game\_state}\SetKwFunction{learningtime}{learning\_time}\SetKwFunction{processing}{processing}\SetKwFunction{update}{update}

\SetKwFunction{actionselection}{action\_selection}\SetKwFunction{terminal}{terminal}\SetKwFunction{leaf}{leaf}
\SetKwProg{myproc}{Function}{}{}

\myproc{\rootlearning{$t_{\max}$, $\tau$}}{

$t_{0}\leftarrow$ \time{}\;

\While{\time{}$-\,t_{0}<t_{\max}$ }{

$s\leftarrow$\initial{}\;

$S\leftarrow\emptyset$\;

$T\leftarrow\{\}$\;

$D\leftarrow\emptyset$\;

\While{$\neg$\terminal{$s$}}{

$S,\,T\leftarrow$\search{$s$, $S$, $T$, $\hadapt$, $\hterminal$,
$\tau$}\;

$a\leftarrow$\actionselection{$s$, $S$, $T$}\;

$D\leftarrow D\cup\left\{ \left(s,v(s)\right)\right\} $\;

$s\leftarrow a(s)$

}\;

$D\leftarrow D\cup\left\{ \left(s,v(s)\right)\right\} $\;

$D\leftarrow$ \processing{$D$}

\update{$\hadapt$, $D$}

}\;

}\;

\protect\protect

\caption{Root learning (root bootstrapping) algorithm (see Table~\ref{tab:Index-of-symbols}
for the definitions of symbols).\label{alg:root-learning}}
\end{algorithm}
\begin{algorithm}[!bh]
\DontPrintSemicolon\SetAlgoNoEnd

\SetKwFunction{terminalearning}{terminal\_learning}\SetKwFunction{search}{search}\SetKwFunction{initial}{initial\_game\_state}\SetKwFunction{processing}{processing}\SetKwFunction{update}{update}

\SetKwFunction{actionselection}{action\_selection}\SetKwFunction{terminal}{terminal}
\SetKwProg{myproc}{Function}{}{}

\myproc{\terminalearning{$t_{\max}$, $\tau$}}{

$t_{0}\leftarrow$ \time{}\;

\While{\time{}$-\,t_{0}<t_{\max}$ }{

$s\leftarrow$\initial{}\;

$S\leftarrow\emptyset$\;

$T\leftarrow\{\}$\;

$G\leftarrow\left\{ s\right\} $\;

\While{$\neg$\terminal{$s$}}{

$S,\,T\leftarrow$\search{$s$, $S$, $T$, $\hadapt$, $\hterminal$,
$\tau$}\;

$a\leftarrow$\actionselection{$s$, $S$, $T$}\;

$s\leftarrow a(s)$\;

$G\leftarrow G\cup\left\{ s\right\} $\;

}\;

$D\leftarrow\{\left(s',\hterminal(s)\right)\ |\ s'\in G\}$\;

$D\leftarrow$ \processing{$D$}\;

\update{$\hadapt$, $D$}\;

}\;

}\;

\protect\protect

\caption{Terminal learning algorithm (see Table~\ref{tab:Index-of-symbols}
for the definitions of symbols).\label{alg:terminal-learning}}
\end{algorithm}
\begin{algorithm}[!bh]
\DontPrintSemicolon\SetAlgoNoEnd

\SetKwFunction{replay}{experience\_replay}\SetKwProg{myproc}{Function}{}{}

\myproc{\replay{$D$, $\mu$, $\sigma$}}{

elements of $D$ are added in $M$\;

\If{$\left|M\right|>\mu$}{

remove the oldest items of $M$ to have $\left|M\right|=\mu$

}

\If{$\left|M\right|\leq\sigma\cdot\mu$}{

return $M$

}

return a list of random items of $M$ whose size is $\sigma\cdot\mu$

}\;

\protect\protect

\caption{Experience replay (replay buffer) algorithm used in the experiments
of this article. $\mu$ is the memory size and $\sigma$ is the sampling
rate. $M$ is the memory buffer (global variable initialized by an
empty queue). If the number of data is less than $\sigma\cdot\mu$,
then it returns all data (no sampling). Otherwise, it returns $\sigma\cdot\mu$
random elements.\label{alg:exp_replay}}
\end{algorithm}
\begin{algorithm}[!bh]
\DontPrintSemicolon\SetAlgoNoEnd

\SetKwFunction{sgd}{stochastic\_gradient\_descent}\SetKwProg{myproc}{Function}{}{}

\myproc{\sgd{$\hadapt$, $D$, $B$}}{

Split $D$ in $m$ disjoint sets, denoted $\left\{ D_{i}\right\} _{i=1}^{m}$,
such that $D=\bigcup_{i=1}^{m}D_{i}$ and $\left|D_{i}\right|=B$
for each $i\in\left\{ 1,\ldots,m\right\} $\;

\ForEach{$i\in\left\{ 1,\ldots,m\right\} $}{

minimize $\sum_{(s,v)\in D_{i}}\left(\hadapt(s)-v\right)^{2}$ by
using Adam and $L_{2}$ regularization\;

}\;

}\;

\protect\protect

\caption{Stochastic gradient descent algorithm used in the experiments of this
article. It is based on Adam optimization ($1$ epoch per update)
\citep{kingma2014adam} and $L_{2}$ regularization (with $\lambda=0.001$
as parameter) \citep{ng2004feature} and implemented with tensorflow.
$B$ is the batch size (see Table~\ref{tab:Index-of-symbols} for
the definitions of the other symbols)\label{alg:sgd}}
\end{algorithm}

\subsection{Stratified Experience Replay\label{sec:Modified-Experience-Replay}}

In this section, we introduce the technique we call \emph{Stratified
Experience Replay}, which is a variation of Experience Replay. The
targeted objective of Stratified Experience Replay is to reduce the
variance of learning processes. 

Stratified Experience Replay consists in performing a stratified sampling
without replacement on the replay buffer of Experience Replay whose
data has been grouped in relation to the matches that generated it.
Stratified sampling imposes the same proportion of data for each previous
match, so we have a good level of variability during each learning
phase. The absence of replacement makes that no data is unnecessarily
repeated and no data is forgotten. 

Stratified Experience Replay depends on two parameters: $\mu$ the
number of previous matches kept in memory and $\delta$ the duplication
factor: the number of times a piece of data must be learned during
the learning phases. The data is stored in two datasets, the data
to be learned stored in $\mu$ sets $D_{i}$ with $i\in\{1,\ldots,\mu\}$,
each corresponding to the data of the $i$-th previous match that
must be learned and $\mu$ other sets $D'_{i}$ with $i\in\{1,\ldots,\mu\}$,
the archived data, which correspond to the data already learned from
the $i$-th previous match. The set of data to learn for the new learning
phase is then samples of each $D_{i}$ of size $\left\lceil \frac{\delta}{\mu}\left(\left|D_{i}\right|+\left|D'_{i}\right|\right)\right\rceil $
(the total number of learned data is the number of generated data
times the duplication factor $\delta$). The selected samples are
removed from the $D_{i}$ and added to the $D'_{i}$ . The selected
data is then separated into minibatches, so as to preserve the stratification
as much as possible (the minibatches are thus in a way also stratified)
and learning is then performed on each minibatch. The formalization
of Stratified Experience Replay is described in Algorithm~\ref{alg:smooth-exp_replay}.

\begin{algorithm}[!bh]
\DontPrintSemicolon\SetAlgoNoEnd

\SetKwFunction{replay}{SGD\_with\_stratified\_experience\_replay}\SetKwProg{myproc}{Function}{}{}

\myproc{\replay{$\hadapt$, $D$, $\mu$, $\delta$, $B$}}{

append $\left(D,\emptyset\right)$ in $M$\;

\If{$\left|M\right|>\mu$}{

remove the oldest items of $M$ to have $\left|M\right|=\mu$

}

$n\leftarrow\sum\liste{\left\lceil \frac{\delta}{\mu}\left(\left|D\right|+\left|D'\right|\right)\right\rceil }{\left(D,D'\right)\in M}$

$g\leftarrow\max\left(1,\left\lfloor n/B\right\rfloor \right)$

\If{$\left|\left\lfloor n/g\right\rfloor -B\right|>\left|\left\lfloor n/\left(g+1\right)\right\rfloor -B\right|$}{

$g\leftarrow g+1$

}

$G\leftarrow\left\{ 1,\ldots,g\right\} $

\ForEach{$i\in G$}{

$S_{i}\leftarrow\emptyset$

}

\ForEach{$\left(D,D'\right)\in M$}{

$l\leftarrow\left\lceil \frac{\delta}{\mu}\left(\left|D\right|+\left|D'\right|\right)\right\rceil $

\ForEach{$c\in\left\{ 1,\ldots,l\right\} $}{

$\left(s,v\right)\leftarrow\mathrm{choice}\left(D\right)$

remove $\left(s,v\right)$ from $D$

add $\left(s,v\right)$ in $D'$

\If{$\left|D\right|=0$}{

$D\leftarrow D'$

$D'\leftarrow\emptyset$

}

$j\leftarrow\mathrm{choice}\left(G\right)$

$G\leftarrow G\backslash\{j\}$

\If{$\left|G\right|=0$}{

$G\leftarrow\left\{ 1,\ldots,g\right\} $

}

add $\left(s,v\right)$ in $S_{j}$

}\;

}\;

\ForEach{$i\in\left\{ 1,\ldots,g\right\} $}{

minimize $\sum_{(s,v)\in S_{i}}\left(\hadapt(s)-v\right)^{2}$ \;

}

}\;

\protect\protect

\caption{Stochastic gradient descent with Stratified experience replay algorithm
($B$ is the batch size, $\mu$ is the memory size, $\delta$ is the
duplication factor, $M$ is the memory buffer: a global variable initialized
by an empty queue ; choice($E$): uniformly randomly chooses an element
of $E$ ; see Table~\ref{tab:Index-of-symbols} for the definitions
of the other symbols)\label{alg:smooth-exp_replay}}
\end{algorithm}

\begin{rem}
We have not been able to evaluate the advantages or disadvantages
of this variant with respect to the basic experience replay: the performances,
we obtained during an experiment on this subject (not presented in
this document), are so close that we would need increase the number
of repetitions by 10 or maybe even 100 to be able to separate them,
which is currently complicated or even impossible. We are not claiming
that this algorithm is better than the classic replay experience,
nor vice versa.
\end{rem}

\subsection{Common Experiment Protocol\label{subsec:Common-Experience-Protocol}}

The experiments of several sections share the same protocol. It is
presented in this section. The protocol is used to compare different
variants of reinforcement learning algorithms. A variant corresponds
to a certain combination of elementary algorithms. More specifically,
a combination consists of the association of a search algorithm (iterative
deepening alpha-beta (with move ordering), MCTS (UCT with $c=0.4$
as exploration constant), $\ubfm$, ...), of an action selection method
($\epsilon$-greedy distribution (used by default), softmax distribution,
...), a terminal evaluation function $\hterminal$ (the classic game
gain (used by default), ...), and a procedure for selecting the data
to be learned (root learning, tree learning, or terminal learning).
The protocol consists in carrying out a reinforcement learning process
of $48$ hours for each variant (per repetition). At several stages
of the learning process, each combination is evaluated. This evaluation
consists in matches played by using the adaptive evaluation function
generated by the evaluated combination. Each variant is thus characterized
by a winning percentage at each stage of the reinforcement learning
process. More formally, we denote by $\hadaptnum h^{c}$ the evaluation
generated by the combination $c$ at the hour $h$. Each combination
is evaluated every hour by a winning percentage. The winning percentage
of a combination $c$ at a hour $h\leq48$ is computed from matches
by using the minimax search at depth $1$ with $\hadaptnum h^{c}$
as evaluation function confronting the following opponents: the minimax
search at depth $1$ with $\hadaptnum{48}^{c'}$ as evaluation for
all studied combination $c'$ (there is one match in first player
and another in second player per pair of combination). 

This protocol is repeated several times for each experiment in order
to reduce the statistical noise in the winning percentages obtained
for each variant (the obtained percentage is the average of the percentages
of repetitions). The winning percentages are then represented in a
graph showing the evolution of the winning percentages during training. 

In addition to the curve, the different variants are also compared
in relation to their final winning percentage, i.e. at the end of
the learning process. Unlike the evolution of winning percentages,
in the comparison of final performances, each evaluation $\hadaptnum{48}^{c}$
confronts each other evaluation $\hadaptnum{48}^{c'}$ from all repetitions.
In other words, this experiment consists in performing an all-play-all
tournament with all the evaluation functions generated during the
different repetitions. The presented winning percentage of a combination
is still the average over the repetitions. The matches are also made
by using minimax at depth $1$. These percentages are shown in tables.

\begin{rem}
The used version of MCTS does not performed random simulations for
evaluating the leaves. Instead, leaves are evaluated by a neural
network. No policies are used (unless it is explicitly specified).
\end{rem}

\begin{rem}
All the experiments based on this protocol involving MCTS were also
performed with $c=\sqrt{2}$ as exploration constant. The results
are similar.
\end{rem}

\subsubsection{Technical Details\label{subsec:Technical-details}}

We now present the technical details of this protocol.

The used parameters are: search time per action $\tau=2s$, batch
size $B=128$, memory size $\mu=10^{6}$, sampling rate $\sigma=4\%$
(see Section~\ref{subsec:Tree-Learning}). Moreover, the used adaptive
evaluation function for each combination is a convolutional neural
network \citep{krizhevsky2012imagenet} having three convolution layers\footnote{There is an exception: for the game Surkarta, there is only two convolution
layers.} followed by a fully connected hidden layer. For each convolutional
layer, the kernel size is $3\times3$ and the filter number is $64$.
The number of neurons in the fully connected layer is $100$. The
margin of each layer is zero. After each layer except the last one,
the ReLU activation function \citep{glorot2011deep} is used. The
output layer contains a neuron. When the classical terminal evaluation
is used, $\tanh$ is the output activation function. Otherwise, there
is no activation function for the output.
\begin{rem}
In this paper, filter numbers and numbers of neurons are chosen in
order to there are about the same number of variables in the convolution
layers and in the dense layers.
\end{rem}

\subsection{Comparison of Learning Data Selection Algorithms\label{subsec:Comparison-data-selection}}

We now compare tree learning, root learning and terminal learning,
using the protocol of Section~\ref{subsec:Common-Experience-Protocol}.
Each combination uses either tree learning, root learning, or terminal
learning. Moreover, each combination uses either iterative deepening
alpha-beta (denoted by $\id$) or MCTS. Furthermore, each combination
uses $\epsilon$-greedy as action selection method (see Section~\ref{subsec:Tree-Learning})
and the classical terminal evaluation ($1$ if the first player wins,
$-1$ if the first player loses, and $0$ in case of a draw). There
are a total of $6$ combinations. The experiment was repeated $32$
times. The winning percentage of a combination for each game and
for each evaluation step (i.e. each hour) is therefore calculated
from $384$ matches (recall that there is a first player match and
a second player match per pair of combinations). 

\begin{figure}
\begin{centering}
\includegraphics[scale=0.3]{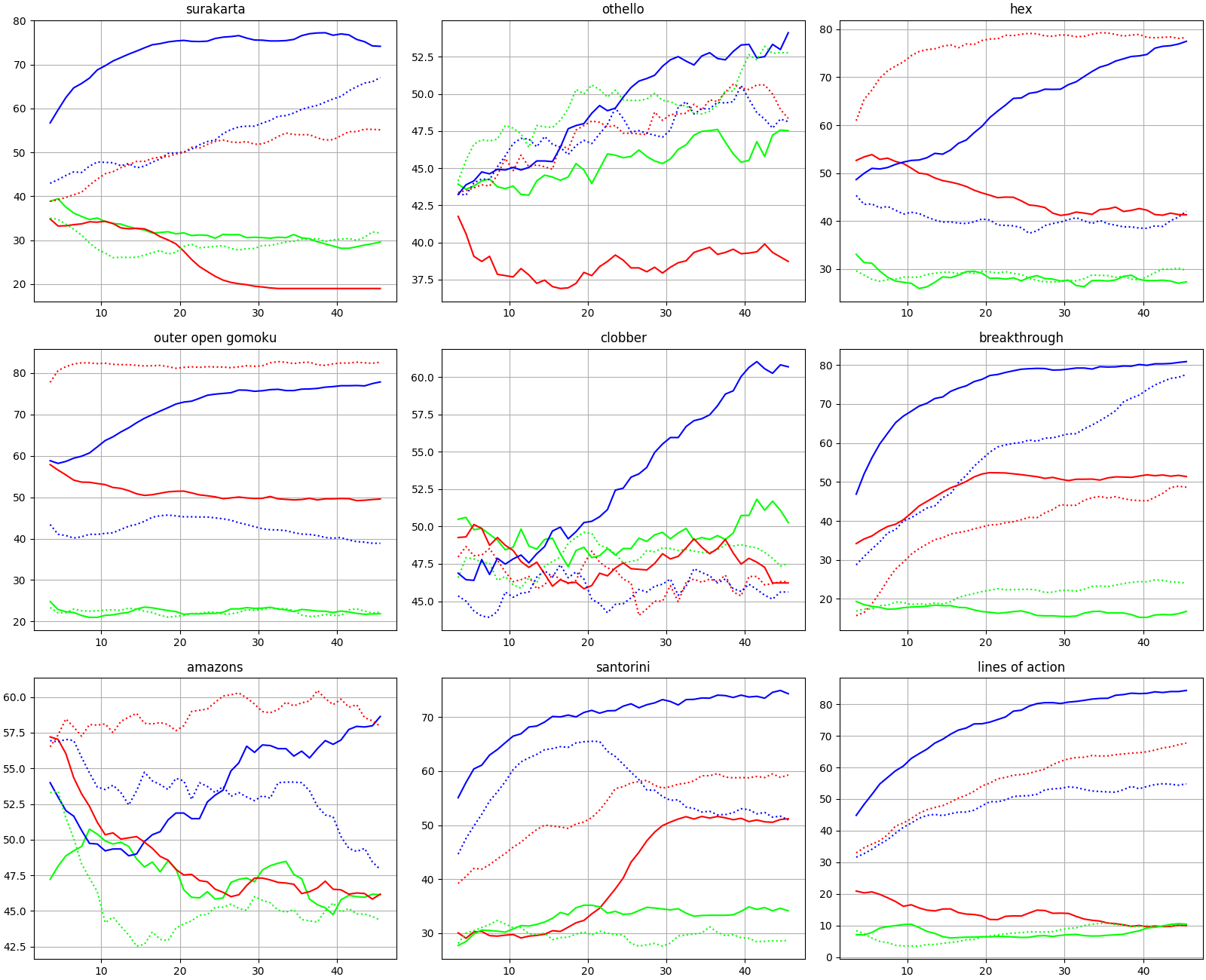}
\par\end{centering}
\caption{Evolutions of the winning percentages of the combinations of the experiment
of Section~\ref{subsec:Comparison-data-selection}, i.e. MCTS (dotted
line) or iterative deepening alpha-beta (continuous line) with tree
learning (blue line), root learning (red line), or terminal learning
(green line). The display uses a simple moving average of 6 data.\label{fig:comparison-data}}
\end{figure}

The winning percentage curves are shown in Figure~\ref{fig:comparison-data}.
The final winning percentages are shown in Table~\ref{tab:comparison-data}.
Each percentage of the table has required $12,288$ matches. $\id$
is first on all games except in Outer Open Gomoku where it is second
(MCTS root learning is first) and in Surakarta (MCTS with tree learning
is first). MCTS with root learning is better than MCTS with tree learning
except in Breakthrough and Surakarta. At Hex and Amazons, MCTS with
root learning gives better results throughout the learning process
but ends up being caught up by $\id$ with terminal learning. Terminal
learning performs worse everywhere, except in a few cases where it
is very slightly better. On average, $\id$ with tree learning is
better ($71\%$ win), then MCTS with root learning is second ($9\%$
lower win percentage), followed by MCTS with tree learning ($18\%$
lower to $\id$).

In conclusion, tree learning with $\id$ performs much better than
other combinations, although the results are very tight at Amazons,
Hex, and Outer Gomoku with MCTS with root learning.
 
\begin{table}
\begin{centering}
\begin{tabular}{|c|c|c|c|c|c|c|}
\hline 
 & \multicolumn{2}{c|}{tree learning} & \multicolumn{2}{c|}{root learning} & \multicolumn{2}{c|}{terminal learning}\tabularnewline
\hline 
 & MCTS & $\id$ & MCTS & $\id$ & MCTS & $\id$\tabularnewline
\hline 
\hline 
Othello & $48.8\%$  & $55.4\%$  & $51.0\%$  & $31.9\%$  & $52.9\%$  & $43.7\%$ \tabularnewline
\hline 
Hex & $45.1\%$  & $79.8\%$  & $79.8\%$  & $44.1\%$  & $29.8\%$  & $27.8\%$ \tabularnewline
\hline 
Clobber & $41.5\%$  & $62.5\%$  & $50.0\%$  & $45.5\%$  & $45.7\%$  & $49.8\%$ \tabularnewline
\hline 
Outer Open Gomoku & $40.3\%$  & $80.0\%$  & $87.3\%$  & $48.8\%$  & $21.6\%$  & $23.1\%$ \tabularnewline
\hline 
Amazons & $46.2\%$  & $58.8\%$  & $56.1\%$  & $44.5\%$  & $39.4\%$  & $46.0\%$ \tabularnewline
\hline 
Breakthrough & $78.1\%$  & $79.2\%$  & $45.5\%$  & $50.6\%$  & $22.3\%$  & $14.3\%$ \tabularnewline
\hline 
Santorini & $50.2\%$  & $73.8\%$  & $60.5\%$  & $51.3\%$  & $29.4\%$  & $36.5\%$ \tabularnewline
\hline 
Surakarta & $69.4\%$  & $65.2\%$  & $56.2\%$  & $20.8\%$  & $28.9\%$  & $23.6\%$ \tabularnewline
\hline 
Lines of Action & $58.8\%$  & $81.7\%$  & $67.7\%$  & $9.6\%$  & $6.9\%$  & $2.7\%$ \tabularnewline
\hline 
\textbf{mean} & $53.1\%$  & $70.7\%$  & $61.6\%$  & $38.5\%$  & $30.8\%$  & $29.7\%$ \tabularnewline
\hline 
\end{tabular}
\par\end{centering}
\caption{Final winning percentages of the combinations of the experiment of
Section~\ref{subsec:Comparison-data-selection} ($\protect\id$:
iterative deepening alpha-beta). Reminder: the percentage is the average
over the repetitions, of the winning percentage of a combination against
each other combination, in first and second player (see~\ref{subsec:Common-Experience-Protocol}
; $95\%$ confidence intervals: max $\pm0.85\%$). \label{tab:comparison-data}}
\end{table}

\section{Tree Search Algorithms for Game Learning\label{sec:Search-Algorithms-for-Learning}}

In this section, we introduce a new tree search algorithm, that we
call \emph{Descent Minimax} or more succinctly \emph{Descent}, dedicated
to be used during the learning process. After presenting Descent,
we compare it to MCTS with root learning and with tree learning, to
iterative deepening alpha-beta with root learning and with tree learning
and to $\ubfm$ with tree learning.

\subsection{Descent: Generate Better Data\label{subsec:Descente-:-g=00003D0000E9n=00003D0000E9rer}}

\begin{algorithm}[!bh]
\DontPrintSemicolon\SetAlgoNoEnd

\SetKwFunction{descenteiteration}{descent\_iteration}

\SetKwFunction{descente}{descent}\SetKwFunction{time}{time}\SetKwFunction{actions}{actions}\SetKwFunction{bestaction}{best\_action}\SetKwFunction{terminal}{terminal}\SetKwFunction{premier}{first\_player}
\SetKwProg{myproc}{Function}{}{}

\myproc{\descenteiteration{$s$, $S$, $T$, $\hadapt$, $\hterminal$}}{

\eIf{\terminal{$s$}}{

$S\leftarrow S\cup\{s\}$\;

$v(s)\leftarrow\hterminal(s)$

}{

\If{$s\notin S$}{

$S\leftarrow S\cup\{s\}$\;

\ForEach{$a\in$ \actions{$s$}}{

\eIf{\terminal{$a(s)$}}{

$v(s,a)\leftarrow$ \descenteiteration{$a(s),S,T,\hadapt,\hterminal$}\;

}{

$v(s,a)\leftarrow\hadapt\left(a(s)\right)$\;

}

}}

$a\leftarrow$ \bestaction{$s$}\;

$v(s,a)\leftarrow$ \descenteiteration{$a(s),S,T,\hadapt,\hterminal$}\;

$a\leftarrow$ \bestaction{$s$}\;

$v(s)\leftarrow v(s,a)$\;

}

return $v(s)$\;

}

\;

\myproc{\bestaction{$s$}}{

\eIf{\premier{$s$}}{return ${\displaystyle \argmax_{a\in\actions{s}}v\left(s,a\right)}$\;}{return
${\displaystyle \argmin_{a\in\actions{s}}v\left(s,a\right)}$\;}

}

\;

\myproc{\descente{$s$, $S$, $T$, $\hadapt$, $\hterminal$,
$\tau$}}{

$t=$ \time{}\;

\lWhile{\time{}$-\,t<\tau$}{\descenteiteration{$s$, $S$,
$T$, $\hadapt$, $\hterminal$}}

return $S$, $T$\;

}\;

\protect\protect

\caption{Descent minimax algorithm (see Table~\ref{tab:Index-of-symbols}
for the definitions of symbols ; note: $S$ is the set of states which
are non-leaves or terminal and $T=(v,)$).\label{alg:descente}}
\end{algorithm}

Thus, we present \emph{Descent.} It is a modification of $\ubfm$
which builds a different, deeper, game tree, to be combined with tree
learning. 
\begin{rem}
Note that, combining it with root learning or terminal learning is
of no interest. The reason is that Descent spend most of its time
exploring a part of the game tree whose only purpose is to provide
states for learning. With root or terminal learning, these states
are not learned and therefore generated for nothing. 
\end{rem}

The idea of Descent is to combine $\ubfm$ with deterministic end-game
simulations providing interesting values from the point of view of
learning. 

The \emph{Descent} algorithm (Algorithm~\ref{alg:descente}) recursively
selects the best child of the current node, which becomes the new
current node. It performs this recursion from the root (the current
state of the game) until reaching a terminal node (an end game). It
then updates the value of the selected nodes (minimax value). The
Descent algorithm repeats this recursive operation starting from the
root as long as there is some search time left. \emph{Descent} is
almost identical to $\ubfm$. The only difference is that Descent
performs an iteration until reaching a terminal state while $\ubfm$
performs this iteration until reaching a leaf of the tree ($\ubfm$
stops the iteration much earlier). In other words, during an iteration,
$\ubfm$ just extends one of the leaves of the game tree while Descent
recursively extends the best child from this leaf until reaching the
end of the game. 

The Descent algorithm has the advantage of $\ubfm$, i.e. to perform
a longer search to determine a better action to play. By learning
the values of the game tree (by using for example tree learning),
it also has the advantage of a minimax search at depth $1$, i.e.
to raise the values of the terminal nodes to the other nodes more
quickly. In addition, the states thus generated are closer to the
terminal states. Their values are therefore better approximations.
\begin{rem}
In the experiments of this article, when there is a value tie in best\_action(s),
the tie is broken at random. An alternative could be to choose the
subtree whose principal variation is the least deep.
\end{rem}

\subsection{Comparison of Search Algorithms for Game Learning\label{subsec:Comparison-of-algorithms-for-Learning}}

We now compare Descent with tree learning to MCTS with root learning
and with tree learning, to iterative deepening alpha-beta with root
learning and with tree learning, and to $\ubfm$ with tree learning,
using the protocol of Section~\ref{subsec:Common-Experience-Protocol}.
There are a total of $6$ combinations. The experiment was repeated
$32$ times. The winning percentage of a combination for each game
and for each evaluation step (i.e. each hour) is therefore calculated
from $384$ matches (recall that there is a first player match and
a second player match per pair of combinations). 

\begin{figure}
\begin{centering}
\includegraphics[scale=0.3]{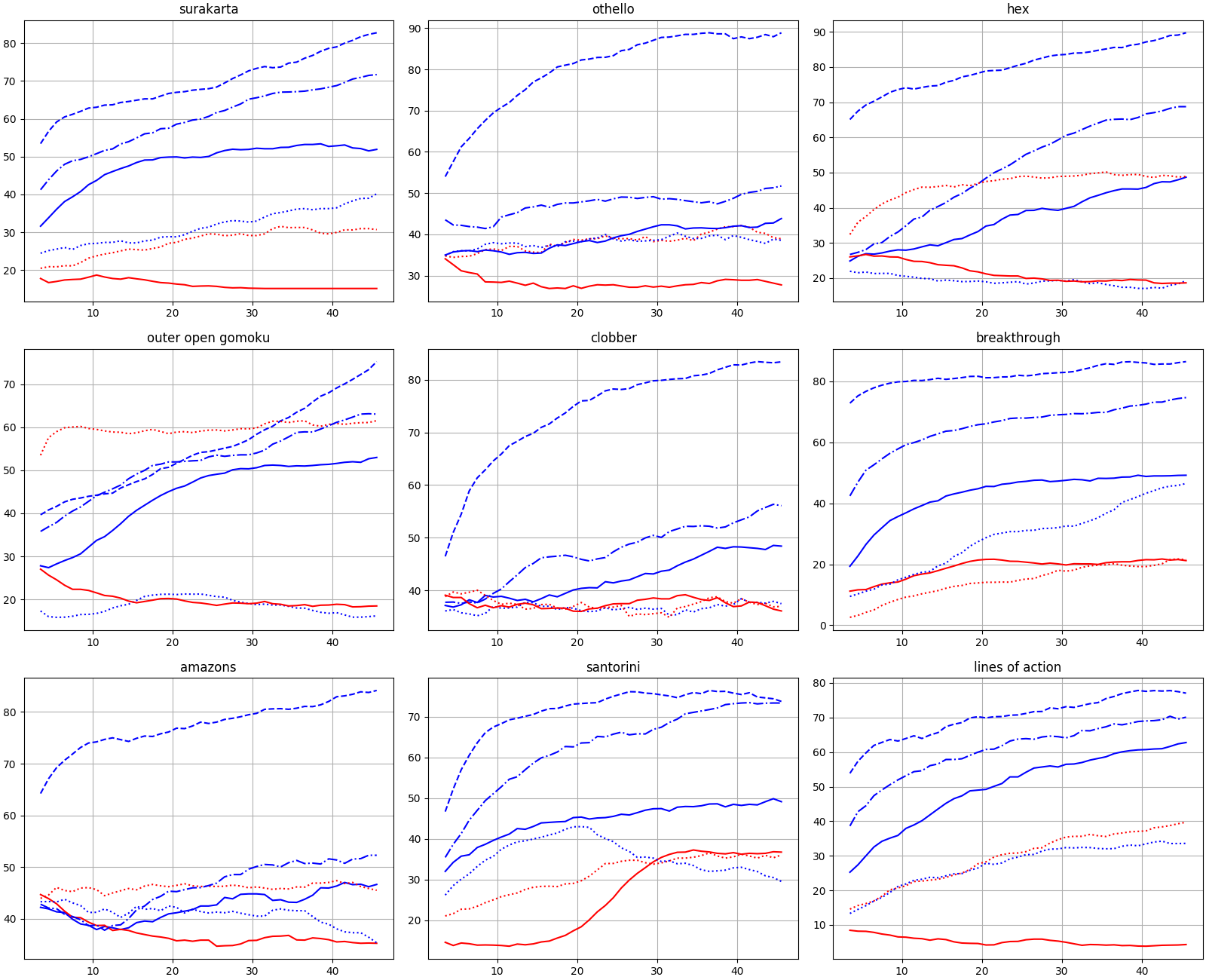}
\par\end{centering}
\caption{Evolutions of the winning percentages of the combinations of the experiment
of Section~\ref{subsec:Comparison-of-algorithms-for-Learning}, i.e.
of Descent (dashed line), $\protect\ubfm$ (dotted dashed line), MCTS
(dotted line), and iterative deepening alpha-beta (continuous line)
with tree learning (blue line) or root learning (red line). The display
uses a simple moving average of 6 data.\label{fig:comparison-search}}
\end{figure}

The winning percentage curves are shown in Figure~\ref{fig:comparison-search}.
The final winning percentages are shown in Table~\ref{tab:comparison-search}.
Each percentage of the table has required $12,288$ matches. It is
Descent which gets the best curves on all games. For two games (Surakarta
and Outer Open Gomoku), the difference with $\ubfm$ is very narrow
but the results remain better than the classic approaches (MCTS and
alpha-beta). On each game, Descent obtains a final percentage higher
than all the other combinations (except in Santorini where it is $2\%$
lower than $\ubfm$, the best algorithm at this game). On average
over all games, Descent has $82\%$ win and is above $\ubfm$, the
second best combination, by $18\%$. It is also above $\id$ with
tree learning, the third best combination, by $34\%$.
\begin{table}
\begin{centering}
\begin{tabular}{|c|c|c|c|c|c|c|}
\hline 
 & \multicolumn{4}{c|}{tree learning} & \multicolumn{2}{c|}{root learning}\tabularnewline
\hline 
 & Descent & $\ubfm$ & MCTS & $\id$ & MCTS & $\id$\tabularnewline
\hline 
\hline 
Othello & $89.4\%$  & $47.2\%$  & $37.7\%$  & $42.7\%$  & $44.9\%$  & $22.1\%$ \tabularnewline
\hline 
Hex & $94.9\%$  & $71.7\%$  & $20.5\%$  & $50.7\%$  & $50.2\%$  & $20.5\%$ \tabularnewline
\hline 
Clobber & $83.0\%$  & $56.7\%$  & $32.9\%$  & $48.9\%$  & $42.0\%$  & $35.5\%$ \tabularnewline
\hline 
Outer Open Gomoku & $77.6\%$  & $63.9\%$  & $18.0\%$  & $51.6\%$  & $64.7\%$  & $18.2\%$ \tabularnewline
\hline 
Amazons & $84.3\%$  & $55.3\%$  & $32.5\%$  & $46.3\%$  & $43.7\%$  & $31.7\%$ \tabularnewline
\hline 
Breakthrough & $86.5\%$  & $72.5\%$  & $45.5\%$  & $47.6\%$  & $19.2\%$  & $21.0\%$ \tabularnewline
\hline 
Santorini & $69.9\%$  & $71.8\%$  & $31.9\%$  & $47.6\%$  & $37.7\%$  & $40.2\%$ \tabularnewline
\hline 
Surakarta & $82.8\%$  & $69.7\%$  & $41.2\%$  & $42.1\%$  & $29.4\%$  & $14.5\%$ \tabularnewline
\hline 
Lines of Action & $73.4\%$  & $66.9\%$  & $36.1\%$  & $57.4\%$  & $39.4\%$  & $3.7\%$ \tabularnewline
\hline 
\textbf{mean} & $82.4\%$  & $64.0\%$  & $32.9\%$  & $48.3\%$  & $41.2\%$  & $23.1\%$ \tabularnewline
\hline 
\end{tabular}
\par\end{centering}
\caption{Final winning percentages of the combinations of the experiment of
Section~\ref{subsec:Comparison-of-algorithms-for-Learning} ($\protect\id$:
iterative deepening alpha-beta ; see~\ref{subsec:Common-Experience-Protocol}
; $95\%$ confidence intervals: max $\pm0.85\%$)\label{tab:comparison-search}}
\end{table}

In conclusion, Descent is the best search algorithm for learning
evaluation functions. $\ubfm$ (with tree learning) is the second
best algorithm, sometimes very close to Descent performances and sometimes
very far, but always superior to the other algorithms (slightly or
largely depending on the game).

\section{Completion\label{subsec:Completion}}

In this section, we propose a complementary technique, that we call
\emph{completion}, which corrects state evaluation functions taking
into account the resolution of states. 
\begin{rem}
However, we do not evaluate this technique in this paper, to avoid
making this paper unnecessarily cumbersome and to focus on the learning
aspects that constitute the main contributions of this paper. The
evaluation of this technique is however available in \citep{cohen2025little}
which studies the impact of improvements of Unbounded Minimax. This
study shows that completion improves the performance of Unbounded
Minimax by $6\%\pm0.7\%$. 
\end{rem}

\subsection{Completion Concept}

\begin{figure}
\begin{centering}
\includegraphics[scale=0.28]{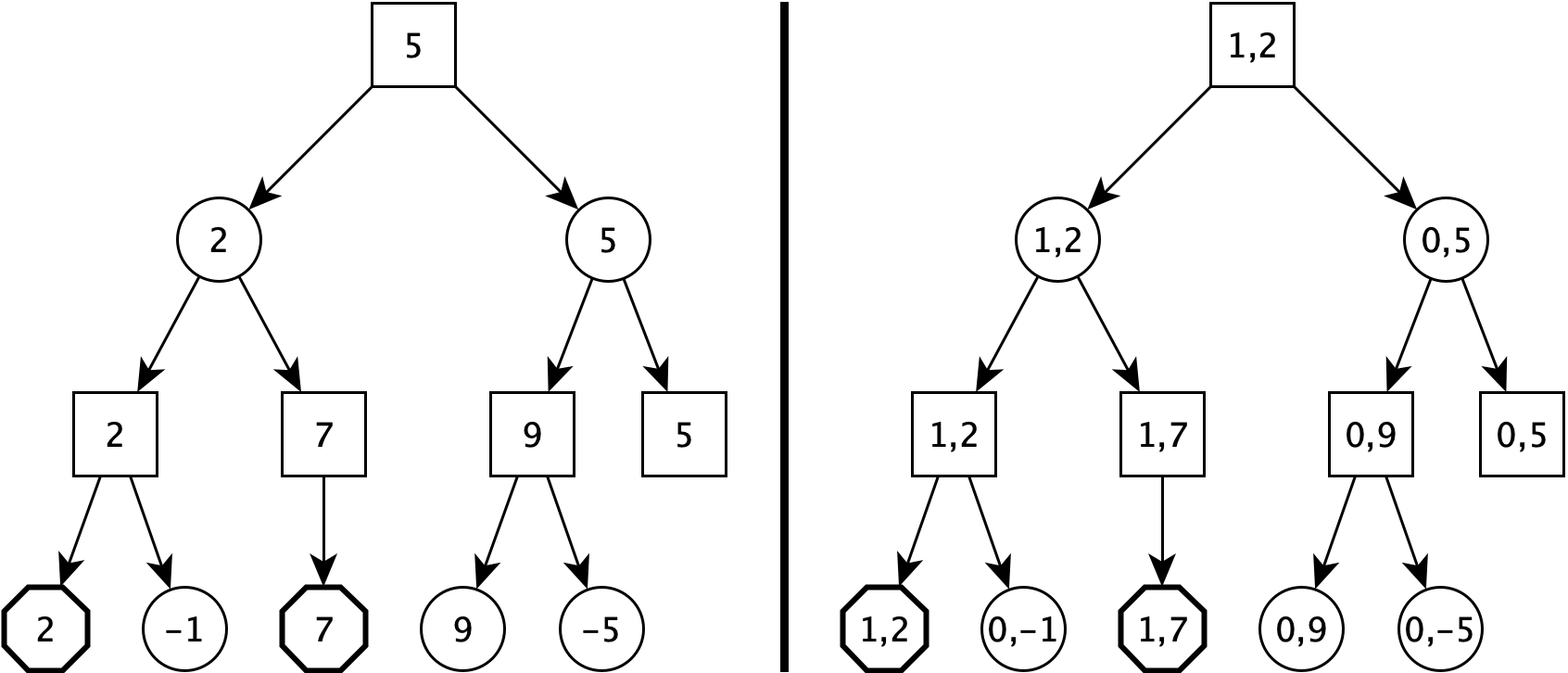}
\par\end{centering}
\caption{The left graph is a game tree where maximizing does not lead to the
best decision ; the right graph is the left game tree with completion
(nodes are labeled by a pair of values) and thus maximizing leads
to the best decision (square node: first player node (max node), circle
node: second player node (min node), octagon: terminal node)\label{fig:completion}.}
\end{figure}

Relying solely on the value of states calculated from the terminal
evaluation function \emph{and} the adaptive evaluation function can
sometimes lead to certain aberrant behaviors. More precisely, if we
only seek to maximize the value of states, we will then choose to
play a state $s$ rather than another state $s'$ when $s$ is of
greater value than $s'$ even if $s'$ is a winning resolved state
(a state is \emph{resolved} if we know the result of the match starting
from that state in which the two players play optimally). A search
algorithm can resolve a state. This happens when all the leaves of
the subtree starting from that state are terminal. Choosing $s$ rather
than $s'$, a winning resolved state, is an error\footnote{There is perhaps, in certain circumstances, an interest in making
this error from the point of view of learning.} when $s$ is not resolved (or when $s$ is resolved and is not winning).
By choosing $s$, guarantee of winning is lost. The left graph of
Figure~\ref{fig:completion} illustrates such a scenario. 
It is therefore necessary to take into account both the value of states
and the resolution of states. 

\subsection{Completed Algorithms}

The completion technique, which we propose in this section, is one
way of doing it. It consists, on the one hand, in associating with
each state $s$ a completion value $c(s)$ and a resolution value
$r(s)$. The completion value $c(s)$ of a leaf state $s$ is $0$
if the state $s$ is not terminal or if it is a draw, $1$ if it
is a winning terminal state, and $-1$ if the state is a losing terminal
state. The value $c(s)$ of a non-leaf state $s$ is computed as the
minimax value of the subtree of the partial game tree starting from
$s$ where the leaves are evaluated by their completion value. The
resolution value $r(s)$ of a leaf state $s$ is $0$ if the state
$s$ is not terminal and $1$ if it is terminal. The resolution value
$r(s)$ of a non-leaf state $s$ is $1$ if $\left|c\left(s\right)\right|=1$.
Otherwise, $r(s)$ is the minimum of the resolution values of the
childen of $s$. 

The completion technique consists, on the other hand, in using $c\left(\cdot\right)$
to compute $v\left(\cdot\right)$. For this, states are compared from
pairs $\left(c\left(\cdot\right),v\left(\cdot\right)\right)$, by
using the lexicographic order (instead of just compare states from
$v\left(\cdot\right)$). More precisely, the value $v\left(s\right)$
of a state $s$ is computed in calculating $\left(c(s),v(s)\right)$
as the minimax value of the subtree of the partial game tree starting
from $s$ where the leaves $l$ are evaluated by $\left(c(l),v(l)\right)$.
Thus, the value $v\left(s\right)$ of a winning state $s$ is always
the value of the corresponding winning terminal leaf. The right graph
of Figure~\ref{fig:completion} illustrates the use of completion. 

Finally, with the completion technique, to decide which action to
play during the search, we choose the best unresolved action if it
exists and otherwise the best resolved action (i.e. we choose the
action which maximize $\left(-r\left(s\right),c\left(s\right),v\left(s\right)\right)$
in a max state and which minimize $\left(r\left(s\right),c\left(s\right),v\left(s\right)\right)$
in a min state). 

Without completion, algorithms such as Descent or Unbounded Minimax
can get stuck in a non-optimal fixed point. The completion technique
ensures that with sufficient thinking time, the optimal strategy is
found~\citep{cohen2021completeness}. 

Moreover, the use of the resolution of states also makes it possible
to stop the search in the resolved subtrees and thus to save computing
time. Descent algorithm modified to use the completion and the resolution
stop is described in Algorithm~\ref{alg:descente-completion}. With
completion, after the search, Unbounded Minimax always chooses an
action leading to a winning resolved state and never chooses, if possible,
an action leading to a losing resolved state. Unbounded Minimax algorithm
modified to use the completion and the resolution stop is described
in Algorithm~\ref{alg:ubfms-decision-1}.

\begin{algorithm}[!bh]
\DontPrintSemicolon\SetAlgoNoEnd

\SetKwFunction{completiondescenteiteration}{descent\_iteration}

\SetKwFunction{completiondescente}{completed\_descent}\SetKwFunction{time}{time}\SetKwFunction{actions}{actions}\SetKwFunction{completedbestaction}{completed\_best\_action}\SetKwFunction{completedbestactiondual}{completed\_best\_action\_dual}\SetKwFunction{terminal}{terminal}\SetKwFunction{premier}{first\_player}
\SetKwFunction{backupresolution}{backup\_resolution} \SetKwProg{myproc}{Function}{}{}

\myproc{\completiondescenteiteration{$s$, $S$, $T$, $\hadapt$,
$\hterminal$}}{

\eIf{\terminal{$s$}}{

$S\leftarrow S\cup\{s\}$\;

$r\left(s\right),c\left(s\right),v\left(s\right)\leftarrow1,\hfb(s),\hterminal(s)$

}{

\If{$s\notin S$}{

$S\leftarrow S\cup\{s\}$\;

\ForEach{$a\in$ \actions{$s$}}{

\eIf{\terminal{$a\left(s\right)$}}{

$r(s,a),c(s,a),v(s,a)\leftarrow$ \completiondescenteiteration{$a(s),S,T,\hadapt,\hterminal$}\;\;

}{

$v(s,a)\leftarrow\hadapt\left(a(s)\right)$\;

}

}

$a_{b}\leftarrow$ \completedbestaction{$s$, $\actions{s}$}\;

$c(s),v(s)\leftarrow c\left(s,a_{b}\right),v\left(s,a_{b}\right)$\;

$r\left(s\right)\leftarrow$ \backupresolution{$s$}\;

}

\If{$r\left(s\right)=0$}{

$A\leftarrow\liste{a\in\actions{s}}{r\left(s,a\right)=0}$

$a\leftarrow$ \completedbestactiondual{$s$, $A$}\;

$n(s,a)\leftarrow n(s,a)+1$\;

$r\left(s,a\right),c\left(s,a\right),v\left(s,a\right)\leftarrow$
\completiondescenteiteration{$a(s),S,T,\hadapt,\hterminal$}\;

$a\leftarrow$ \completedbestaction{$s$, $\actions{s}$}\;

$c(s),v(s)\leftarrow c\left(s,a\right),v\left(s,a\right)$\;

$r\left(s\right)\leftarrow$ \backupresolution{$s$}\;

}

}

return $r\left(s\right),c\left(s\right),v\left(s\right)$\;

}

\;

\myproc{\completiondescente{$s$, $S$, $T$, $\hadapt$, $\hterminal$,
$\tau$}}{

$t=$ \time{}\;

\lWhile{\time{}$-\,t<\tau$ $\wedge$ $r\left(s\right)=0$}{\completiondescenteiteration{$s$,
$S$, $T$, $\hadapt$, $\hterminal$}}

return $S$, $T$\;

}\;

\protect\protect

\caption{\emph{Descent} tree search algorithm with completion and resolution
stop (see Table~\ref{tab:Index-of-symbols} for the definitions of
symbols and Algorithm~\ref{alg:descente-completion-suite} for the
definitions of completed\_best\_action($s$) and backup\_resolution($s$)).
Note: $T=(v,c,r)$ and $S$ is the set of states of the partial game
tree which are non-leaves or terminal.\label{alg:descente-completion}}
\end{algorithm}
\begin{algorithm}[!bh]
\DontPrintSemicolon\SetAlgoNoEnd

\SetKwFunction{completiondescenteiteration}{completed\_descent\_iteration}

\SetKwFunction{completiondescente}{completed\_descent}\SetKwFunction{time}{time}\SetKwFunction{actions}{actions}\SetKwFunction{completedbestaction}{completed\_best\_action}\SetKwFunction{completedbestactiondual}{completed\_best\_action\_dual}\SetKwFunction{terminal}{terminal}\SetKwFunction{premier}{first\_player}
\SetKwFunction{backupresolution}{backup\_resolution} \SetKwProg{myproc}{Function}{}{}

\myproc{\completedbestaction{$s$, $A$}}{

\eIf{\premier{$s$}}{return ${\displaystyle \argmax_{a\in A}\left(c\left(s,a\right),v\left(s,a\right),n\left(s,s'\right)\right)}$\;}{return
${\displaystyle \argmin_{a\in A}\left(c\left(s,a\right),v\left(s,a\right),-n\left(s,s'\right)\right)}$\;}

}

\;

\myproc{\completedbestactiondual{$s$, $A$}}{

\eIf{\premier{$s$}}{return ${\displaystyle \argmax_{a\in A}\left(c\left(s,a\right),v\left(s,a\right),-n\left(s,s'\right)\right)}$\;}{return
${\displaystyle \argmin_{a\in A}\left(c\left(s,a\right),v\left(s,a\right),n\left(s,s'\right)\right)}$\;}

}

\;

\myproc{\backupresolution{$s$}}{

\eIf{$\left|c\left(s\right)\right|=1$}{return $1$ \;}{return
$\minimum{}_{a\in\actions{s}}r\left(s,a\right)$\;}

}

\;

\protect\protect

\caption{Definition of the algorithms completed\_best\_action($s$, $A$),
which computes the \emph{a priori }best action by using completion,
and backup\_resolution($s$), which updates the resolution of $s$
from its child states. \label{alg:descente-completion-suite}}
\end{algorithm}

We also propose to use the resolution of states with action selections,
to reduce the duration of games and therefore \emph{a priori} the
duration of the learning process: always play an action leading to
a winning resolved state if it exists and never play an action leading
to a losing resolved state if possible. Thus, if among the available
actions we know that one of the actions is winning, we play it. If
there is none, we play according to the chosen action selection method
among the actions not leading to a losing resolved state (if possible).
 We call it \emph{completed action selection}.
\begin{rem}
It is not clear, however, that \emph{completed action selection} improves,
in practice, performance as it prunes a portion of the game tree whose
values could be useful for learning (but it would be surprising if
this were not the case).
\end{rem}

\begin{algorithm}[!bh]
\DontPrintSemicolon\SetAlgoNoEnd

\SetKwFunction{completionubfmsiteration}{ubfms\_iteration}

\SetKwFunction{completionubfms}{ubfms\_tree\_search}\SetKwFunction{time}{time}\SetKwFunction{actions}{actions}\SetKwFunction{completedbestaction}{completed\_best\_action}\SetKwFunction{terminal}{terminal}\SetKwFunction{premier}{first\_player}
\SetKwProg{myproc}{Function}{}{}

\myproc{\completionubfmsiteration{$s$, $S$, $T$, $\hadapt$,
$\hterminal$}}{

\eIf{\terminal{$s$}}{

$S\leftarrow S\cup\{s\}$\;

$r(s),c(s),v(s)\leftarrow1,\hfb(s),\hterminal(s)$

}{

\If{$r\left(s\right)=0$}{

\eIf{$s\notin S$}{

$S\leftarrow S\cup\{s\}$\;

\ForEach{$a\in$ \actions{$s$}}{

\eIf{\terminal{$a\left(s\right)$}}{

$r(s,a),c(s,a),v(s,a)\leftarrow$ \completionubfmsiteration{$a(s),S,T,\hadapt,\hterminal$}\;\;

}{

$v(s,a)\leftarrow\hadapt\left(a(s)\right)$\;

}

}

}{

$A\leftarrow\liste{a\in\actions{s}}{r\left(s,a\right)=0}$

$a\leftarrow$ \completedbestactiondual{$s$, $A$}\;

$n(s,a)\leftarrow n(s,a)+1$\;

$r\left(s,a\right),c\left(s,a\right),v\left(s,a\right)\leftarrow$
\completionubfmsiteration{$a(s),S,T,\hadapt,\hterminal$}}\;

$a\leftarrow$ \completedbestaction{$s$, $\actions{s}$}\;

$c(s),v(s)\leftarrow c\left(s,a\right),v\left(s,a\right)$\;

$r\left(s\right)\leftarrow$ \backupresolution{$s$}\;

}

}

return $r(s),c(s),v(s)$\;

}

\;

\myproc{\completionubfms{$s$, $S$, $T$, $\hadapt$, $\hterminal$,
$\tau$}}{

$t=$ \time{}\;

\lWhile{\time{}$-\,t<\tau\wedge r\left(s\right)=0$}{\completionubfmsiteration{$s$,
$S$, $T$, $\hadapt$, $\hterminal$}}

return $S$, $T$\;

}\;

\protect\protect

\caption{\emph{$\protect\ubfmt$} tree search algorithm with completion and
resolution stop (see Table~\ref{tab:Index-of-symbols} for the definitions
of symbols and Algorithm~\ref{alg:descente-completion-suite} for
the definitions of completed\_best\_action($s$) and backup\_resolution($s$)).
Note: $T=(v,c,r,n)$. Note: Adding terminal states in $S$ is only
useful during training with tree learning, so it should not be done
during confrontations.\label{alg:ubfms_search-completed}}
\end{algorithm}
\begin{algorithm}[!bh]
\DontPrintSemicolon\SetAlgoNoEnd

\SetKwFunction{completionubfmsiteration}{ubfms\_iteration}

\SetKwFunction{completiontreeubfms}{ubfms\_tree\_search}\SetKwFunction{completionubfms}{ubfms}\SetKwFunction{time}{time}\SetKwFunction{actions}{actions}\SetKwFunction{completedsafestaction}{safest\_action}\SetKwFunction{terminal}{terminal}\SetKwFunction{premier}{first\_player}
\SetKwProg{myproc}{Function}{}{}

\myproc{\completedsafestaction{$s$, $T$}}{

\eIf{\premier{$s$}}{return ${\displaystyle \argmax_{a\in\actions{s}}\left(c(s,a),n(s,a),v\left(s,a\right)\right)}$\;}{return
${\displaystyle \argmin_{a\in\actions{s}}\left(c(s,a),-n(s,a),v\left(s,a\right)\right)}$\;}

}

\;

\myproc{\completionubfms{$s$, $S$, $T$, $\hadapt$, $\hterminal$,
$\tau$}}{

$S,T\leftarrow$\completiontreeubfms{$s$, $S$, $T$, $\hadapt$,
$\hterminal$,$\tau$}\;

return \completedsafestaction{$s$, $T$}

}\;

\protect\protect

\caption{\emph{$\protect\ubfmt$} action decision algorithm with completion
(see Algorithm~\ref{alg:ubfms_search-completed} for the definition
of the method ubfms\_tree\_search() ; see Table~\ref{tab:Index-of-symbols}
for the definitions of symbols). Note: $T=(v,c,r,n)$ and $S$ is
the set of states of the game tree which are non-leaves or terminal.\label{alg:ubfms-decision-1}}
\end{algorithm}

\begin{rem}
In some application cases, we will prefer to ensure the draw rather
than trying to win. Algorithm~\ref{alg:ubfms-decision-1} must then
be adapted to decide the action to play. If the first player prefers
to guarantee draws, it must instead maximize:
\[
\left(c(s,a),r\left(s,a\right),n(s,a),v\left(s,a\right)\right)\text{.}
\]
If the second player prefers to guarantee draws, it must instead minimize:
\[
\left(c(s,a),-r\left(s,a\right),n(s,a),v\left(s,a\right)\right)\text{.}
\]
\end{rem}

\begin{rem}
For a game where there is no draw, the computation of the resolution
value is not necessary (all necessary information is in the completion
value).
\end{rem}

\begin{rem}
For games without draw, the completion of Unbounded Minimax can be
simply implemented by giving a value of $+\infty$ to winning terminal
states and a value of $-\infty$ to losing terminal states (that is
to say, it is sufficient to replace the terminal function by this
infinite terminal function).

However, in the additional special case of adaptive evaluation functions
with values in $]-1,1[,$ it is not sufficient that the terminal evaluation
functions are in $\left\{ -1,1\right\} $. Indeed, in practice, with
rounding errors, the adaptive evaluation function being real value
will have values in $[-1,1]$.
\end{rem}

\section{Reinforcement Heuristic to Improve Learning Performance\label{sec:Reinforcement-Heuristic}}

In this section, we propose the technique of \emph{reinforcement heuristic},
which consists to replace the classical terminal evaluation function
-- that we denote by $\hfb$, which returns $1$ if the first player
wins, $-1$ if the second player wins, and $0$ in case of a draw
\citep{young2016neurohex,silver2017mastering,gao2018three} -- by
another heuristic to evaluate terminal states during the learning
process. By using the reinforcement heuristic technique, non-terminal
states are therefore evaluated differently, partial game trees and
thus matches during the learning process are different, which can
impact the learning performances. We start by defining what we call
a reinforcement heuristic and we offer several reinforcement heuristics.
Finally, we compare the reinforcement heuristics that we propose to
the classical terminal evaluation function. 

\subsection{Reinforcement Heuristic Definition}

A reinforcement heuristic is a terminal evaluation function that is
more expressive than the classical terminal function, i.e. the game
gain.
\begin{defn}
Let $\hfb$ the game gain function of a game (i.e. $\hfb$ returns
$1$ if the first player wins, $-1$ if the second player wins, and
$0$ in case of a draw). 

A reinforcement heuristic $h_{r}$ is a function that preserves the
order of the game gain function: for any two terminal states of the
game $s,s'$, $\hfb\left(s\right)<\hfb\left(s'\right)$ implies $h_{r}\left(s\right)<h_{r}\left(s'\right)$.
\end{defn}

\subsection{Some Reinforcement Heuristics}

In the following subsections, we propose different reinforcement heuristics.

\subsubsection{Scoring}

Some games have a natural reinforcement heuristic: the game score.
For example, in the case of the game Othello (and in the case of the
game Surakarta), the game score is the number of its pieces minus
the number of pieces of his opponent (the goal of the game is to have
more pieces than its opponent at the end of the game). The scoring
heuristic used as a reinforcement heuristic consists of evaluating
the terminal states by the final score of the game. With that reinforcement
heuristic, the adaptive evaluation function will seek to learn the
score of states. In the context of an algorithm based on minimax,
the score of a non-terminal state is the minimax value of the subtree
starting from that state whose terminal leaves are evaluated by their
scores. After training, the adaptive evaluation function then contains
more information than just an approximation of the result of the game,
it contains an approximation of the score of the game. If the game
score is \emph{intuitive}, the scoring heuristic should improve learning
performances. 
\begin{rem}
In the context of the game of the Amazons, the score is the size of
the territory of the winning player, i.e. the squares which can be
reached by a piece of the winning player. It is approximately the
number of empty squares.
\end{rem}

\subsubsection{Additive and Multiplicative Depth Heuristics\label{subsec:Depth-Heuristic}}

Now we offer the following reinforcement heuristic: the \emph{depth
heuristic}. It consists in giving a better value to the winning states
close to the start of the game than to the winning states far from
the start. Reinforcement learning with the depth heuristic is learning
the duration of matches in addition to their results. This learned
information is then used to try to win as quickly as possible and
try to lose as late as possible. If learning durations can be done,
this should reduce the length of games and therefore speed up learning
(by increasing the number of matches played in the same time but also
by reducing the propagation time of end-of-game information to the
start-of-game states). Seeking to lose for as long as possible helps
avoid missing out on potential long-term wins (even if it makes games
last longer).

We propose two realizations of the depth heuristic: the \emph{additive
depth heuristic}, that we denote by $\hfp$, and the\emph{ multiplicative
depth heuristic}, that we denote by $\hfp'$. The evaluation function
$\hfp$ returns the value $l$ if the first player wins, the value
$-l$ if the second player wins, and $0$ in case of a draw, with
$l=P-p+1$ where $P$ is the maximum number of playable actions in
a game and $p$ is the number of actions played since the beginning
of the game. For the game of Hex, $l$ is the number of empty cells
on the board plus $1$. For the games where $P$ is very large or
difficult to compute, we can instead use $l=\max\left(1,\tilde{P}-p\right)$
with $\tilde{P}$ a constant approximating $P$ (close to the empirical
maximum length of matches). The evaluation function $\hfp'$ is identical
except that $l$ satisfies $l=\frac{P'}{p}$, and $P'$ can be the
exact or empirical average length of matches.
\begin{rem}
Note that the idea of fast victory and slow defeat has already been
proposed but not used in a learning process~\citep{CazenaveSST16}.
\end{rem}

\subsubsection{Cummulative Mobility\label{subsec:Cummulative-Mobility}}

The next reinforcement heuristic that we propose is \emph{cummulative
mobility}. It consists in favoring the matchs where the player has
more possibility of action and where his opponent has less. The implementation
used in this article is as following. The value of a terminal state
is $\frac{M_{1}}{M_{2}}$ if the first player wins, $-\frac{M_{2}}{M_{1}}$
if the second player wins, and $0$ in case of a draw, where $M_{1}$
is the mean of the number of available actions in each turn of the
first player since the start of the game and $M_{2}$ is the mean
of the number of available actions in each turn of the second player
since the start of the game.

\subsubsection{Piece Counting: Presence}

Finally, we propose as reinforcement heuristic: the \emph{presence}
heuristic. It consists in taking into account the number of pieces
of each player. This heuristic is based on the principle that the
more a player has pieces the more this one has an advantage. There
are several implementations for the presence heuristic. We use in
this article the following implementation: the heuristic value is
$\max(n_{1}-n_{2},1)$ if the first player wins, $\min(n_{1}-n_{2},-1)$
if the second player wins, and $0$ in case of a draw, where $n_{1}$
is the number of pieces of the first player and $n_{2}$ is the number
of pieces of the second player. Note that in the games Surakarta
and Othello, the score corresponds to a presence heuristic.

\subsection{Comparison of Reinforcement Heuristics\label{subsec:Comparison-of-Reinforcement}}

We now compare the different heuristics that we have proposed to the
classical terminal evaluation function $\hfb$ on different games,
using the protocol of Section~\ref{subsec:Common-Experience-Protocol}.
Each combination uses Descent with completion (Algorithm~\ref{alg:descente-completion}),
completed $\epsilon$-greedy (see Algorithm~\ref{alg:e-greedy},
and Section~\ref{subsec:Completion}). Each combination uses a different
terminal evaluation function. These terminal evaluations are the classical
game gain evaluation function $\hfb$, the additive depth heuristic,
the multiplicative depth heuristic, the scoring heuristic, the cummulative
mobility, and the presence heuristic. Other parameters are the same
as Section~\ref{subsec:Comparison-data-selection}. There are, at
most, a total of $6$ combinations per game (on some games, some
heuristics are not evaluated because they are trivially of no interest
or equivalent to another heuristic). The experiment was repeated $48$
times. The winning percentage of a combination for each game and
for each evaluation step (i.e. each hour) is therefore calculated
from $288$ to $576$ matches. 

\begin{figure}
\begin{centering}
\includegraphics[scale=0.3]{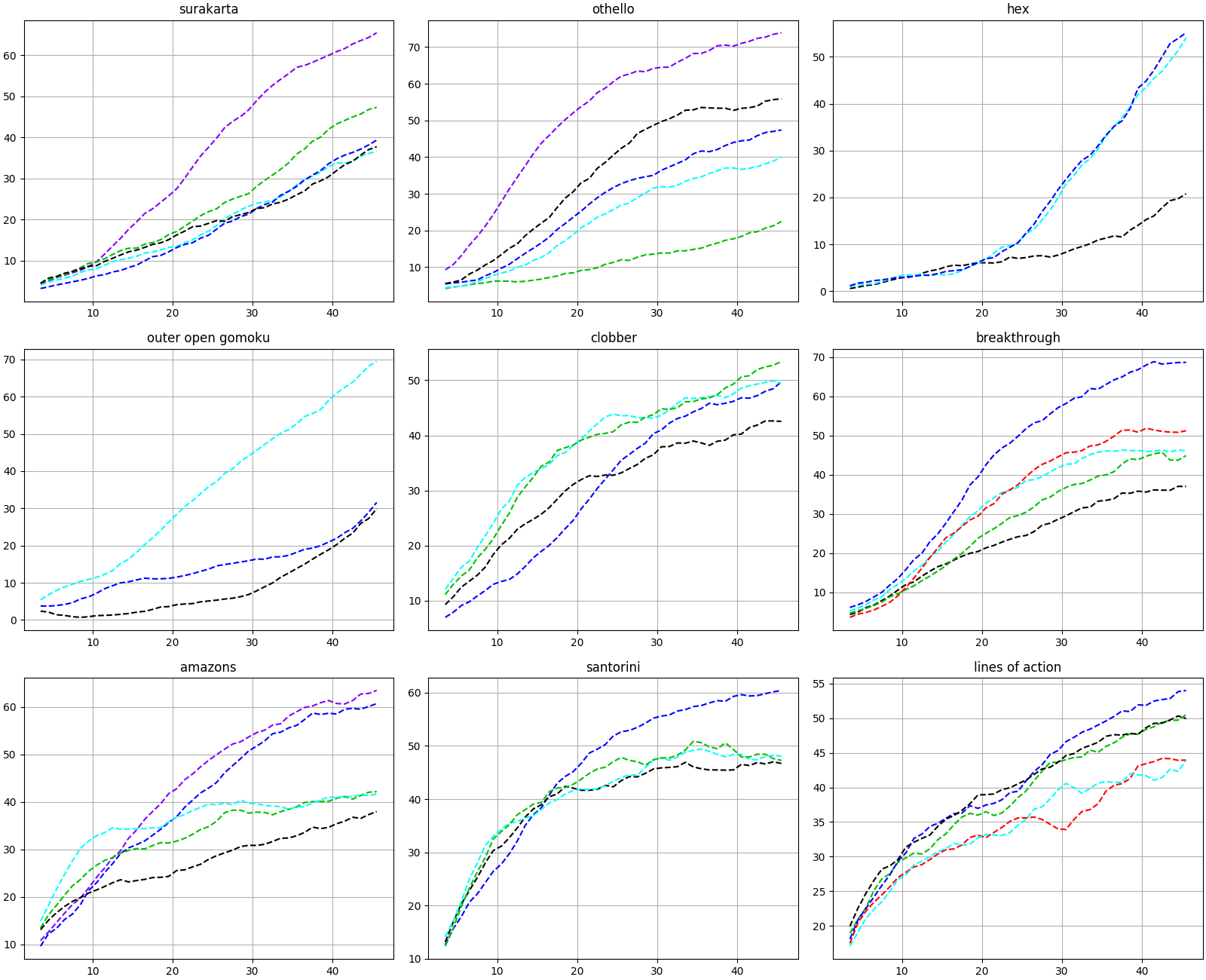}
\par\end{centering}
\caption{Evolutions of the winning percentages of the combinations of the experiment
of Section~\ref{subsec:Comparison-of-Reinforcement}, i.e. the use
of the following heuristics: classic (black line), score (purple line),
additive depth (blue line), multiplicative depth (turquoise line),
cumulative mobility (green line), and presence (red line). The display
uses a simple moving average of 6 data.\label{fig:comparison-h}}
\end{figure}

The final winning percentages are shown in Table~\ref{tab:comparison-h}.
Each percentage of the table has required between $13,824$ and $27,648$
matches. On average and for $7$ of the $9$ games, the classic terminal
heuristic has the worst percentage (exception are Othello and Lines
of Action). In scoring games, scoring is the best heuristic, as we
might expect. Leaving aside the score heuristic, with the exception
of Surakarta, Othello and Clobber, it is one of the two depth heuristics
that has the best winning percentage. In Surakarta and Clobber, mobility
is just ahead of the depth heuristics. On average, using the additive
depth heuristic instead of using the classic evaluation increases
the winning percentage by $15\%$, and using the best depth heuristic
increases the winning percentage by $19\%$. The winning percentage
curves are shown in Figure~\ref{fig:comparison-h}. The final percentages
summarizes the curves quite well. Note that the positive impact of
the additive depth heuristic compared to the other heuristics (except
score) is particulary clear at Breakthrough, Amazons, Hex, and Santorini.
In a similar manner, the positive impact of the multiplicative depth
heuristic is particulary clear at Clobber, Hex, and Open Outer Gomoku. 

In conclusion, the use of generic reinforcement heuristics has significantly
improved performances and the depth heuristics are prime candidates
as a powerful generic reinforcement heuristic. 
\begin{table}
\begin{centering}
\begin{tabular}{|c|c|c|c|c|c|c|}
\hline 
 &  &  & \multicolumn{2}{c|}{depth} &  & \tabularnewline
\hline 
 & classic & score & additive  & multiplicative  & mobility & presence\tabularnewline
\hline 
\hline 
Othello & $49.8\%$  & $70.6\%$  & $50.1\%$  & $48.9\%$  & $18.5\%$  & score\tabularnewline
\hline 
Hex & $33.3\%$  & X & $66.1\%$  & $60.4\%$  & X & X\tabularnewline
\hline 
Clobber  & $43.7\%$  & X & $47.0\%$  & $49.8\%$  & $53.5\%$  & X\tabularnewline
\hline 
Outer Open Gomoku & $33.0\%$  & X & $41.4\%$  & $74.4\%$  & X & X\tabularnewline
\hline 
Amazons & $36.8\%$  & $67.9\%$  & $60.0\%$  & $50.7\%$  & $49.0\%$  & X\tabularnewline
\hline 
Breakthrough & $39.0\%$  & X & $69.5\%$  & $40.4\%$  & $43.9\%$  & $48.5\%$ \tabularnewline
\hline 
Santorini & $42.7\%$  & X & $59.7\%$  & $46.6\%$  & $43.3\%$  & X\tabularnewline
\hline 
Surakarta & $33.5\%$  & $68.9\%$  & $43.1\%$  & $35.3\%$  & $55.6\%$  & score\tabularnewline
\hline 
Lines of Action & $50.9\%$  & X & $57.1\%$  & $46.8\%$  & $53.7\%$  & $44.0\%$ \tabularnewline
\hline 
\textbf{mean} & $40.3\%$  & $69.1\%$  & $54.9\%$  & $50.4\%$  & $45.4\%$  & $46.3\%$ \tabularnewline
\hline 
\end{tabular}
\par\end{centering}
\caption{Final winning percentages of the combinations of the experiment of
Section~\ref{subsec:Comparison-of-Reinforcement} (X: heuristic without
interest in this experiment or for the associated game ; presence
coincides with score in Surakarta and Othello ; $95\%$ confidence
intervals: max $\pm1.04\%$)\label{tab:comparison-h}}

\end{table}

\section{Ordinal Distribution\label{sec:Ordinal-Distribution}\label{subsec:Loi-pour-jouer}}

In this section, we propose another technique, a new action selection
distribution. It is based on a new, alternative, probability distribution,
that we call \emph{ordinal distribution}. 

The ordinal distribution does not depend on the value of states. However,
it depends on the order of their values. On the one hand, this distribution
has the advantage of providing different probability distributions
for actions that are not the best, unlike $\epsilon$-greedy. On the
other hand, it has a more intuitive interpretation than the softmax
function. Indeed, with this distribution, we choose with probability
$p$ to play the best action. If we do not choose it, we choose the
second best action with probability approximately $p$. Otherwise,
we choose the third action with probability approximately $p$, etc.
More precisely, with this distribution, the conditional probability
of playing an action knowing that we do not play a better action is
a linear interpolation between the uniform distribution and the max
distribution (with respect to the exploitation parameter $\epsilon'$).
Its formula is:

\[
P\left(c_{i}\right)=\left(\epsilon'+\frac{1-\epsilon'}{n-i}\right)\cdot\left(1-\sum_{j=0}^{j<i}P\left(c_{j}\right)\right)
\]
with $n$ the number of children of the root, $i\in\{0,\ldots,n-1\}$,
$c_{i}$ the $i$-th best child of the root, $P\left(c_{i}\right)$
the probability of playing the action leading to the child $c_{i}$,
and $\epsilon'$ the exploitation parameter ($\epsilon'=1-\epsilon$).
Algorithm~\ref{alg:ordinal} describes the action selection method
resulting from the use of the ordinal distribution with an optimized
calculation.
\begin{algorithm}[!bh]
\DontPrintSemicolon\SetAlgoNoEnd

\SetKwFunction{ordinal}{ordinal}\SetKwFunction{actions}{actions}\SetKwFunction{premier}{first\_player}\SetKwProg{myproc}{Function}{}{}

\myproc{\ordinal{$s$, $v$}}{

\eIf{\premier{$s$}}{

$A\leftarrow\text{\actions{\text{s}}}$ sorted in descending order
by $a\mapsto v(s,a)$ \;

}{

$A\leftarrow\text{\actions{\text{s}}}$ sorted in ascending order
by $a\mapsto v(s,a)$\;

}

$i\leftarrow0$\;

$n\leftarrow\left|A\right|$\;

\For{$a\in A$ }{

\If{ $\random\leq\left(\frac{t}{T}\cdot\left(n-i-1\right)+1\right)/\left(n-i\right)$
}{

return $a$\;

}

$i\leftarrow i+1$\;

}

}\;

\protect\protect

\caption{Ordinal action distribution algorithm with simulated annealing ($\epsilon'=\frac{t}{T}$)
used in the experiments of this article (see Table~\ref{tab:Index-of-symbols}
for the definitions of symbols).\label{alg:ordinal}}
\end{algorithm}

\begin{rem}
We experimentally compared the ordinal distribution to the$\epsilon$-greedy
distribution and to the softmax distribution on many games. The ordinal
distribution performs better on average and on the majority of games
than the other two distributions (the gain is however quite slight
but the results are solid). However, we do not show these results
in this article. They will be the subject of another publication (this
article is long enough). This additional article should normally be
named: Ordinally Randomized Safe Unbounded Minimax.
\end{rem}

\begin{rem}
We also noticed that the ordinal distribution was easier to tune than
the softmax distribution. In particular, the best parameters for the
different games studied are close, unlike what happens for the softmax
distribution.
\end{rem}

\section{Comparison with ExIt\label{sec:Comparison-with-ExIt}}

In this section, we compare ExIt and Athénan. We start by presenting
the ExIt algorithm (Section~\ref{subsec:The-ExIt-algorithm}). Then,
we compare the algorithmic differences of ExIt and Athénan (Section~\ref{subsec:Technical-Comparison-ExIt-Descent}.)
Next, we present the experiment carried out (Section~\ref{subsec:The-Experimental-Comparison-ExIt})
and its technical details (Section~\ref{subsec:Technical-Details-ExIt}).
Finally, we present the results of this experiment (Section~\ref{subsec:Results-ExIt}).

\subsection{The ExIt Algorithm\label{subsec:The-ExIt-algorithm}}

The ExIt (Expert Iteration) algorithm consists of using an expert
algorithm to generate data and using an apprentice algorithm that
learns that data by supervision to imitate the expert. In order for
the process to improve, the expert must use the apprentice as the
basis of its reasoning. The procedure can then be applied in a self-improving
loop: the apprentice imitates the expert to improve, the expert uses
the apprentice to improve, thereby providing better quality data,
etc. In practice, the expert is a search algorithm and in particular
a modified version of MCTS. The apprentice is a neural network used
to evaluate states and provide a policy. The policy consists of probabilities
of playing actions. It is used by the modified MCTS. The general algorithm
of ExIt is given in Algorithm~\ref{alg:exit}. In order to initialize
the self-improvement procedure. The apprentice is initialized from
data generated using the standard MCTS, i.e. without any knowledge
used.

The data generation strategy is as follows. Perform self-play matches
to obtain the states to be evaluated by the expert serving as learning
targets. In order not to have correlated data, only one game state
per match is used. And so that it is not too costly, it is the apprentice
which is used to perform the self-play matches. One of the states
of each match will then be analyzed by the expert in order to label
this data.

At the beginning of the ExIt process, only the policy is learned and
used. The policy loss, called tree-policy target, used for training
the policy network is given by the following formula:

\[
-\sum_{a}\frac{n_{s,a}}{\sum_{a'}n_{s,a'}}\log\pi\left(a|s\right)
\]
where $n_{s,a}$ is the number of times the action $a$ is selected
during the search in the state $s$, and $\pi\left(a|s\right)$ is
the probability of playing $a$ in $s$ according to the neural network.
The modified UCT value of the MCTS part of ExIt is given by the following
formula:
\[
\mathrm{UCT}_{\mathrm{ExIt}}\left(s,a\right)=\mathrm{UCT}\left(s,a\right)+w_{a}\cdot\frac{\pi\left(a|s\right)}{n_{s,a}+1}
\]

where $\mathrm{UCT}\left(s,a\right)$ is the classical $\mathrm{UCT}$
term of MCTS (which manages the exploration exploitation dilemma~\citep{Coulom06})
and $w_{a}$ is a constant. 

Later during the training, the value network is learned and used in
addition to the policy network. The two losses are thus added for
the rest of the training. The value loss, nammed KL loss, used for
training the value network is given by the following formula:

\[
-z\cdot\log\left(v(s)\right)-\left(1-z\right)\cdot\log\left(1-v(s)\right)
\]
 where $z$ is the result (classic gain) of the match and $v\left(s\right)$
is the value of the state $s$ according to the neural network The
modified UCT value of the MCTS part of ExIt when the value network
is used is given by the following formula:
\[
\mathrm{UCT}_{\mathrm{ExIt}}\left(s,a\right)=\mathrm{UCT}\left(s,a\right)+w_{a}\cdot\frac{\pi\left(a|s\right)}{n_{s,a}+1}+w_{v}\cdot\hat{Q}\left(s,a\right)
\]
where $w_{v}$ is a constant, and $\hat{Q}\left(s,a\right)$ is the
backed up average of the network value after playing $a$ in the state
$s$ during the search.

\begin{algorithm}[!bh]
\DontPrintSemicolon\SetAlgoNoEnd

$\hat{\pi}_{0}\leftarrow\mathrm{initial\_policy\left(\right)}$\;

$\pi_{0}^{*}\leftarrow\mathrm{build\_expert\left(\hat{\pi}_{0}\right)}$\;

\ForEach{$i\in\left\{ 1,\ldots,\mathrm{max\_iterations}\right\} $}{

$S_{i}\leftarrow\mathrm{sample\_self\_play}\left(\hat{\pi}_{i-1}\right)$\;

$D_{i}\leftarrow\liste{\left(s,\mathrm{imitation\_learning\_target}\left(\pi_{i-1}^{*}\left(s\right)\right)\right)}{s\in S_{i}}$\;

$\hat{\pi}_{i}\leftarrow\mathrm{train\_policy}\left(\bigcup_{j\leq i}D_{j}\right)$\;

$\pi_{i}^{*}\leftarrow\mathrm{build\_expert\left(\hat{\pi}_{i}\right)}$\;

}\;

\protect\protect

\caption{Algorithm ExIt (online Expert Iteration).\label{alg:exit}}
\end{algorithm}

\subsection{Algorithmic Comparison between ExIt, AlphaZero and Athénan\label{subsec:Technical-Comparison-ExIt-Descent}}

In this section, we compare the following algorithms with each other:
Athénan, ExIt, and also AlphaZero. We perform a detailed comparison
and then provide a summary.

\subsubsection{Detailed Algorithmic Comparison}

AlphaZero and Athénan share some commonalities that differ with ExIt.
We begin by detailing these characteristics common to Athénan and
AlphaZero that differ with ExIt. On the one hand, unlike ExIt, the
same neural network is kept from the start to the end of the learning
process (it is not reset before each learning phase). Experience replay
is used (although, at least in the context of Athénan, this is optional:
it significantly improves learning performances but learning also
works without it). ExIt uses the \emph{early stopping technique} and,
in its best version, relearns all the data generated from the beginning
at each learning phase (performing the learning phases only with the
last data gave less good results during their experiments). 

Furthermore, ExIt does a \textquotedblleft warm-start\textquotedblright :
it performs a pre-training by learning the data of matches generated
from base MCTS (base MCTS does not use any learned policy or value
fonction: the leaf states are evaluated only by statistics of victory
of random games). Then, ExIt only uses and learns the policy with
its modified MCTS. Finally, it uses and learns the policy and the
value of states with its modified MCTS. It learns in this way more
quickly at the beginning of the training. Athénan performs a pre-training
by learning the data of random terminal states. Like AlphaZero, Athénan
uses mean square error for learning (not the KL loss). 

On the other hand, unlike ExIt, with Athénan and AlphaZero, a search
is performed in each state of the match to determine the move to play.
Moreover, with AlphaZero and Athénan, the data of each state of the
match is used during the learning phases. On the contrary, ExIt plays
matches without search according to the policy of the neural network
in order to perform more games. In addition, ExIt performs a search
for only one state of the match. It therefore uses only one data item
per match for learning. 

We now describe the features that differ between Athénan and ExIt
(and which are similar between ExIt and AlphaZero). During confrontations
(evaluation matches), Athénan uses Unbounded Minimax with Safe Decision:
$\ubfmt$. ExIt and AlphaZero use MCTS. Unlike AlphaZero and ExIt,
with Athénan, all pieces of data of searches, i.e. data of the partial
game tree, is learned. In other words, for learning value of states,
tree learning is used instead of terminal learning (thus (much) more
than one piece of data is learned per search). Therefore, there is
no need for large parallelization to generate a significant number
of data for training (as done for ExIt and AlphaZero). In addition,
unlike ExIt, after each match there is a learning phase: learning
is carried out continuously. The acquired experience is thus immediately
used to directly generate better matches. 
\begin{rem}
Learning the minimax value (performed by Descent) instead of the end-game
value (performed by AlphaZero and ExIt) is more informative (under
the assumption that the state evaluation is of quality). 
\end{rem}

\begin{rem}
Note that in the case of our experiments with Athénan, we do not perform
matches in parallel (it would however be possible to perform such
parallelization). Conversely, ExIt and AlphaZero require massive parallelization
of matches (which is done in the experiments in this paper). However,
with Athénan, the evaluation of the child states of each state of
the game tree is carried out in parallel (on a only GPU), which makes
it possible to speed up searches without loss of efficiency. This
lossless parallelization is not possible with ExIt and AlphaZero.
 More details on this parallelization, called Child Batching, in~\citep{cohen2025study}.
\end{rem}

\subsubsection{Summary}

In summary, the two points that seem most important are that, on the
one hand, Athénan is based on Descent to generate matches, whereas
AlphaZero uses MCTS and ExIt uses the neural network (without search).
On the other, Athénan uses tree learning whereas AlphaZero and ExIt
uses terminal learning and a target policy. In particular, Athénan
does not use a policy, thus there is no need to encode actions (which
poses a problem in certain contexts~\citep{soemers2021deep}). Additionally,
with Athénan, learning is done after each match. Finally, for confrontations,
Athénan uses Unbounded Minimax with Safe Decision but AlphaZero and
ExIt use a variant of MCTS biased to use its policy-value neural network.

\subsection{The Experimental Comparison\label{subsec:The-Experimental-Comparison-ExIt}}

We perform a comparison between ExIt and Athénan. Specifically, for
the training performed with Athénan, we use completed Descent (Algorithm~\ref{alg:descente-completion})
with tree learning (Algorithm~\ref{alg:tree-learning}), completed
ordinal distribution (see Section~\ref{subsec:Completion} and Algorithm~\ref{alg:ordinal}),
the classic gain of the game as reinforcement heuristic\footnote{We could have used the score heuristic when available and the depth
heuristic otherwise, which would have strongly increased Athénan's
performance, but since it might be possible to inject such heuristics
into ExIt, it is fair not to use them.} (see Section~\ref{sec:Reinforcement-Heuristic}), and the stratified
experience replay (Section~\ref{alg:smooth-exp_replay}).

We test these two algorithms on the following games: Go $9\times9$
, Go $11\times11$, Hex $11\times11$, Hex $13\times13$, Othello
$8\times8$, Othello $10\times10$, Connect6, Outer-Open-Gomoku, Havannah
$8$, Havannah $10$. We performed 40 repetitions per game and per
algorithm. Each training process lasted $15$ days.

\subsection{Technical Details\label{subsec:Technical-Details-ExIt}}

\subsubsection{Common Parameters}

We use the following neural network architecture for each of the training
carried out (used as the adaptative evaluation function): a residual
network with a convolutional layer with $132$ filters, followed by
$8$ residual blocks (two $3\times3$ convolutions per block with
$132$ filters each), followed by a flat layer, and followed by two
fully connected hidden layers (with each $N$ neurons), followed by
the final fully connected layer with $1$ neuron (the output). The
activation function used is the ReLU. The value $N$ for each game
is described in Table~\ref{tab:F_N_values_ExIt_Ath=0000E9nan}. The
number of variables in each network is approximately $5\cdot10^{6}$.
The Adam parameters are $\lambda=0.0001$, $\beta_{1}=0.9$, $\beta_{2}=0.999$
and $\epsilon=10^{-7}$. $L_{2}$ regularization is used with $0.001$
as parameter. 
\begin{table}
\begin{centering}
\begin{tabular}{|c|c|}
\hline 
 & $N$\tabularnewline
\hline 
\hline 
Go $9\times9$ & 365\tabularnewline
\hline 
Go $11\times11$ & 228\tabularnewline
\hline 
Hex $11\times11$ & 228\tabularnewline
\hline 
Hex $13\times13$ & 155\tabularnewline
\hline 
Othello $8\times8$ & 477\tabularnewline
\hline 
Othello $10\times10$ & 286\tabularnewline
\hline 
Outer-Open-Gomoku & 111\tabularnewline
\hline 
Connect6 & 65\tabularnewline
\hline 
Havannah $8$ & 111\tabularnewline
\hline 
Havannah $10$ & 65\tabularnewline
\hline 
\end{tabular}\caption{Number of hidden dense neurons $N$ used by ExIt and Athénan for each
studied game.\label{tab:F_N_values_ExIt_Ath=0000E9nan}}
\par\end{centering}
\end{table}

\subsubsection{ExIt Parameters}

The neural network has an additional head used to calculate the policy
(parallel to the dense layers described above). It is composed of
a convolutional $1\times1$ layer with $1$ filter followed by a final
dense layer with as many neurons as there are actions available (i.e.
the number of neurons is the output policy size). Finally, the softmax
function is applied on the output with the temperature parameter $T_{\mathrm{softmax}}$
(provided at the end of this section).

As described in the ExIt paper, the neural networks is initialized
by the data from matches of the base MCTS algorithm against itself.
In this paper, that data were generated and learned during 12 hours
(10 MCTS has been used in parallel by using 10 CPU). From the 3th
day, the value network is learned and used, in addition to the policy
network. We use the default settings of ExIt, except for the number
of rollouts and the number of matches parallelization. We use 2500
rollouts for the MCTS algorithm used by ExIt instead of $10,000$.
Recall that the parallelization of matches is used to perform matches
synchronousy so that their states are evaluated simultaneously by
the neural network on the graphic card. The inference batch size of
this parallelization procedure has been reduced from 1024 to 500 (except
for Othello $10\times10$ where it is reduced to 333). Based on our
testings, these changes have very little impact on performance. However,
the original larger values cause many memory overflows, some of which
are unpredictable. Recall that ExIt has been applied in its introduction
article only to Hex $9\times9$, which is a game with a quite small
board.

For the other parameters, we thus use the default settings: UCT constant
$C$ during the initialization: $0.25$, UCT constant $C$ after the
initialization: $0.05$, learning batch size: $250$, softmax temperature
$T_{\mathrm{softmax}}=0.1$, search constants of ExIt MCTS: $w_{a}=100$
and $w_{v}=0.75$.

\subsubsection{Athénan Parameters}

During training, search time per action is $\tau=3s$. The stratified
experience replay parameters used are: the batch size $B=3000$, the
memory size $\mu=100$, the duplication factor $\delta=3$ . Moreover,
when the children of a state are evaluated by the neural network,
they are batched and thus evaluated in parallel (on the only GPU).
The evaluation function has been pre-initialized by learning the values
of random terminal states (of the order of $10,000,000$). This pre-initialized
lasted $12$ hours and used only one CPU. Resolved states are kept
in memory (the memory of the resolved states is emptied every 12 learning
hours). No simulated annealing was used for the ordinal distribution.
The parameter value of this distribution is a random number drawn
uniformly from {[}0,1{]} each time the ordinal distribution is used.
Other parameters are the same as Section~\ref{subsec:Technical-details}. 

\subsection{Results\label{subsec:Results-ExIt}}

Athénan and ExIt were evaluated against the base MCTS algorithm throughout
their training. An evaluation has been performed approximately every
day (for a total of 15 evaluations). The search time per action during
the evaluations is $2$ seconds. During an evaluation, each algorithm
played 100 matches as first player and 100 more matches as second
player. 

The performance of an algorithm is its win rate minus its loss rate
against MCTS, averaged over the 40 repetitions. The evolution during
the learning process of the average performances over all games of
Athénan and Exit against MCTS are described in Table~\ref{fig:Evolutions-Ath=0000E9nan-vs-ExIt-moy}.
The evolution curves for each game are in the following figures: Go
$9\times9$: Table~\ref{fig:Evolutions-Ath=0000E9nan-vs-ExIt-go9},
Go $11\times11$: Table~\ref{fig:Evolutions-Ath=0000E9nan-vs-ExIt-go11},
Hex $11\times11$: Table~\ref{fig:Evolutions-Ath=0000E9nan-vs-ExIt-hex11},
Hex $13\times13$: Table~\ref{fig:Evolutions-Ath=0000E9nan-vs-ExIt-hex13},
Othello $8\times8$: Table~\ref{fig:Evolutions-Ath=0000E9nan-vs-ExIt-oth8},
Othello $10\times10$: Table~\ref{fig:Evolutions-Ath=0000E9nan-vs-ExIt-oth10},
Connect6: Table~\ref{fig:Evolutions-Ath=0000E9nan-vs-ExIt-con},
Outer-Open-Gomoku: Table~\ref{fig:Evolutions-Ath=0000E9nan-vs-ExIt-gom},
Havannah $8$: Table~\ref{fig:Evolutions-Ath=0000E9nan-vs-ExIt-hava8},
Havannah $10$: Table~\ref{fig:Evolutions-Ath=0000E9nan-vs-ExIt-hava10}. 

The final performance improvement of Athénan compared to ExIt is as
follows: $+62.51\%$ on average over all games, $+162\%$ at Go $9\times9$,
$+158\%$ at Go $11\times11$, $+32\%$ at Hex $11\times11$, $+47\%$
at Hex $13\times13$, $+110\%$ at Othello $8\times8$, $+89\%$ to
Othello $10\times10$, $+9.5\%$ at Outer-Open-Gomoku, $+5.45\%$
to Connect6 (but learning is at least 10 times much faster), $+92\%$
at Havannah $8$, $+23\%$ at Havannah $10$.

In conclusion, Athénan performs better than ExIt on all games and
is strongly better on average and on the majority of games.

\begin{figure}
\begin{centering}
\includegraphics[scale=0.75]{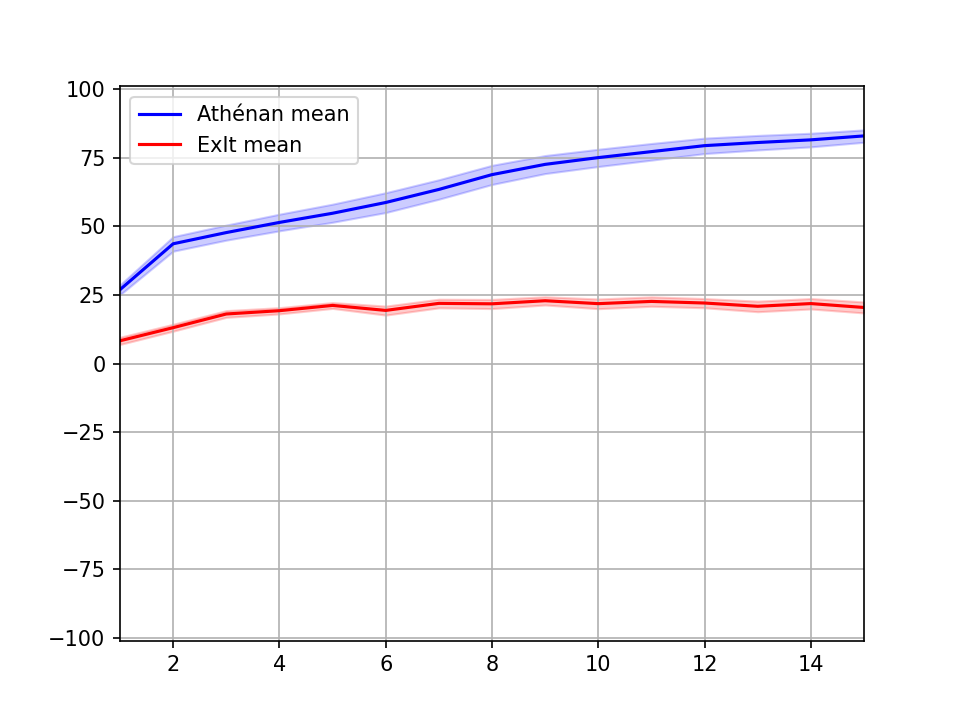}
\par\end{centering}
\caption{Evolutions over 15 training days of performances of the 40 learning
repetitions of Athénan and ExIt against base MCTS over all tested
games (mean of performance and its stratified bootstrapping 5\% confidence
interval).\label{fig:Evolutions-Ath=0000E9nan-vs-ExIt-moy} }

\end{figure}

\begin{figure}
\begin{centering}
\includegraphics[scale=0.75]{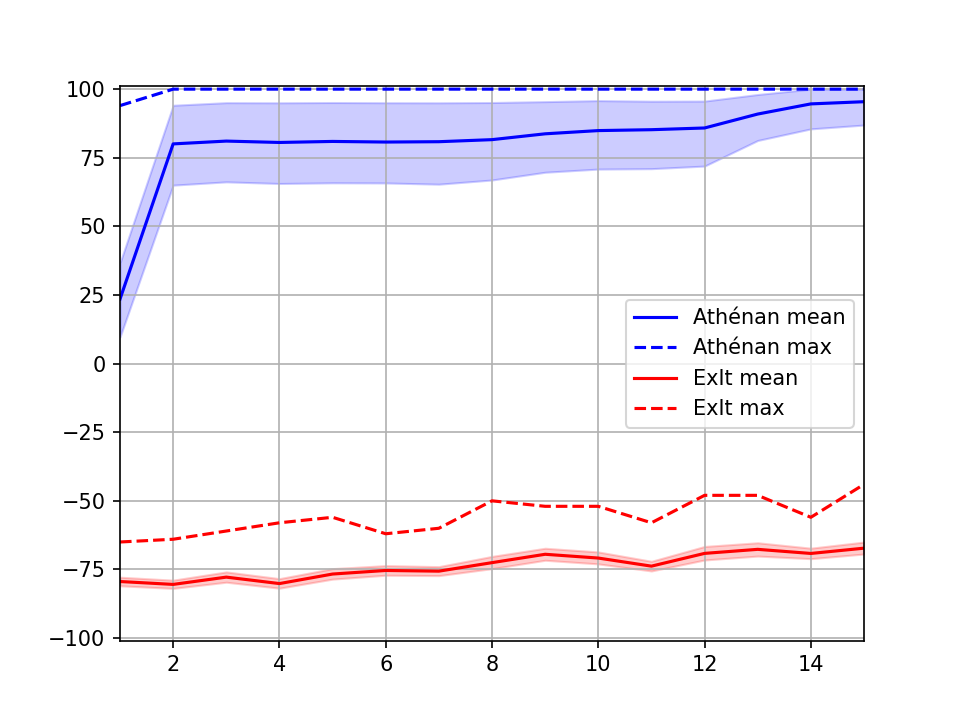}
\par\end{centering}
\caption{Evolutions over 15 training days of performances of the 40 learning
repetitions of Athénan and ExIt against base MCTS for Go $9\times9$
(max of performance and mean of performance and its bootstrapping
5\% confidence interval).\label{fig:Evolutions-Ath=0000E9nan-vs-ExIt-go9} }
\end{figure}

\begin{quotation}
\begin{figure}
\begin{centering}
\includegraphics[scale=0.75]{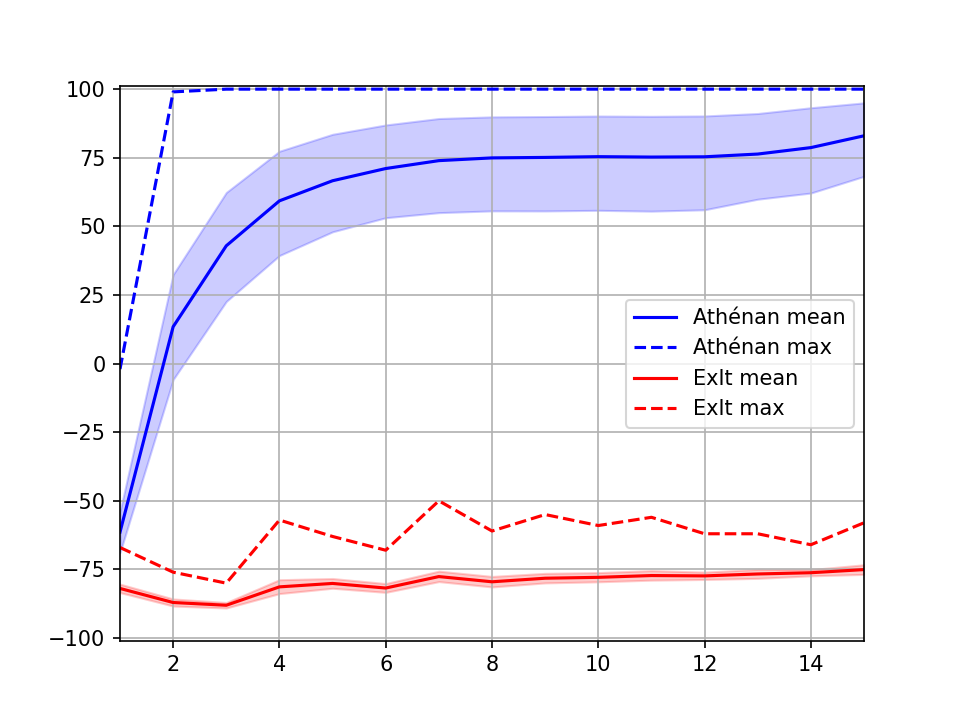}
\par\end{centering}
\caption{Evolutions over 15 training days of performances of the 40 learning
repetitions of Athénan and ExIt against base MCTS for Go $11\times11$
(max of performance and mean of performance and its bootstrapping
5\% confidence interval).\label{fig:Evolutions-Ath=0000E9nan-vs-ExIt-go11} }
\end{figure}
\begin{figure}
\begin{centering}
\includegraphics[scale=0.75]{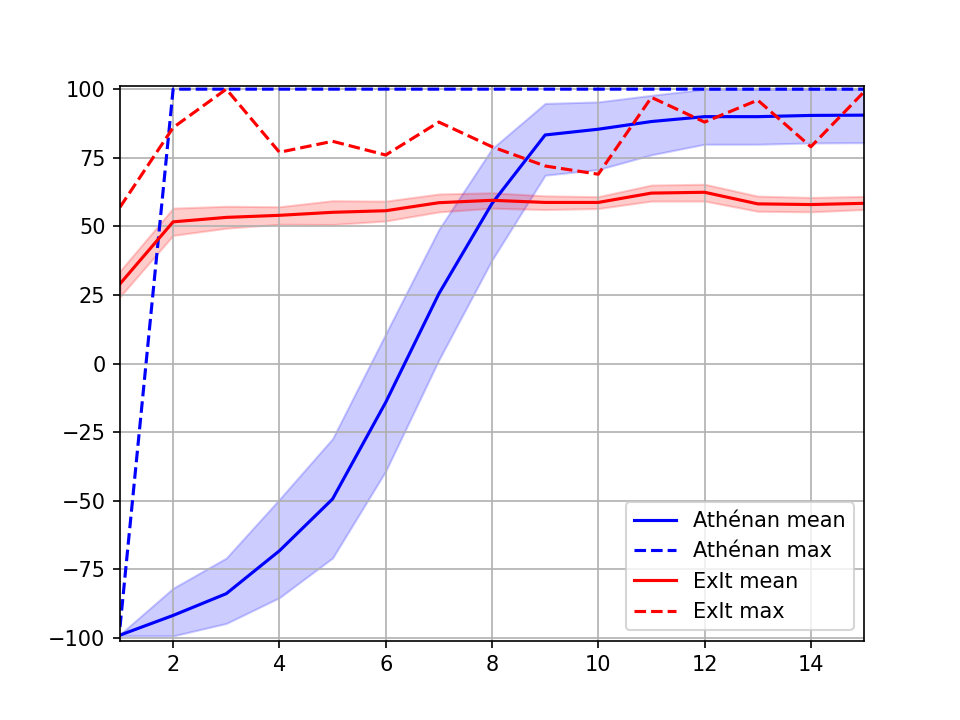}
\par\end{centering}
\caption{Evolutions over 15 training days of performances of the 40 learning
repetitions of Athénan and ExIt against base MCTS for Hex $11\times11$
(max of performance and mean of performance and its bootstrapping
5\% confidence interval).\label{fig:Evolutions-Ath=0000E9nan-vs-ExIt-hex11} }
\end{figure}
\begin{figure}
\begin{centering}
\includegraphics[scale=0.75]{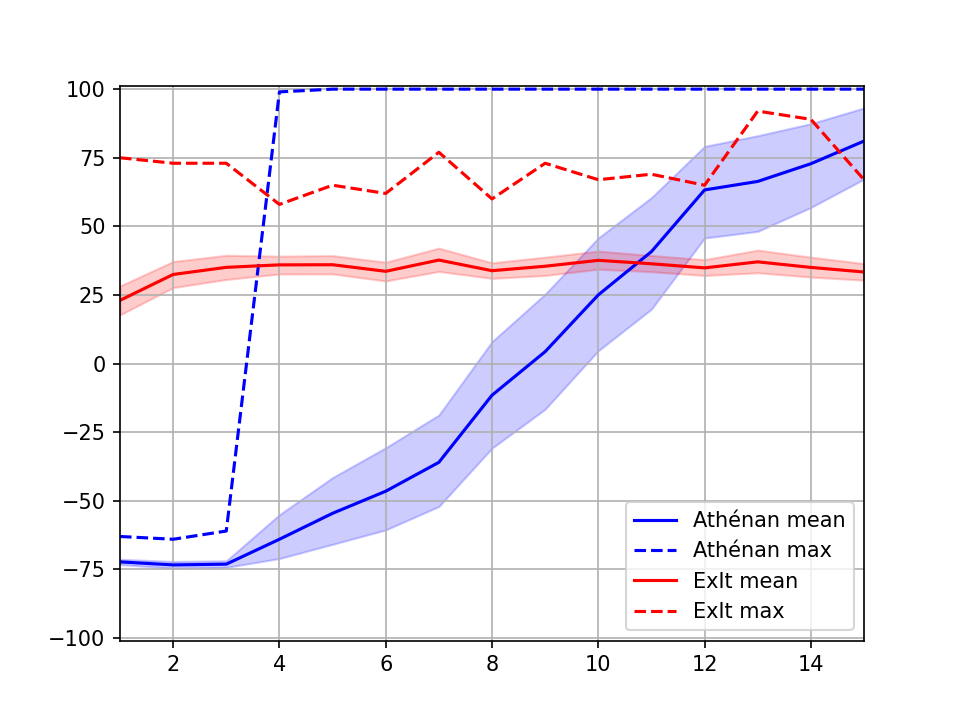}
\par\end{centering}
\caption{Evolutions over 15 training days of performances of the 40 learning
repetitions of Athénan and ExIt against base MCTS for Hex $13\times13$
(max of performance and mean of performance and its bootstrapping
5\% confidence interval).\label{fig:Evolutions-Ath=0000E9nan-vs-ExIt-hex13} }
\end{figure}

\begin{figure}
\begin{centering}
\includegraphics[scale=0.75]{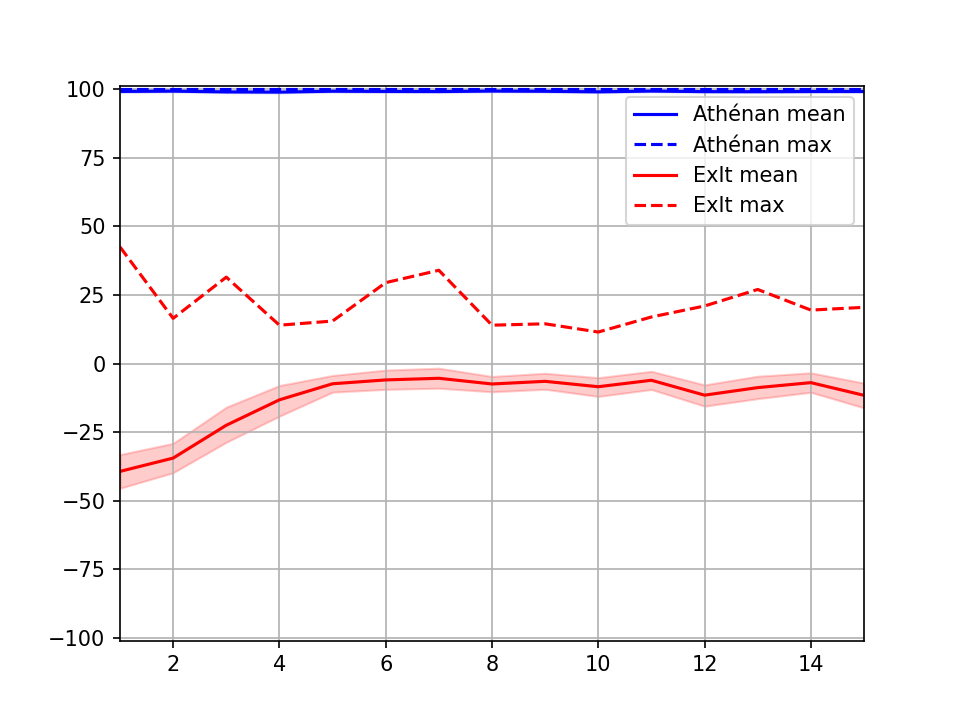}
\par\end{centering}
\caption{Evolutions over 15 training days of performances of the 40 learning
repetitions of Athénan and ExIt against base MCTS for Othello $8\times8$
(max of performance and mean of performance and its bootstrapping
5\% confidence interval).\label{fig:Evolutions-Ath=0000E9nan-vs-ExIt-oth8} }
\end{figure}

\begin{figure}
\begin{centering}
\includegraphics[scale=0.75]{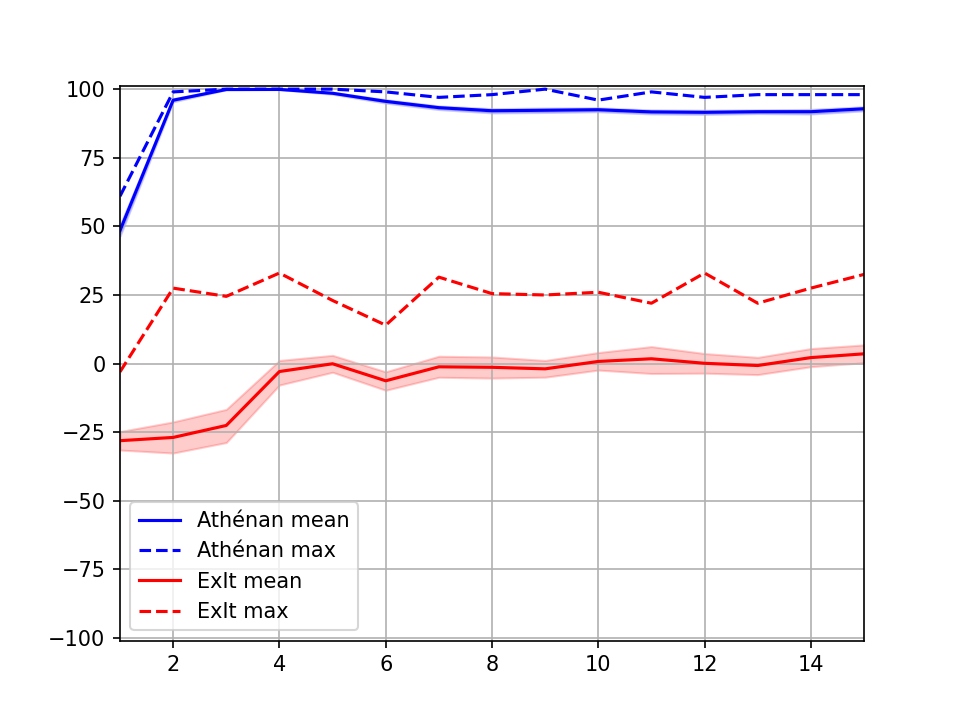}
\par\end{centering}
\caption{Evolutions over 15 training days of performances of the 40 learning
repetitions of Athénan and ExIt against base MCTS for Othello $10\times10$
(max of performance and mean of performance and its bootstrapping
5\% confidence interval).\label{fig:Evolutions-Ath=0000E9nan-vs-ExIt-oth10} }
\end{figure}
\begin{figure}
\begin{centering}
\includegraphics[scale=0.75]{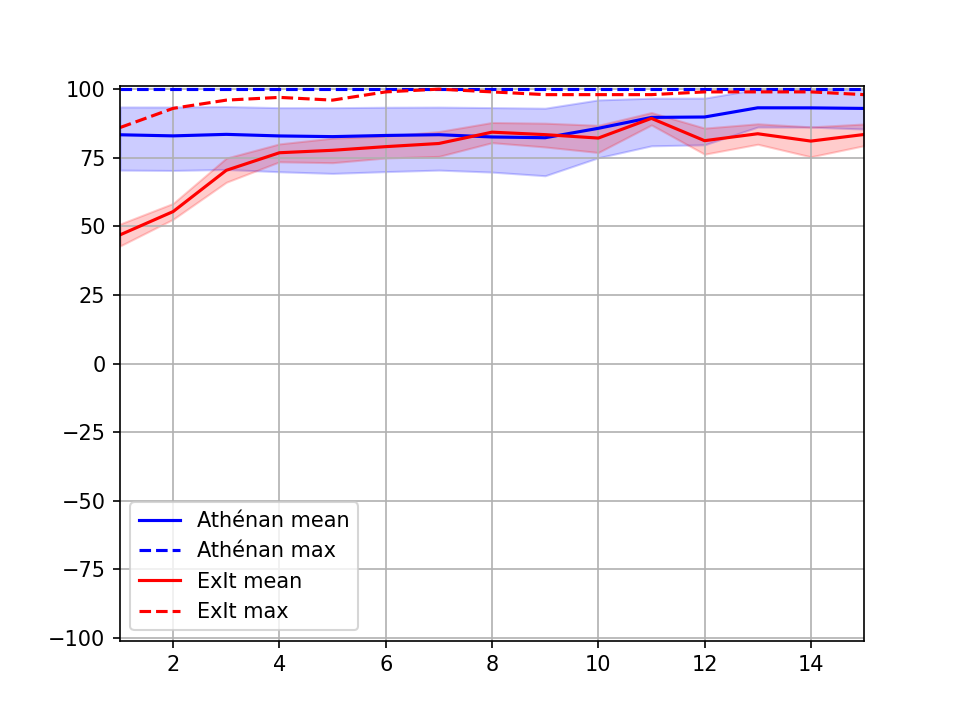}
\par\end{centering}
\caption{Evolutions over 15 training days of performances of the 40 learning
repetitions of Athénan and ExIt against base MCTS for Outer-Open-Gomoku
(max of performance and mean of performance and its bootstrapping
5\% confidence interval).\label{fig:Evolutions-Ath=0000E9nan-vs-ExIt-gom}}
\end{figure}
\begin{figure}
\begin{centering}
\includegraphics[scale=0.75]{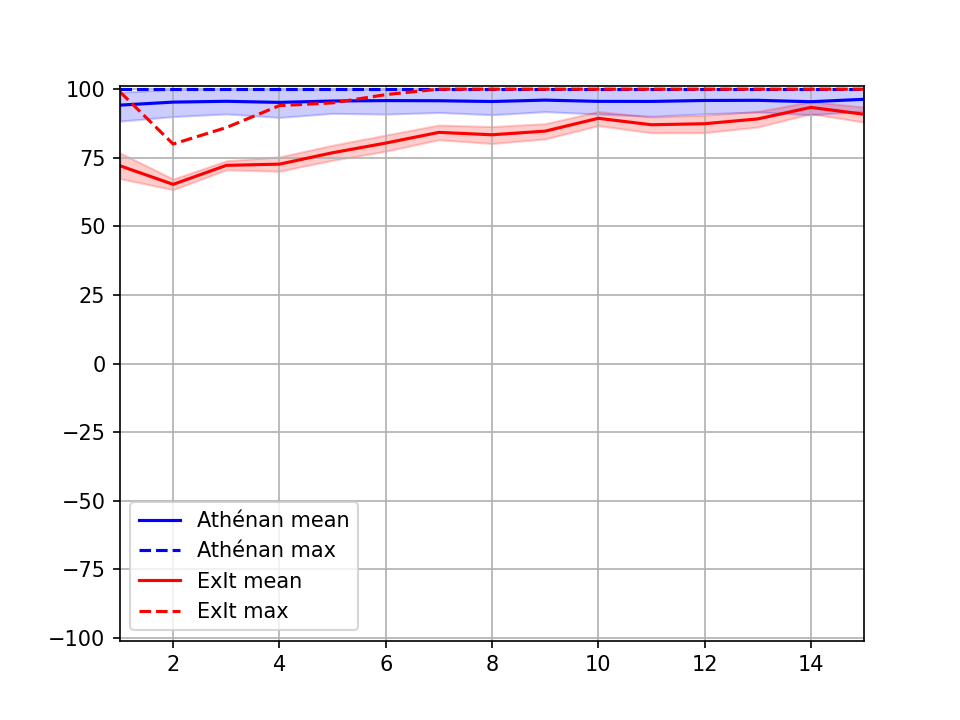}
\par\end{centering}
\caption{Evolutions over 15 training days of performances of the 40 learning
repetitions of Athénan and ExIt against base MCTS for Connect6 (max
of performance and mean of performance and its bootstrapping 5\% confidence
interval).\label{fig:Evolutions-Ath=0000E9nan-vs-ExIt-con}}
\end{figure}
\begin{figure}
\begin{centering}
\includegraphics[scale=0.75]{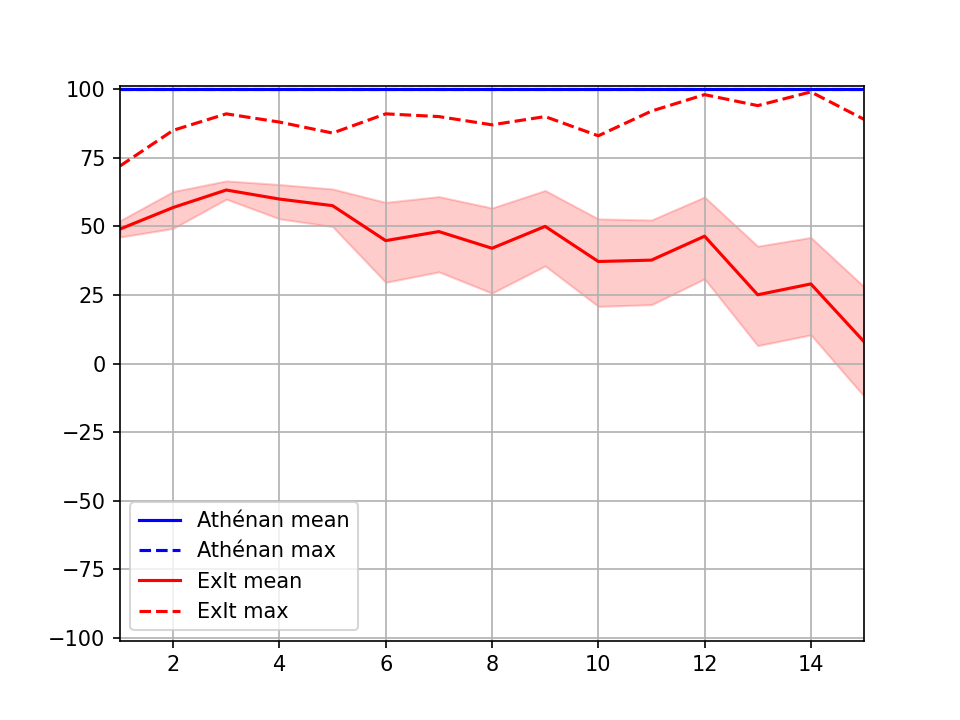}
\par\end{centering}
\caption{Evolutions over 15 training days of performances of the 40 learning
repetitions of Athénan and ExIt against base MCTS for Havannah $8$
(max of performance and mean of performance and its bootstrapping
5\% confidence interval).\label{fig:Evolutions-Ath=0000E9nan-vs-ExIt-hava8} }
\end{figure}
\begin{figure}
\begin{centering}
\includegraphics[scale=0.75]{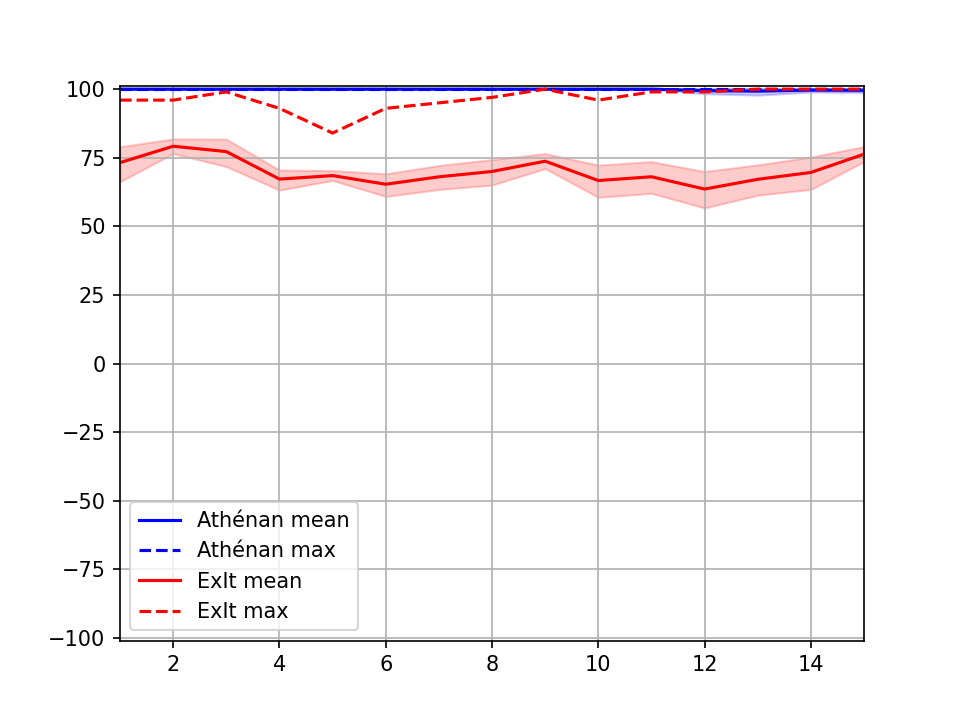}
\par\end{centering}
\caption{Evolutions over 15 training days of performances of the 40 learning
repetitions of Athénan and ExIt against base MCTS for Havannah $10$
(max of performance and mean of performance and its bootstrapping
5\% confidence interval).\label{fig:Evolutions-Ath=0000E9nan-vs-ExIt-hava10} }
\end{figure}
\end{quotation}

\section{Application to Hex\label{sec:Application-to-Hex}}

In this section, we apply Athénan to design program-players to the
game of Hex. We begin by giving details regarding Hex, its rules and
its computational properties. Next, we present the algorithms from
the literature playing to Hex (Section~\ref{subsec:Programmes-joueurs-au-Hex}).
Then, we apply Athénan to Hex with the board size $11\times11$ and
we exceed the level of 3HNN, the best Hex program before Athénan (Section~\ref{subsec:A-Long-Training-11}).
Finally, we apply Athénan to Hex again but this time with the board
size $13\times13$ and we still go beyond the level of 3HNN (Section~\ref{subsec:A-Long-Training-13}).

\subsection{Game of Hex}

\begin{figure}[t]
\begin{centering}
\includegraphics[scale=0.2]{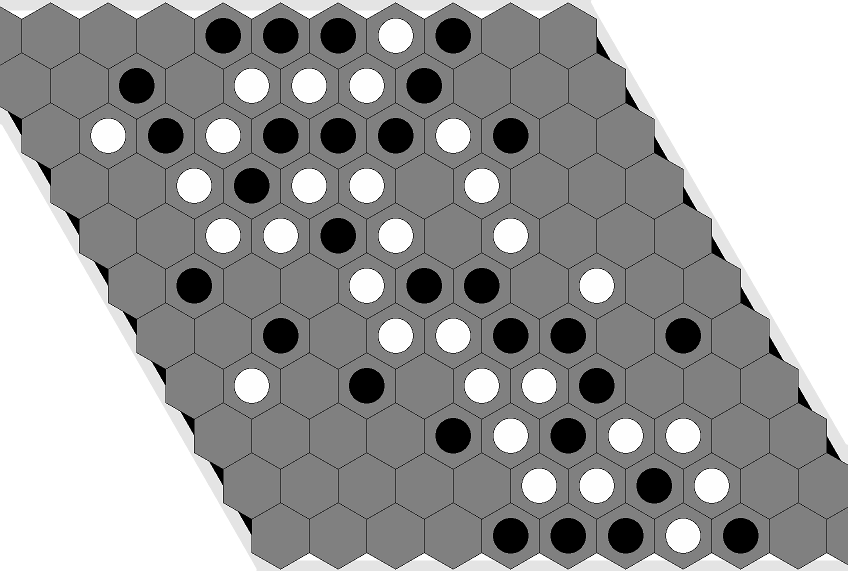}
\par\end{centering}
\centering{}\caption{A Hex end game of size 11 (white wins)\label{fig:Fin-de-partie}}
\end{figure}

The game of Hex \citep{browne2000hex} is a two-player combinatorial
strategy game. It is played on an empty $n\times n$ hexagonal board.
We say that a $n\times n$ board is of size $n$. The board can be
of any size, although the classic sizes are $11$, $13$ and $19$.
In turn, each player places a stone of his color on an empty cell
(each stone is identical). The goal of the game is to be the first
to connect the two opposite sides of the board corresponding to its
color. Figure~\ref{fig:Fin-de-partie} illustrates an end game. Although
its rules are simplistic, Hex tactics and strategies are complex.
The number of states and the number of actions per state are very
large, similar to the game of Go. From the board size $11$, the number
of states is, for example, higher than that of chess (Table~$6$
of \citep{van2002games}). For any board size, the first player has
a winning strategy \citep{berlekamp2018winning} which is unknown,
except for board sizes smaller than or equal to $10$ \citep{pawlewicz2013scalable}
(the game is weaky solved up to the size $10$). In fact, resolving
a particular state is PSPACE-complete \citep{reisch1981hex,bonnet2016complexity}.
There is a variant of Hex using a swap rule. With this variant, the
second player can play in first action a special action, called \emph{swap},
which swaps the color of the two players (i.e. they swap their pieces
and their sides). The swap rule reduces the imbalance between the
two players (without the swap rule, the first player has a very strong
advantage). It is generally used in competitions. It is always used
at the Computer Olympiad and in this paper.

\subsection{Hex Programs\label{subsec:Programmes-joueurs-au-Hex}}

Many Hex player programs have been developed. For example, Mohex 1.0
\citep{huang2013mohex} is a program based on Monte Carlo tree search.
It also uses many techniques dedicated to Hex, based on specific theoretical
results. In particular, it is able to quickly determine a winning
strategy for some states (without expanding the search tree) and to
prune at each state many actions that it knows to be \emph{inferior}.
It also uses ad hoc knowledge to bias simulations of Monte Carlo tree
search.

Mohex 2.0 \citep{huang2013mohex} is an improvement of Mohex 1.0 that
uses learned knowledge through supervised learning (namely correlations
between victory and board patterns) to guide both tree exploration
and simulations.

Other work then focused on predicting best actions, through supervised
learning of a database of games, using a neural network \citep{michalski2013machine,lecun2015deep,goodfellow2016deep}.
The neural network is used to learn a \emph{policy}, i.e. a prior
probability distribution on the actions to play. These prior probabilities
are used to guide the exploration of Monte Carlo tree search. First,
there is Mohex-CNN \citep{gao2017move} which is an improvement of
Mohex 2.0 using a convolutional neural network \citep{krizhevsky2012imagenet}.
A new version of Mohex was then proposed: Mohex-3HNN \citep{gao2018three}.
Unlike Mohex-CNN, it is based on a residual neural network \citep{he2016deep}.
It calculates, in addition to the policy, a value for states \emph{and}
actions. The value of states replaces the evaluation of states based
on simulations of Monte Carlo tree search. Adding a value to actions
allows Mohex-HNN to reduce the number of calls of the neural network,
improving performance. Mohex-3HNN is the best Hex program. It wons
Hex size 11 and 13 tournaments at 2018 Computer Olympiad \citep{gao2019hex}.

Programs which learn the evaluation function by reinforcement have
also been designed. These programs are NeuroHex \citep{young2016neurohex},
EZO-CNN \citep{takada2017reinforcement}, DeepEzo \citep{takada2019reinforcement},
and ExIt \citep{anthony2017thinking}. They learn from self-play.
Unlike the other three programs, NeuroHex performs supervised learning
(of a common Hex heuristic) followed by reinforcement learning. NeuroHex
also starts its matches with a state from a database of games. EZO-CNN
and DeepEzo use knowledge to learn winning strategies in some states.
DeepEzo also uses knowledge during confrontations. ExIt learns a policy
in addition to the value of states and it is based on MCTS (see Section~\ref{sec:Comparison-with-ExIt}
for details). It is the only program to have learned to play Hex without
using knowledge. That result is, however, limited to the board size
$9$. A comparison of the main characteristics of these different
programs is presented in Table~\ref{tab:comparatif}.

\begin{table*}[t]
\begin{centering}
{\footnotesize{}{}}%
\begin{tabular}{|c|c|c|c|c|c|}
\hline 
{\footnotesize{}{}Programs} & {\footnotesize{}{}Size} & {\footnotesize{}{}Search} & {\footnotesize{}{}Learning} & {\footnotesize{}{}Network} & {\footnotesize{}{}Use}\tabularnewline
\hline 
\hline 
{\footnotesize{}{}Mohex-CNN} & {\footnotesize{}{}13} & {\footnotesize{}{}MCTS} & {\footnotesize{}{}supervised} & {\footnotesize{}{}convolutional} & {\footnotesize{}{}policy}\tabularnewline
\hline 
{\footnotesize{}{}Mohex-3HNN} & {\footnotesize{}{}13} & {\footnotesize{}{}MCTS} & {\footnotesize{}{}supervised} & {\footnotesize{}{}residual} & {\footnotesize{}{}policy, state, action}\tabularnewline
\hline 
{\footnotesize{}{}NeuroHex} & {\footnotesize{}{}13} & {\footnotesize{}{}none} & {\footnotesize{}{}supervised, reinforcement} & {\footnotesize{}{}convolutional} & {\footnotesize{}{}state}\tabularnewline
\hline 
{\footnotesize{}{}EZO-CNN} & {\footnotesize{}{}7, 9, 11} & {\footnotesize{}{}Minimax} & {\footnotesize{}{}reinforcement} & {\footnotesize{}{}convolutional} & {\footnotesize{}{}state}\tabularnewline
\hline 
{\footnotesize{}{}DeepEZO} & {\footnotesize{}{}13} & {\footnotesize{}{}Minimax} & {\footnotesize{}{}reinforcement} & {\footnotesize{}{}convolutional} & {\footnotesize{}{}policy, state}\tabularnewline
\hline 
{\footnotesize{}{}ExIt} & {\footnotesize{}{}9} & {\footnotesize{}{}MCTS} & {\footnotesize{}{}reinforcement} & {\footnotesize{}{}convolutional} & {\footnotesize{}{}policy, state}\tabularnewline
\hline 
\end{tabular}
\par\end{centering}
\caption{Comparison of the main features of the latest Hex programs. These
characteristics are respectively the board sizes on which learning
is based, the used tree search algorithm, the type of learning, the
type of neural network, and its use (to approximate the values of
states, actions, and/or policy).\label{tab:comparatif}}
\end{table*}

\subsection{A Long Training for Hex $11$\label{subsec:A-Long-Training-11}}

We now apply all the techniques that we have proposed to carry out
a long self-play reinforcement learning on Hex size 11. More precisely,
we use completed Descent (Algorithm~\ref{alg:descente-completion})
with tree learning (Algorithm~\ref{alg:tree-learning}), completed
ordinal distribution (see Section~\ref{subsec:Completion} and Algorithm~\ref{alg:ordinal}),
and the additive depth heuristic (see Section~\ref{subsec:Depth-Heuristic}). 

\subsubsection{Technical details}

In addition, we use a classical data augmentation: the adding of symmetrical
states. Symmetrical states are added in $D$, the set of pairs $(s,v)$
of the game tree (see Section~\ref{sec:Data-Usage} for details).
The addition of symmetric states is performed after the end of each
match and before the application of experience replay. Formally, $D\leftarrow D\cup\left\{ \left(r_{180\text{°}}\left(s\right),v\right)\ |\ (s,v)\in D\right\} $
where $r_{180\text{°}}\left(s\right)$ is $s$ rotated by $180{^{\circ}}$.
More precisely, the processing($D)$ method of Algorithm~\ref{alg:tree-learning}
is experience\_replay(symmetry($D$), $\mu$, $\sigma$)) where symmetry($D$)
adds symetrical states in $D$ as described above and returns $D$. 

The used learning parameters are: search time per action $\tau=2s$
and batch size $B=3000$. The experience replay parameters are: games
memory size $\mu=100$\footnote{A variant of replay experience was applied, it memorizes the pairs
of the last $100$ games and not the last $100$ pairs.} and sampling rate $\sigma=5\%$. 

We use the following neural network as adaptative evaluation function:
a residual network with a convolutional layer with $83$ filters,
followed by $4$ residual blocks ($2$ convolutions per block with
$83$ filters each), and followed by a fully connected hidden layers
(with $74$ neurons). The activation function used is the ReLU. 

The input of the neural network is a game board extended by one line
at the top, bottom, right and left (in the manner of \citep{young2016neurohex,anthony2017thinking}).
More precisely, each of these lines is completely filled with the
stones of the player of the side where it is located. This extension
is simply used to explicitly represent the sides of the board and
their membership. 

Moreover, when the children of a state are evaluated by the neural
network, they are batched and thus evaluated in parallel (on the only
GPU). 

The evaluation function has been pre-initialized by learning the values
of random terminal states (their number is $15,168,000$). 

The other settings are the same as those in Section~\ref{subsec:Technical-details}.
Resolved states are kept in memory (the memory of the resolved states
is emptied every two learning day).

The reinforcement learning process lasted $34,321$ matches. Note
that the number of data used during the learning process is of the
order of $59\cdot10^{7}$, the number of neural network evaluations
is of the order of $196\cdot10^{6}$, and the number of state evaluations
is of the order of $15\cdot10^{9}$.

\subsubsection{Results}

The winning percentages against Mohex 3HNN of Athénan (i.e. of $\ubfmt$
using the learned evaluation function generated by Descent) are described
in Table~\ref{tab:Winning-percentages-11-1.5}. Note that the proposed
techniques have made it possible to exceed the level of Mohex 3HNN
on Hex size 11, and without the use of knowledge (which had not been
done until then). 
\begin{table}
\begin{centering}
\begin{tabular}{|c|c|c|c|c|c|}
\hline 
search time & Mohex 2nd & Mohex 1st & \textbf{mean} & $95\%$ confidence interval & total matchs\tabularnewline
\hline 
\hline 
$2.5s$ & $98\%$ & $84\%$ & $91\%$ & $2\%$ & $1000$\tabularnewline
\hline 
$10s$ & $91\%$ & $86\%$ & $88\%$ & $2\%$ & $1600$\tabularnewline
\hline 
\end{tabular}
\par\end{centering}
\caption{Winning percentages against Mohex 3HNN of $\protect\ubfmt$ using
the learned evaluation function of Section~\ref{subsec:A-Long-Training-11}
(the search time per action is the same for each player ; default
settings are used for Mohex ; there are as many matches in first
as in second player).\label{tab:Winning-percentages-11-1.5}}
\end{table}

\subsection{A Long Training for Hex $13$\label{subsec:A-Long-Training-13}}

We carry out the same experiment as the previous section, but on Hex
size 13. 

\subsubsection{Technical differences}

 The architecture of the neural network is a convolutional layer
with $186$ filters, followed by $8$ residual blocks ($2$ convolutions
per block with $186$ filters each), and followed by $2$ fully connected
hidden layers (with $220$ neurons each). The activation function
used is the ReLU. The network was not initialized by random end state
values. The experience replay parameters are: games memory size $\mu=250$\footnote{A variant of replay experience was applied, it memorizes the pairs
of the last $250$ games and not the last $250$ pairs.} and sampling rate $\sigma=2\%$. The search time per action $\tau$
is $5s$. The reinforcement learning process lasted $5,288$ matches.
Note that the number of data used during the learning process is of
the order of $16\cdot10^{7}$, the number of neural network evaluations
is of the order of $64\cdot10^{6}$, and the number of state evaluations
is of the order of $7\cdot10^{9}$.

\subsubsection{Results}

The winning percentages against Mohex 3HNN of Athénan (i.e. of $\ubfmt$
using the learned evaluation function generated by Descent) are described
in Table~\ref{tab:Winning-percentages-13-1.5}. Thus, Athénan has
also exceed the level of Mohex 3HNN at Hex size 13 (with swap). This
had not been done before and was achieved without using any prior
knowledge about Hex strategies.

\begin{table}
\begin{centering}
\begin{tabular}{|c|c|c|c|c|c|}
\hline 
search & Mohex 2nd & Mohex 1st & \textbf{mean} & $95\%$ confidence interval & total matchs\tabularnewline
\hline 
\hline 
$2.5s$ & $100\%$ & $100\%$ & $100\%$ & $0\%$ & $1200$\tabularnewline
\hline 
$10s$ & $100\%$ & $100\%$ & $100\%$ & $0\%$ & $800$\tabularnewline
\hline 
\end{tabular}
\par\end{centering}
\caption{Winning percentages against Mohex 3HNN of $\protect\ubfmt$ using
the learned evaluation function of Section~\ref{subsec:A-Long-Training-13}
(the search time per action is the same for each player ; default
settings are used for Mohex ; there are as many matches in first as
in second player).\label{tab:Winning-percentages-13-1.5}}
\end{table}

\section{Application at Othello\label{sec:Application-at-Othello}}

In this section, we apply Athénan on the game of Othello. After completing
the reinforcement learning process without knowledge, we evaluated
Athénan against the state-of-the-art at Othello: the Edax program\footnote{Edax version 4.4: \url{https://github.com/abulmo/edax-reversi}}.
We start by presenting the state-of-the-art at Othello (Section~\ref{subsec:Othello-Related-Work}),
then we describe the experiment carried out (Section~\ref{subsec:Experiment-Othello})
and its technical details (Section~\ref{subsec:Technical-Details-Othello}),
and finally we present the results (Section~\ref{subsec:Results-Othello}).

\subsection{Othello Related Work\label{subsec:Othello-Related-Work}}

Although Othello is one of the first games in which artificial intelligence
has reached a superhuman level, there is still research on Othello.
Edax and Saio are the Othello state-of-the-art programs~\citep{wiki:othello_computer,norelli2022olivaw,liskowski2018learning,wiki:othello_computer2}.
Saio is not free and is similar to Edax. Edax is an open source highly
optimized program. It is based on minimax and dedicated Othello techniques:
\emph{multi-probcut tree search}, \emph{tabular value functions} (as
evaluation function) generated from expert knowledge, and opening
sequences~\citep{buro1997experiments}.

There are other programs at Othello.

A move predictor \citep{liskowski2018learning} was designed at Othello
using a convolutional neural network learned by supervised learning.
However, its level is low. In particular, it only beats Edax when
Edax plans at depth 2 (which takes it on the order of $10^{-4}$ seconds),
whereas Edax plans with a depth between 20 and 30 for a search time
of the order of seconds.

AlphaZero has been applied recently at Othello \citep{norelli2022olivaw}.
This AlphaZero program, named Olivaw, is as strong as Edax with a
search depth of 8. Their program reached a high level at Othello,
and was able to beat a national champion at Othello.

See \citep{wiki:othello_computer,norelli2022olivaw,liskowski2018learning}
for a detailed Related Work about Computer Othello.

\subsection{Experiment\label{subsec:Experiment-Othello}}

We now apply Athénan to carry out a long self-play reinforcement learning
at Othello. More precisely, we use completed Descent (Algorithm~\ref{alg:descente-completion})
with tree learning (Algorithm~\ref{alg:tree-learning}), completed
ordinal distribution (see Section~\ref{subsec:Completion} and Algorithm~\ref{alg:ordinal}),
the game score as reinforcement heuristic (see Section~\ref{sec:Reinforcement-Heuristic}),
and the stratified experience replay (Section~\ref{alg:smooth-exp_replay}).

\subsection{Technical Details\label{subsec:Technical-Details-Othello}}

In addition, we use a classical data augmentation: the adding of symmetrical
states. Symmetrical states are added in $D$, the set of pairs $(s,v)$
of the game tree (see Section~\ref{sec:Data-Usage} for details).
This addition is performed after the end of each game and before the
application of experience replay. Formally, $D\leftarrow D\cup\left\{ \left(\mathrm{sym}\left(s\right),v\right)\ |\ (s,v)\in D\right\} $
where $\mathrm{sym}\left(s\right)$ returns one of the 8 symmetric
board states of $s$. In other words, the processing($D)$ method
of Algorithm~\ref{alg:tree-learning} is experience\_replay(symmetry($D$),
$\mu$, $\sigma$)) where symmetry($D$) adds symetrical states in
$D$ as described above and returns $D$. 

Search time per action is $\tau=5s$. The stratified experience replay
parameters used are: the batch size $B=4000$, the memory size $\mu=100$,
the duplication factor $\delta=3$ . 

We use the following neural network as adaptative evaluation function:
a residual network with a convolutional layer with $373$ filters,
followed by $8$ residual blocks (two $3\times3$ convolutions per
block with $373$ filters each following with a squeeze-and-excitation
layer \citep{hu2018squeeze} whose ratio is $16$), followed by a
$1\times1$ convolution with $4453$ filters, followed by a global
sum pooling \citep{aich2018global}, followed by a flat layer, and
followed by a fully connected hidden layers (with $4453$ neurons).
The activation function used is the ReLU. Number of weights of the
neural network is $41,695,147$. he Adam parameters are $\lambda=0.0001$,
$\beta_{1}=0.9$, $\beta_{2}=0.999$ and $\epsilon=10^{-5}$. The
parameter of the $L_{2}$ regularization is $0.99$. 

Moreover, when the children of a state are evaluated by the neural
network, they are batched and thus evaluated in parallel (on the only
GPU). The evaluation function has been pre-initialized by learning
the values of random terminal states (of the order of $10,000,000$).
Resolved states are kept in memory (the memory of the resolved states
is emptied every 12 learning hours). TNo simulated annealing was used
for the ordinal distribution. The parameter of the ordinal distribution
is a random number drawn uniformly from {[}0,1{]} each time this distribution
is used. The other settings are the same as those in Section~\ref{subsec:Technical-details}.

The reinforcement learning process lasted $52,470$ matches. Note
that the number of data used during the learning process is of the
order of $426\cdot10^{7}$, the number of neural network evaluations
is of the order of $606\cdot10^{6}$, and the number of state evaluations
is of the order of $3.49\cdot10^{9}$.

\subsection{Results\label{subsec:Results-Othello}}

The winning percentages against Edax of Athénan (i.e. of $\ubfmt$
using the learned evaluation function generated by Descent) are described
in Table~\ref{tab:Winning-percentages-othello}. Thus, the proposed
techniques have made it possible to exceed the level of Edax at Othello.
Recall that this reinforcement learning was carried out without using
prior knowledge about Othello strategies.

\begin{table}
\begin{centering}
\begin{tabular}{|c|c|c|c|c|c|c|}
\hline 
search time & win - loss & win & draw & loss & $95\%$ confidence radius & total matchs\tabularnewline
\hline 
\hline 
$1.5s$ & $20\%$ & $39.37\%$ & $41.25\%$ & $19.37\%$ & $3,4\%$ & $800$\tabularnewline
\hline 
$15s$ & $12.25\%$ & $25.37\%$ & $61.5\%$ & $13.12\%$ & $3,4\%$ & $800$\tabularnewline
\hline 
\end{tabular}
\par\end{centering}
\caption{Winning percentages against Edax of $\protect\ubfmt$ using the learned
evaluation function of Section~\ref{subsec:Experiment-Othello} (the
search time per action is the same for each player ; there are as
many matches in first as in second player).\label{tab:Winning-percentages-othello}}
\end{table}

\section{Application at Arimaa\label{sec:Application-at-Arimaa}}

In this section, we apply Athénan on the game of Arimaa and we evaluate
the performed process of reinforcement learning without knowledge
against the state-of-the-art of Arimaa: the Sharp program~\citep{wu2015designing}.
We start by presenting the state-of-the-art at Arimaa (Section~\ref{subsec:Arimaa-Related-Work}),
then we describe the experiment carried out (Section~\ref{subsec:Experiment-Arimaa})
and its technical details (Section~\ref{subsec:Technical-Details-Arimaa}),
and finally we present the results (Section~\ref{subsec:Results-Arimaa}).

\subsection{Arimaa Related Work \label{subsec:Arimaa-Related-Work}}

Since 2015, state-of-the-art at Arimaa is the program Sharp \citep{wu2015designing}.
Sharp combined traditional alpha--beta pruning with handcrafted heuristic
functions. But it uses a lot of improvements of alpha-beta, called
\emph{killer moves}, \emph{history heuristic}, \emph{quiescence search},
\emph{extensions}, \emph{late move reduction}, (theses techniques
required dedicated knowledge to be used). It also uses techniques
dedicated to Arimaa: specific algorithms and knowledge used by alpha-beta
and its improvements (mainly for move ordering, pruning, and state
evaluating). 

Alternative approaches to alpha-beta at Arimaa have been studied,
notably based on Monte Carlo, but without success \citep{wu2015designing}.
There have also been some attempts to apply AlphaZero at Arimaa, but
they have all failed to our knowledge~\citep{arimaasite}.

\subsection{Experiment\label{subsec:Experiment-Arimaa}}

We now apply Athénan to carry out a long self-play reinforcement learning
at Arimaa. More precisely, we use completed Descent (Algorithm~\ref{alg:descente-completion})
with tree learning (Algorithm~\ref{alg:tree-learning}), completed
ordinal distribution (see Section~\ref{subsec:Completion} and Algorithm~\ref{alg:ordinal}),
the stratified experience replay (Section~\ref{alg:smooth-exp_replay}),
and the Arimaa reinforcement heuristic described in the following
section.

\subsection{Arimaa Reinforcement Heuristic}

The reinforcement heuristic (see Section~\ref{sec:Reinforcement-Heuristic})
we used for Arimaa is a variation of the presence heuristic (Section~\ref{sec:Reinforcement-Heuristic})
which takes into account the importance of the pieces which comes
directly from the rules. 

We use the following state evaluation function, denotedy by $\mathrm{h}_{\mathrm{arimaa}}$,
as reinforcement heuristic. Let $s$ be a game state. Then, $\mathrm{h}_{\mathrm{arimaa}}\left(s\right)=\begin{cases}
69+V\left(s\right) & \text{if }\text{first player wins}\\
-69+V\left(s\right) & \text{if }\text{second player wins}
\end{cases}$ where $P_{1}\left(s\right)$ (resp. $P_{2}\left(s\right)$) the set
of pieces of the first player (resp. second player) in $s$, $V\left(s\right)$
is the difference of the values of the two players pieces: 
\[
V\left(s\right)=\sum_{p\in P_{1}\left(s\right)}v_{\mathrm{arimaa}}\left(p\right)-\sum_{p\in P_{2}\left(s\right)}v_{\mathrm{arimaa}}\left(p\right)\text{,}
\]
and $v_{\mathrm{arimaa}}\left(p\right)$ is the value the piece $p$
corresponding to the importance of that piece according to the rule.
The value $v_{\mathrm{arimaa}}\left(p\right)$ is defined as follows:
$v_{\mathrm{arimaa}}\left(\mathrm{elephant}\right)=5$, $v_{\mathrm{arimaa}}\left(\mathrm{camel}\right)=4$,
$v_{\mathrm{arimaa}}\left(\mathrm{horse}\right)=3$, $v_{\mathrm{arimaa}}\left(\mathrm{dog}\right)=2$,
$v_{\mathrm{arimaa}}\left(\mathrm{cat}\right)=1$, $v_{\mathrm{arimaa}}\left(\mathrm{rabbit}\right)=6$.

\subsection{Technical Details\label{subsec:Technical-Details-Arimaa}}

In addition, we use a classical data augmentation: the adding of symmetrical
states. Symmetrical states are added in $D$, the set of pairs $(s,v)$
of the game tree (see Section~\ref{sec:Data-Usage} for details).
This addition is performed after the end of each game and before the
application of experience replay. Formally, $D\leftarrow D\cup\left\{ \left(\mathrm{sym}\left(s\right),v\right)\ |\ (s,v)\in D\right\} $
where $\mathrm{sym}\left(s\right)$ returns one of the symmetric board
states of $s$. In other words, the processing($D)$ method of Algorithm~\ref{alg:tree-learning}
is experience\_replay(symmetry($D$), $\mu$, $\sigma$)) where symmetry($D$)
adds symetrical states in $D$ as described above and returns $D$. 

Search time per action is $\tau=5s$. The stratified experience replay
parameters used are: the batch size $B=4000$, the memory size $\mu=100$,
the duplication factor $\delta=3$ . 

We use the following neural network as adaptative evaluation function:
a residual network with a convolutional layer with $263$ filters,
followed by $4$ residual blocks (two $3\times3$ convolutions per
block with $263$ filters each), followed by a $1\times1$ convolution
with $2217$ filters, followed by a global sum pooling \citep{aich2018global},
followed by a flat layer, and followed by a fully connected hidden
layers (with $2217$ neurons). The activation function used is the
ReLU. Number of weights of the neural network is $10,511,017$. The
Adam parameters are $\lambda=0.0001$, $\beta_{1}=0.9$, $\beta_{2}=0.99$
and $\epsilon=10^{-7}$. 

Moreover, when the children of a state are evaluated by the neural
network, they are batched and thus evaluated in parallel (on the only
GPU). The evaluation function has been pre-initialized by learning
the values of random terminal states (of the order of $10,000,000$).
Resolved states are kept in memory (the memory of the resolved states
is emptied every 12 learning hours). The other settings are the same
as those in Section~\ref{subsec:Technical-details}.

The reinforcement learning process lasted $6,317$ matches. Note that
the number of data used during the learning process is of the order
of $768\cdot10^{7}$, the number of neural network evaluations is
of the order of $10^{9}$, and the number of state evaluations is
of the order of $30\cdot10^{9}$.

\subsection{Results\label{subsec:Results-Arimaa}}

The winning percentages against Sharp of Athénan (i.e. of $\ubfmt$
using the learned evaluation function generated by Descent) are described
in Table~\ref{tab:Winning-percentages-arimaa}. Thus, the proposed
techniques have made it possible to exceed the level of Sharp at Arimaa.
Recall that this reinforcement learning was carried out without using
prior knowledge about Arimaa strategies.
\begin{table}
\begin{centering}
\begin{tabular}{|c|c|c|c|c|}
\hline 
search time & win & loss & $95\%$ confidence interval & total matchs\tabularnewline
\hline 
\hline 
$2.5s$ & $90.33\%$ & $9.66\%$ & $4\%$ & $600$\tabularnewline
\hline 
$10s$ & $92.5\%$ & $7.5\%$ & $4\%$ & $600$\tabularnewline
\hline 
\end{tabular}
\par\end{centering}
\caption{Winning percentages against Sharp of $\protect\ubfmt$ using the learned
evaluation function of Section~\ref{subsec:Experiment-Othello} (the
search time per action is the same for each player ; there are as
many matches in first as in second player).\label{tab:Winning-percentages-arimaa}}
\end{table}

\section{Application at Morpion Solitaire}

In this section, we apply Athénan on the game of \emph{Morpion Solitaire}
(also called \emph{Join Five}), a puzzle game (one-player game). We
start by detailing the rules of Morpion Solitaire (Section~\ref{subsec:Morpion-Solitaire-Rules}),
then presenting the state-of-the-art at Morpion Solitaire (Section~\ref{subsec:Morpion-Solitaire-Related-Worl}).
Next, we describe the experiment carried out (Section~\ref{subsec:Experiment-Morpion})
and its technical details (Section~\ref{subsec:Technical-Details-Morpion}),
and finally we present our results (Section~\ref{subsec:Results-Morpion}).

\subsection{Morpion Solitaire Rules\label{subsec:Morpion-Solitaire-Rules}}

\begin{figure}
\begin{centering}
\includegraphics[scale=0.18]{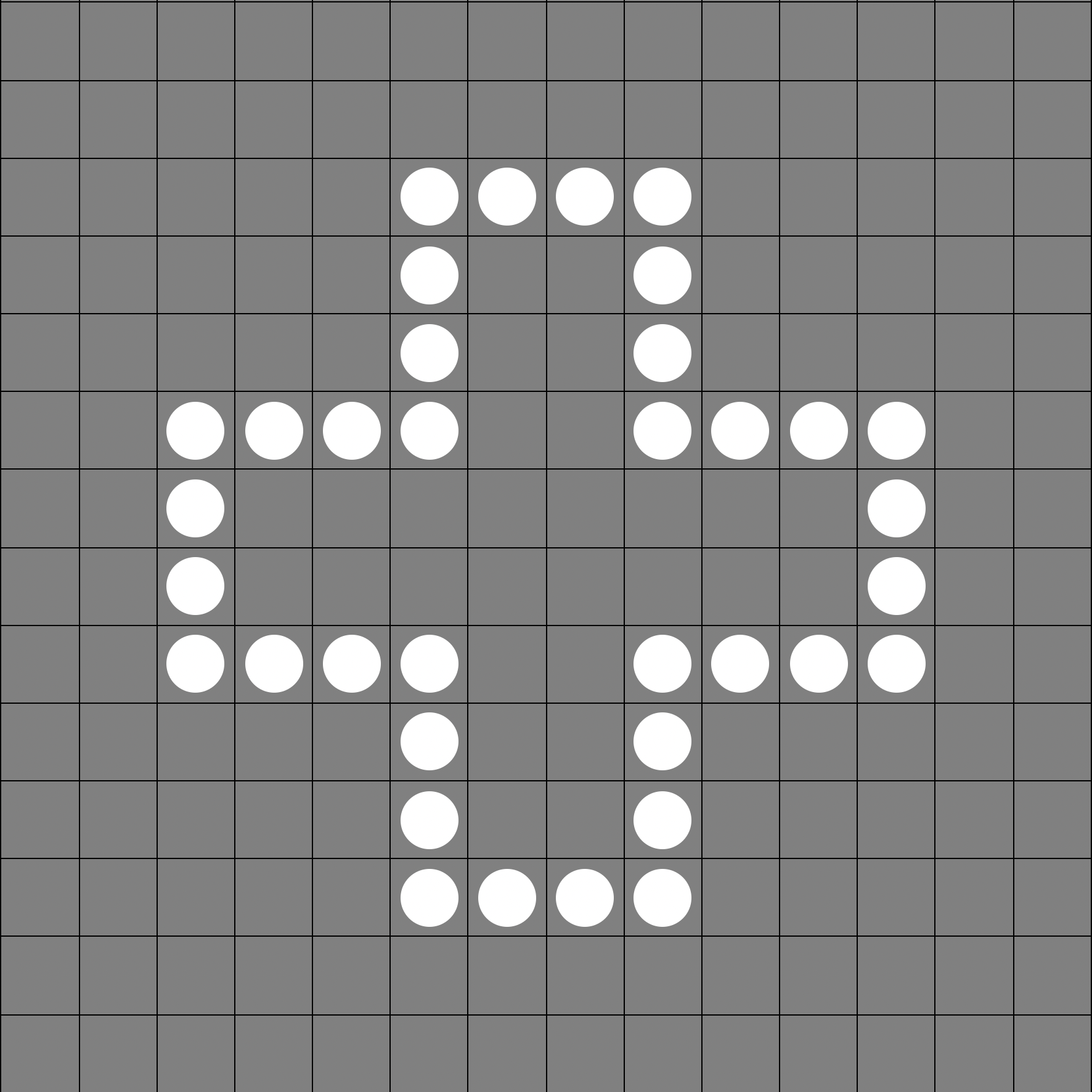}\quad{}\includegraphics[scale=0.18]{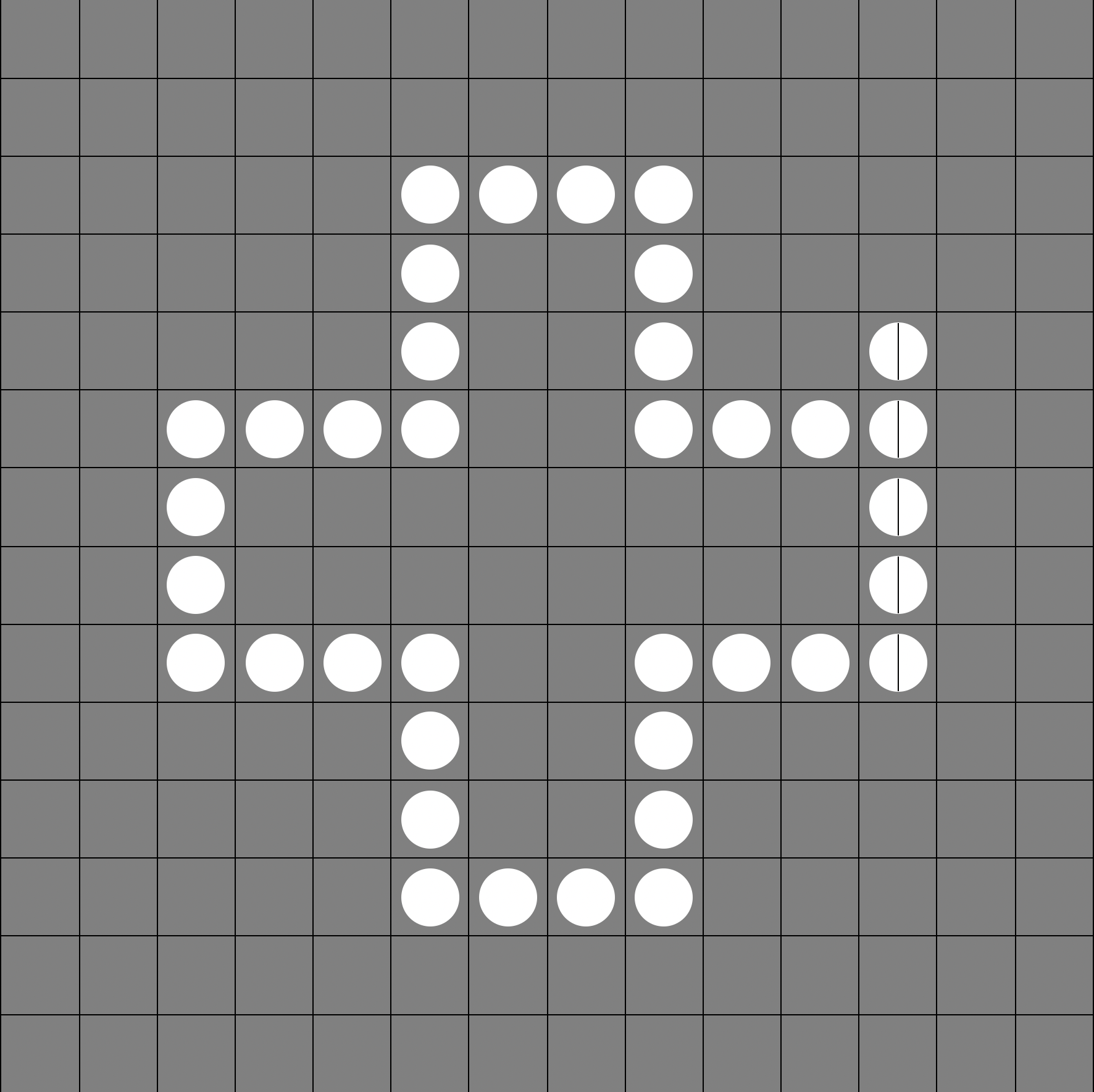}
\par\end{centering}
\begin{centering}
\caption{Initial board for Morpion Solitaire (left) ; The board of Morpion
Solitaire after a first move (right).\label{fig:morpion}}
\par\end{centering}
\centering{}
\end{figure}

The goal of the Morpion Solitaire is to play as long as possible.
The score of the Morpion Solitaire is therefore the number of moves
played since the start of the game. The Morpion Solitaire board is
an infinite size square board. At the start of the game, some points
are already placed in the shape of a cross. At each turn, the player
places a point on the board and crosses out an alignment of 5 points
not already crossed out in that direction. A point can therefore only
be crossed out at most 4 times (horizontally, vertically, diagonally
and antidiagonally). The 5 points must include the point just placed.
A point can only be placed if it will belong to an alignment of 5
points that can be crossed out. A move includes the point placement
plus the choice of the 5 aligned points and their crossing-out. As
soon as the player cannot place any more points, the game is over.
The initial board and the board of an example of first move are represented
in Figure~\ref{fig:morpion}.

The Morpion Solitaire has a variant, called the \emph{touching version}
(the Morpion Solitaire, described above, is called the \emph{disjoint
version}). With the touching variant, the endpoints of an already
drawn line can be used as an endpoint for a new line aligned in the
same direction. This variant is not studied in this paper.

\subsection{Morpion Solitaire Related Work\label{subsec:Morpion-Solitaire-Related-Worl}}

An upper bound on the maximum score at Morpion Solitaire (disjoint
version) is $84$ \citep{michalewski2016upper}. The human record
is 68 \citep{wang2020tackling}. The best actual record is 82. The
record of 82 was obtained with \emph{Nested Rollout Policy Adaptation}
\citep{rosin2011nested} and \emph{Beam Nested Rollout Policy Adaptation}~\citep{cazenave2012beam},
which are Monte Carlo algorithms, using online learning, dedicated
to one-player games.

AlphaZero has been applied with the \emph{Ranked Reward} technique
to Morpion Solitaire~\citep{wang2020tackling}. The record obtained
by this AlphaZero program is only $67$.

ExIt, the reinforcement learning algorithm without knowledge, has
been combined with Nested Rollout Policy Adaptation and applied to
Morpion Solitaire \citep{doux2021deep}. The record of this ExIt program
is $73$.

The touching version has also been studied. The record for this variant
of Morpion Solitaire is 178 \citep{nagorko2019parallel}.

\subsection{Experiment\label{subsec:Experiment-Morpion}}

We now apply Athénan to carry out a long self-play reinforcement learning
without knowledge at Morpion Solitaire. More precisely, we use Descent
(Algorithm~\ref{alg:descente}) with tree learning (Algorithm~\ref{alg:tree-learning}),
ordinal distribution (see Section~\ref{sec:Ordinal-Distribution}),
the stratified experience replay (Section~\ref{alg:smooth-exp_replay}),
and the score heuristic as reinforcement heuristic (Section~\ref{sec:Reinforcement-Heuristic}). 
\begin{rem}
Note that completion is not used since it only concerns two-player
games. However, resolved states are still used. In this context, a
state is resolved only if all its children are resolved.
\end{rem}

\subsection{Technical Details\label{subsec:Technical-Details-Morpion}}

All endgame are draw unless the score of 84 is reached (win).

We use a classical data augmentation: the adding of symmetrical states.
Symmetrical states are added in $D$, the set of pairs $(s,v)$ of
the game tree (see Section~\ref{sec:Data-Usage} for details). This
addition is performed after the end of each game and before the application
of experience replay. Formally, $D\leftarrow D\cup\left\{ \left(\mathrm{sym}\left(s\right),v\right)\ |\ (s,v)\in D\right\} $
where $\mathrm{sym}\left(s\right)$ returns one of the symmetric board
states of $s$. In other words, the processing($D)$ method of Algorithm~\ref{alg:tree-learning}
is experience\_replay(symmetry($D$), $\mu$, $\sigma$)) where symmetry($D$)
adds symetrical states in $D$ as described above and returns $D$. 

Search time per action is $\tau=1s$. The stratified experience replay
parameters used are: the batch size $B=3000$, the memory size $\mu=100$,
the duplication factor $\delta=3$. 

We use the following neural network as adaptative evaluation function:
a residual network with a convolutional layer with $186$ filters,
followed by $8$ residual blocks (two $3\times3$ convolutions per
block with $186$ filters each, following with a squeeze-and-excitation
layer \citep{hu2018squeeze} whose ratio is $16$), followed by a
flat layer, and followed by two fully connected hidden layers (with
each $104$ neurons). The activation function used is the ReLU. Number
of weights of the neural network is $9,989,285$. The Adam parameters
are $\lambda=0.0001$, $\beta_{1}=0.9$, $\beta_{2}=0.99$ and $\epsilon=10^{-8}$. 

Moreover, when the children of a state are evaluated by the neural
network, they are batched and thus evaluated in parallel (on the only
GPU). The evaluation function has been pre-initialized by learning
the values of random terminal states (of the order of $10,000,000$).
Resolved states are kept in memory (the memory of the resolved states
is emptied every 12 learning hours). The other settings are the same
as those in Section~\ref{subsec:Technical-details}.

The reinforcement learning process lasted $58,971$ matches. Note
that the number of data used during the learning process is of the
order of $126\cdot10^{7}$, the number of neural network evaluations
is of the order of $0.19\cdot10^{9}$, and the number of state evaluations
is of the order of $1.8\cdot10^{9}$.

\subsection{Results\label{subsec:Results-Morpion}}

\begin{figure}
\begin{centering}
\includegraphics[scale=0.3]{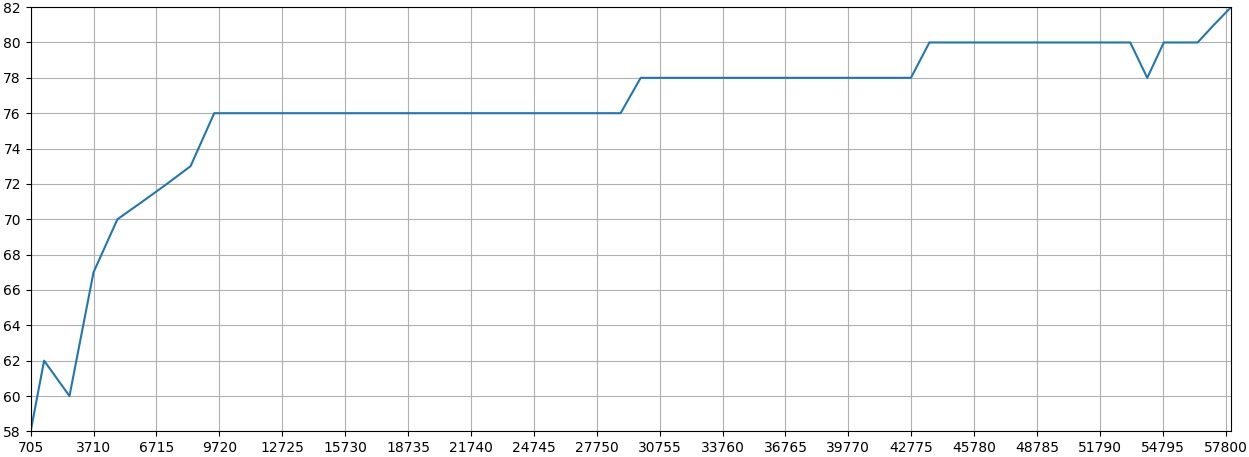}
\par\end{centering}
\centering{}\caption{Evolution of the score obtained in Morpion Solitaire using the evaluation
function learned and $\protect\ubfmt$ with $0.1$ second of search
per action.\label{fig:Morpion_courbe}}
\end{figure}

The evolution of the score obtained in Morpion Solitaire using Athénan
(i.e. using $\ubfmt$ based on the evaluation function generated by
Descent) with $0.1$ second of search per action is described in Figure~\ref{fig:Morpion_courbe}.
The score of $82$ is reached in the $58,971$ match. 

Thus, the state-of-the-art level has been reached with a more general
algorithm than the state-of-the-art techniques on this game. Recall
that this result was not achieved with ExIt and AlphaZero, the two
alternative algorithms of Athénan.

This result further shows that Athénan is applicable to single-player
games.

\section{Computer Olympiad Results\label{sec:Computer-Olympiad-Results}}

In this section, we briefly present Athénan's results in the global
board game artificial intelligence competition, namely the Computer
Olympiad.

Athénan won a lot of gold medals at the Computer Olympiad: 48 in total.
It won eleven gold medals in 2024~\citep{cohen2025athenan}, sixteen
gold medals in 2023~\citep{cohen2023athenan}, five gold medals in
2022, eleven in 2021, and again five in 2020~\citep{cohen2021descent}. 

Moreover, Athénan is currently the defending champion on 17 games
(the eleven 2024 gold medals plus 6 previous uncontested gold medals).
All Athénan results are summarized in Table~\ref{tab:Ath=0000E9nan-Computer-Olympiad}.

Note that, no other program has exceeded 5 gold medals in the same
year since the beginning of this competition in 1989. In fact, reaching
5 medals so far was an exceptional achievement. 

\begin{table}
\begin{centering}
\begin{tabular}{|c|c|c|c|c|c|}
\hline 
 & 2020 & 2021 & 2022 & 2023 & 2024\tabularnewline
\hline 
\hline 
Amazons & Gold & Gold & 1 & Gold & Gold\tabularnewline
\hline 
Breakthrough & Gold & Gold & Gold & Gold & Gold\tabularnewline
\hline 
Clobber & Gold & 1 & 1 & Gold & Gold\tabularnewline
\hline 
Surakarta & Gold & Gold & Gold & Gold & Gold\tabularnewline
\hline 
Othello 8x8 & Silver & Gold & Silver & Bronze & \tabularnewline
\hline 
Othello 10x10 & Gold & 1 & 1 & Gold & 1\tabularnewline
\hline 
Othello 16x16 &  &  &  &  & Gold\tabularnewline
\hline 
Hex 11x11 &  & Gold & 1 & Gold & Gold\tabularnewline
\hline 
Hex 13x13 & 0 & Gold & 1 & Gold & Gold\tabularnewline
\hline 
Hex 19x19 & 0 & Gold & 1 & Gold & 1\tabularnewline
\hline 
Havannah 8 & 0 & Gold & 1 & Gold & Gold\tabularnewline
\hline 
Havannah 10 &  & Gold & 1 & Gold & 1\tabularnewline
\hline 
Canadian Draughts &  & Gold & Gold & Gold & 1\tabularnewline
\hline 
Brazilian Draughts &  & Gold & 0 & Silver & \tabularnewline
\hline 
International Draughts &  & Silver & 0 & Silver & \tabularnewline
\hline 
Connect6 &  & Silver & Silver & Silver & 0\tabularnewline
\hline 
Outer-Open-Gomoku &  & Bronze & Silver & Bronze & 0\tabularnewline
\hline 
Ataxx &  &  & Gold & Gold & Gold\tabularnewline
\hline 
Santorini &  &  & Gold & Gold & 0\tabularnewline
\hline 
Lines of Action &  &  & 0 & Gold & Gold\tabularnewline
\hline 
Xiangqi &  &  & 0 & Gold & 1\tabularnewline
\hline 
Arimaa &  &  &  & Gold & 1\tabularnewline
\hline 
Shobu &  &  &  &  & Gold\tabularnewline
\hline 
\end{tabular}
\par\end{centering}
\caption{Athénan Computer Olympiad results ($0$: participation without result
; $1$: participation without opponent, i.e. title not contested ;
empty: no participation).\label{tab:Ath=0000E9nan-Computer-Olympiad}}

\end{table}

\section{Conclusion\label{sec:Conclusion}}

\subsection{New algorithms: Athénan components}

We have proposed several new techniques for reinforcement learning
state evaluation functions.

Firstly, we have generalized tree bootstrapping (tree learning) in
the context of reinforcement learning without knowledge based on non-linear
functions. We have shown that learning the values of the game tree
instead of just learning the value of the root significantly improves
learning performances.

Secondly, we have introduced the Descent algorithm which explores
in the manner of Unbounded Minimax, intended to be used during the
learning process. Unlike Unbounded Minimax, Descent iteratively explores
the sequences of best actions \emph{up to terminal states}. Its objective
is to improve the quality of the data used during the learning phases,
while keeping the advantages of Unbounded Minimax. In the context
of our experiments, the use of Descent gives strongly better performances
than the use of Alpha-Beta, Unbounded Minimax, or MCTS.

Thirdly, we proposed the completion technique which allows to take
into account the resolution of states, notably in the context of Unbounded
Minimax and Descent. Note that the impact assessment of this technique
was carried out in \citep{cohen2025little}. This technique improves
performance but the gain is lower than the techniques evaluated in
this article.

Fourthly, we have suggested to replace the classic gain of a game
($+1/-1$) by different terminal evaluation functions, called reinforcement
heuristics. We have propose different general terminal evaluations,
such as the depth heuristic, which takes into account the duration
of games in order to favor quick wins and slow defeats. Our experiments
have shown that the use of a reinforcement heuristic improves performances.
Our study recommends using the score heuristic when the game has one
and otherwise using the depth heuristic.

Fifth, we have proposed a new action selection distribution which
does not take into account the value of states but only their order.

We then combined all these techniques within Athénan, a zero-knowledge
reinforcement learning algorithm, like AlphaZero or ExIt.

\subsection{Evaluation of Athénan}

In the unpublished initial version of this article~\citep{cohen2020learning},
Athénan has not been compared to the two state-of-the-art zero-knowledge
reinforcement learning algorithms: AlphaZero and ExIt. In the meantime,
Athénan's comparison with AlphaZero was realized in collaboration
with Tristan Cazenave in another article~\citep{cohen2023minimax}.
In this external study, Athénan with the classic game gain as reinforcement
heuristic is at least 7 times faster than AlphaZero on equal hardware.
It is also at least 30 times faster using the reinforcement heuristics
presented in this article. Athénan with one GPU even outperforms AlphaZero
with 100 GPUs on some games.

In the new version of this paper, we have therefore added the missing
experience: the comparison with ExIt. In the context of the experiments
conducted in this article, Athénan performs much better than ExIt.
ExIt's average performance does not even reach in 15 days the performance
that Athénan obtained in one day. Note that in this study, Athénan
only used the classic game gain as reinforcement heuristic. Thus,
we should achieve a result 4 times higher with the advanced reinforcement
heuristics proposed in this article.

We also showed that Athénan arrived, without using any knowledge\footnote{Without using any knowledge other than the rules of the game or arising
trivially from the rules of the game (namely the use of board symmetry,
encoding of board edge membership in Hex, order of importance of pieces
for the Arimaa reinforcement heuristic).}, to surpass the state-of-the-art of many games, namely Hex size 11
and size 13, Othello, and Arimaa. This contrasts with the outperformed
state-of-the-art programs which use dedicated knowledge.

Moreover, Athénan achieved the state-of-the-art level at Morpion Solitaire.
Note that the state-of-the-art record at Morpion Solitaire was achieved
using algorithms restricted to single-player games. Conversely, this
result shows that in addition to being applicable to two-player games,
Athénan is also applicable to single-player games. This is all the
more noteworthy since attempts to achieve state-of-the-art level at
Morpion Solitaire with ExIt and AlphaZero failed.

Finally, we briefly presented Athénan's results at the Computer Olympiad,
the global artificial intelligence board game competition. Athénan
broke all records at this competition, tripling the record of gold
medals obtained in a single year (reaching 16 gold medals), winning
a total of 48 gold medals in 5 years since its first participation,
and being the defending champion on 17 games.

Overall, the results show that Athénan and its components constitute
particularly efficient and above all effective alternatives to other
approaches based on Alpha-Beta and MCTS, in the era of reinforcement
learning.
\begin{rem}
Athénan has been programmed in Python. Switching to a faster performing
language should reduce the learning time by a factor between two and
five and will increase the winning percentages during confrontations.
\end{rem}

\subsection{Perspectives}

We did not evaluate the ordinal distribution in this article. We
will see in a future article that this techniques improves average
performance. However, its significant gain is much lower than the
other techniques proposed in this article (except completion).

Work is underway on the generalization of these algorithms in the
following contexts: stochastic games~\citep{cohen2023learning},
multiplayer games~\citep{cohen2021completeness}, and usage of a
policy.

A promising research perspective is the parallelization of Unbounded
Minimax and Descent when learning: Unbounded Minimax determines the
next action to play and Descent determines the pairs to learn (by
batching evaluations on a single GPU or using two GPUs). 

Another promising perspective is to perform training using Athénan
with two neural networks playing against each other. This could potentially
reduce numerical problems and the risks of local optimums. To avoid
losing learning speed, on the one hand, each player performs a Descent
search during its opponent's turn. On the other hand, the neural networks
are evaluated in parallel, on the same graphics card (or on two different
ones). This approach can also be generalized to a pool of neural networks
performing matches in parallel. In this context, it might be more
interesting to replace the ordinal distribution with the max distribution.
These approaches should speed up training and increase efficiency,
but they will reduce effectiveness.

The other research perspectives include the application of our contributions
to General Game Playing (at first with perfect information). A yet-to-be-determined
Athénan parameterization, perhaps with the use of small neural networks,
could provide interesting performance in the context of General Game
Playing. A modification of Athénan that could be interesting in this
context is to perform \textquotedbl online\textquotedbl{} learning,
i.e. instead of performing the learning phase after each match, to
perform it after each search, or even after each Descent iteration,
or maybe even after each updating of game tree nodes.

The other research perspectives also include the application and adaptation
of our contributions to the contexts of hidden information. 

Finally, they include the application of our contributions to optimization
problems, such that the RNA Inverse Folding problem \citep{cazenave2020monte}.

\section{Acknowledgements}

This project was provided with computer and storage resources by GENCI
at CINES and IDRIS thanks to the grants 2020-AD010114027, 2020-AD011011772,
2020-AD011011714, 2021-AD011011461R1, 2022-AD011011461R2, 2023-AD011011461R3,
2023-A0141011461, and 2024-A0161011461 on the supercomputers Jean
Zay and Adastra (V100 and MI2050x partitions). 

This work was supported in part by the French government under management
of Agence Nationale de la Recherche as part of the \textquotedblleft Investissements
d\textquoteright avenir\textquotedblright{} program, benchmark ANR19-P3IA-0001
(PRAIRIE 3IA Institute). 

I thank GREYC for giving me access to its computation server, which
allowed me to perform several experiments that inspired those in this
article.

\bibliographystyle{model1-num-names}
\bibliography{jeux}
\newpage{}

\end{document}